\documentclass[12pt,letterpaper,english,letter]{report}
 \pdfoutput=1 
\usepackage[T1]{fontenc}
\usepackage[latin9]{inputenc}
\usepackage{color}
\usepackage{babel}
\usepackage{verbatim}
\usepackage{float}
\usepackage{amsmath}
\usepackage{amsthm}
\usepackage{amssymb}
\usepackage{graphicx}
\usepackage{setspace}
\usepackage[unicode=true,
bookmarks=true,bookmarksnumbered=true,bookmarksopen=true,bookmarksopenlevel=1,
breaklinks=true,pdfborder={0 0 0},pdfborderstyle={},backref=section,colorlinks=true]
{hyperref}
\hypersetup{pdftitle={Qualifying},
 pdfauthor={Isma Hadji},
 pdfkeywords={Template, Qualifying, Dissertation, PhD}}

\makeatletter

\pdfpageheight\paperheight
\pdfpagewidth\paperwidth

\providecommand{\tabularnewline}{\\}

\usepackage[OT1]{fontenc}
\makeatletter
\renewcommand\tableofcontents{%
    \if@twocolumn
      \@restonecoltrue\onecolumn
    \else
      \@restonecolfalse
    \fi
    \chapter*{\contentsname
      \@mkboth{\contentsname}{\contentsname}}%
    \@starttoc{toc}%
    \if@restonecol\twocolumn\fi
    }

\renewcommand\listoffigures{%
    \if@twocolumn
      \@restonecoltrue\onecolumn
    \else
      \@restonecolfalse
    \fi
    \chapter*{\listfigurename
      \@mkboth{\listfigurename}{\listfigurename}}%
    \@starttoc{lof}%
    \if@restonecol\twocolumn\fi
    }

\renewcommand\listoftables{%
    \if@twocolumn
      \@restonecoltrue\onecolumn
    \else
      \@restonecolfalse
    \fi
    \chapter*{\listtablename
      \@mkboth{\listtablename}{\listtablename}}%
    \@starttoc{lot}%
    \if@restonecol\twocolumn\fi
    }

\usepackage{fancyhdr}
\usepackage{epsfig}
\usepackage{cite}
\usepackage{graphicx}
\usepackage{latexsym}
\usepackage{epic}
\usepackage{setspace}
\usepackage{psfrag}
\usepackage{xspace}
\usepackage{amsmath}

\DeclareMathOperator*{\argmax}{arg\,max}

\newcounter{ind}

\makeatletter
\DeclareRobustCommand\onedot{\futurelet\@let@token\@onedot}
\def\@onedot{\ifx\@let@token.\else.\null\fi\xspace}

\def\eg{\emph{e.g}\onedot} 
\def\ie{\emph{i.e}\onedot}

\def\etal{\emph{et al}\onedot}
\makeatother

\makeatother

\begin{document}
\newpage
\thispagestyle{empty}
\vspace*{\fill}
\begin{center}
\textbf{\huge{}What Do We Understand About Convolutional Networks?}
\par\end{center}{\huge \par}
\begin{center}
\LARGE{Isma Hadji and Richard P. Wildes}

\Large{Department of Electrical Engineering and Computer Science}

\Large{York University}

\Large{Toronto, Ontario}

\Large{Canada}
\end{center}
\vspace*{\fill}

\pagenumbering{arabic} \setcounter{page}{1} \pagestyle{fancy}

\chapter{Introduction\label{cha:Introduction}}

\section{Motivation}

Over the past few years major computer vision research efforts have
focused on convolutional neural networks, commonly referred to as
ConvNets or CNNs. These efforts have resulted in new state-of-the-art
performance on a wide range of classification (e.g \cite{Krizhevsky2012,Tran2015,He2016})
and regression (e.g \cite{Szegedy2013,Liu2015,Eigen2015}) tasks.
In contrast, while the history of such approaches can be traced back
a number of years (e.g \cite{Fukushima1980,LeCun1998}), theoretical
understanding of how these systems achieve their outstanding results
lags. In fact, currently many contributions in the computer vision
field use ConvNets as a black box that works while having a very vague
idea for why it works, which is very unsatisfactory from a scientific
point of view. In particular, there are two main complementary concerns:
\textbf{(1)} For learned aspects (e.g convolutional kernels), exactly
what has been learned? \textbf{(2)} For architecture design aspects
(e.g number of layers, number of kernels/layer, pooling strategy,
choice of nonlinearity), why are some choices better than others?
The answers to these questions not only will improve the scientific
understanding of ConvNets, but also increase their practical applicability.

Moreover, current realizations of ConvNets require massive amounts
of data for training \cite{Karpathy2014,Krizhevsky2012,LeCun1998}
and design decisions made greatly impact performance \cite{chatfield2014,Jarret2009}.
Deeper theoretical understanding should lessen dependence on data-driven
design. While empirical studies have investigated the operation of
implemented networks, to date, their results largely have been limited
to visualizations of internal processing to understand what is happening
at the different layers of a ConvNet \cite{Zeiler2014,simonyan14deep,MahendranV15}.

\section{Objective}

In response to the above noted state of affairs, this document will
review the most prominent proposals using multilayer convolutional
architectures. Importantly, the various components of a typical convolutional
network will be discussed through a review of different approaches
that base their design decisions on biological findings and/or sound
theoretical bases. In addition, the different attempts at understanding
ConvNets via visualizations and empirical studies will be reviewed.
The ultimate goal is to shed light on the role of each layer of processing
involved in a ConvNet architecture, distill what we currently understand
about ConvNets and highlight critical open problems.

\section{Outline of report}

This report is structured as follows: The present chapter has motivated
the need for a review of our understanding of convolutional networks.
Chapter \ref{cha:chapter2} will describe various multilayer networks
and present the most successful architectures used in computer vision
applications%
. Chapter \ref{cha:chapter3} will more specifically focus on each
one of the building blocks of typical convolutional networks and discuss
the design of the different components from both biological and theoretical
perspectives. Finally, chapter \ref{cha:chapter4} will describe the
current trends in ConvNet design and efforts towards ConvNet understanding
and highlight some critical outstanding shortcomings that remain.

\chapter{Multilayer Networks\label{cha:chapter2}}

This chapter gives a succinct overview of the most prominent multilayer
architectures used in computer vision, in general. Notably, while
this chapter covers the most important contributions in the literature,
it will not to provide a comprehensive review of such architectures,
as such reviews are available elsewhere (\eg\cite{Goodfellow-et-al-2016,LeCunBengioHinton,buduma2017}).
Instead, the purpose of this chapter is to set the stage for the remainder
of the document and its detailed presentation and discussion of what
currently is understood about convolutional networks applied to visual
information processing.

\section{Multilayer architectures}

{} Prior to the recent success of deep learning-based networks, state-of-the-art
computer vision systems for recognition relied on two separate but
complementary steps. First, the input data is transformed via a set
of hand designed operations (\eg convolutions with a basis set, local
or global encoding methods) to a suitable form. The transformations
that the input incurs usually entail finding a compact and/or abstract
representation of the input data, while injecting several invariances
depending on the task at hand. The goal of this transformation is
to change the data in a way that makes it more amenable to being readily
separated by a classifier. Second, the transformed data is used to
train some sort of classifier (\eg Support Vector Machines) to recognize
the content of the input signal. The performance of any classifier
used is, usually, heavily affected by the used transformations.

Multilayer architectures with learning bring about a different outlook
on the problem by proposing to learn, not only the classifier, but
also learn the required transformation operations directly from the
data. This form of learning is commonly referred to as representation
learning \cite{LeCunBengioHinton,Bengio2013}, which when used in
the context of deep multilayer architectures is called deep learning.

Multilayer architectures can be defined as computational models that
allow for extracting useful information from the input data %
{} multiple levels of abstraction. Generally, multilayer architectures
are designed to amplify important aspects of the input at higher layers,
while becoming more and more robust to less significant variations.
Most multilayer architectures stack simple building block modules
with alternating linear and nonlinear functions. Over the years, a
plethora of various multilayer architectures were proposed and this
section will cover the most prominent such architectures adopted for
computer vision applications. In particular, artificial neural network
architectures will be the focus due to their prominence. For the sake
of succinctness, such networks will be referred to more simply as
neural networks in the following.

\subsection{Neural networks\label{subsec:Neural-Networks}}

\begin{figure}
\begin{centering}
\includegraphics[width=0.4\textwidth]{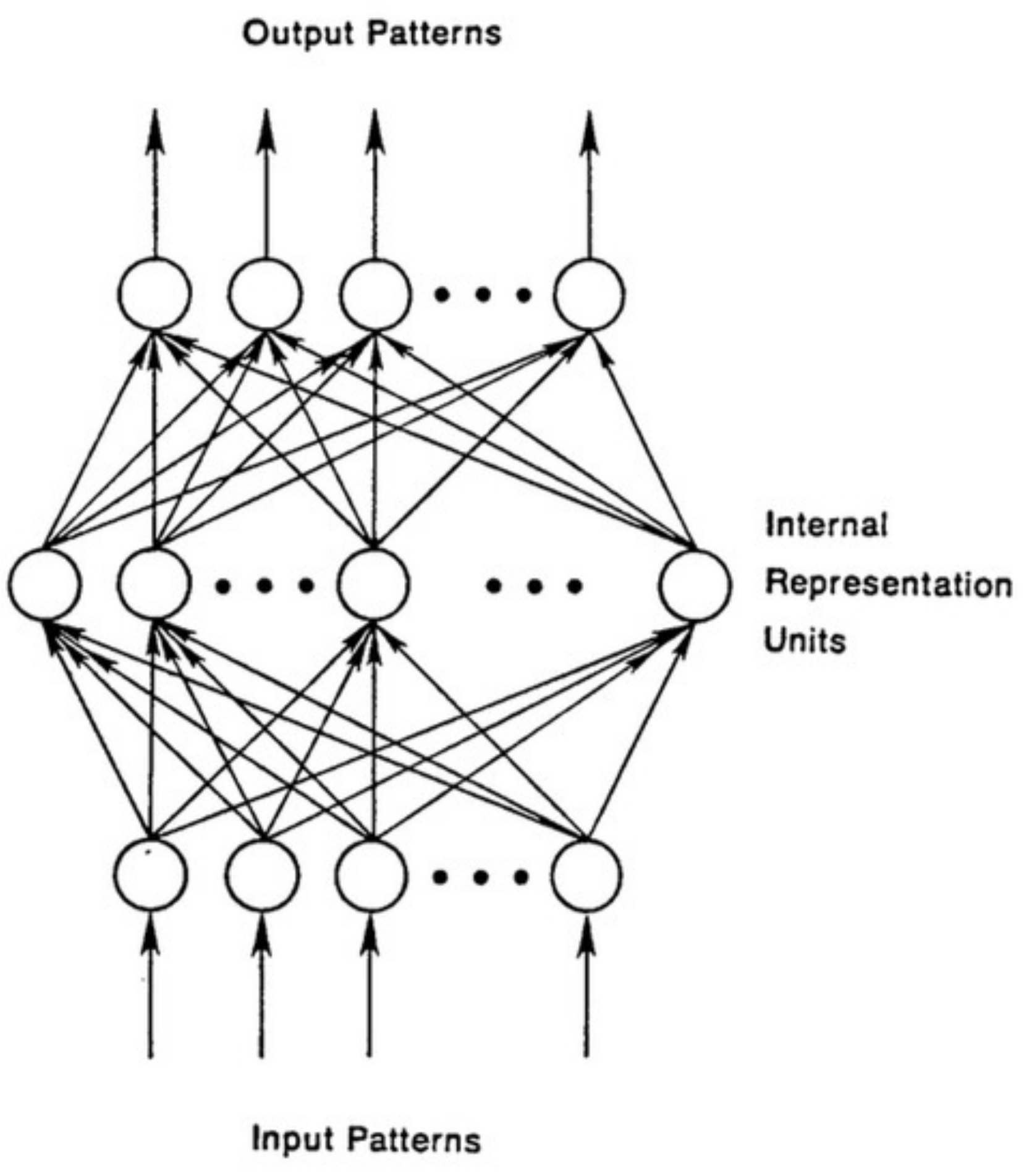}
\par\end{centering}
\caption[Illustration of a typical Neural Network architecture.]{\label{fig:nn} Illustration of a typical Neural Network architecture.
Figure reproduced from \cite{buduma2017}.\vspace{-15pt}}
\end{figure}

A typical neural network architecture is made of an input layer, $\boldsymbol{x}$,
an output layer, $\boldsymbol{y}$, and a stack of multiple hidden
layers, \textbf{$\boldsymbol{h}$}, where each layer consists of multiple
cells or units, as depicted in Figure \ref{fig:nn}. Usually, each
hidden unit, $h_{j}$, receives input from all units at the previous
layer and is defined as a weighted combination of the inputs followed
by a nonlinearity according to 

\begin{equation}
h_{j}=F(b_{j}+\sum_{i}w_{ij}x_{i})\label{eq:nn}
\end{equation}
where, $w_{ij}$, are the weights controlling the strength of the
connections between the input units and the hidden unit, $b_{j}$
is a small bias of the hidden unit and $F(.)$ is some saturating
nonlinearity such as the sigmoid.

Deep neural networks can be seen as a modern day instantiation of
Rosenblatt's perceptron \cite{Rosenblatt1957} and multilayer perceptron
\cite{rosenblatt1962}. Although, neural network models have been
around for many years (\ie since the 1960's) they were not heavily
used until more recently. There were a number of reasons for this
delay. Initial negative results showing the inability of the perceptron
to model simple operations like XOR, hindered further investigation
of perceptrons for a while until their generalizations to many layers
\cite{minsky69perceptrons}. Also, lack of an appropriate training
algorithm slowed progress until the popularization of the backpropagation
algorithm \cite{Rumelhart1986}. However, the bigger roadblock that
hampered the progress of multilayer neural networks is the fact that
they rely on a very large number of parameters, which in turn implies
the need for large amounts of training data and computational resources
to support learning of the parameters. 

A major contribution that allowed for a big leap of progress in the
field of deep neural networks is layerwise unsupervised pretraining,
using Restricted Boltzman Machine (RBM) \cite{Hinton2006}. Restricted
Boltzman Machines can be seen as two layer neural networks where,
in their restricted form, only feedforward connections are allowed.
In the context of image recognition, the unsupervised learning method
used to train RBMs can be summarized in three steps. First, for each
pixel, $x_{i}$, and starting with a set of random weights, $w_{ij}$,
and biases, $b_{j}$, the hidden state, $h_{j}$, of each unit is
set to $1$ with probability, $p_{j}$. The probability is defined
as\vspace{-15pt}

\begin{equation}
p_{j}=\sigma(b_{j}+\sum_{i}x_{i}w_{ij})\label{eq:rbm}
\end{equation}
where, $\sigma(y)=1/(1+exp(-y))$. Second, once all hidden states
have been set stochastically based on equation \ref{eq:rbm}, an attempt
to reconstruct the image is performed by setting each pixel, $x_{i}$,
to $1$ with probability $p_{i}=\sigma(b_{i}+\sum_{j}h_{j}w_{ij})$.
Third, the hidden units are corrected by updating the weights and
biases based on the reconstruction error given by\vspace{-25pt}

\begin{equation}
\Delta w_{ij}=\alpha(\left\langle x_{i}h_{j}\right\rangle _{input}-\left\langle x_{i}h_{j}\right\rangle _{reconstruction})\label{eq:rbm-1}
\end{equation}
where $\alpha$ is a learning rate and $\left\langle x_{i}h_{j}\right\rangle $
is the number of times pixel $x_{i}$ and the hidden unit $h_{j}$
are on together. The entire process is repeated $N$ times or until
the error drops bellow a pre-set threshold, $\tau$. After one layer
is trained its outputs are used as an input to the next layer in the
hierarchy, which is in turn trained following the same procedure.
Usually, after all the network's layers are pretrained, they are further
finetuned with labeled data via error back propagation using gradient
descent \cite{Hinton2006}. Using this layerwise unsupervised pretraining
allows for training deep neural networks without requiring large amounts
of labeled data because unsupervised RBM pretraining provides a way
for an empirically useful initialization of the various network parameters.

\begin{figure}[H]
\begin{centering}
\includegraphics[width=0.45\textwidth]{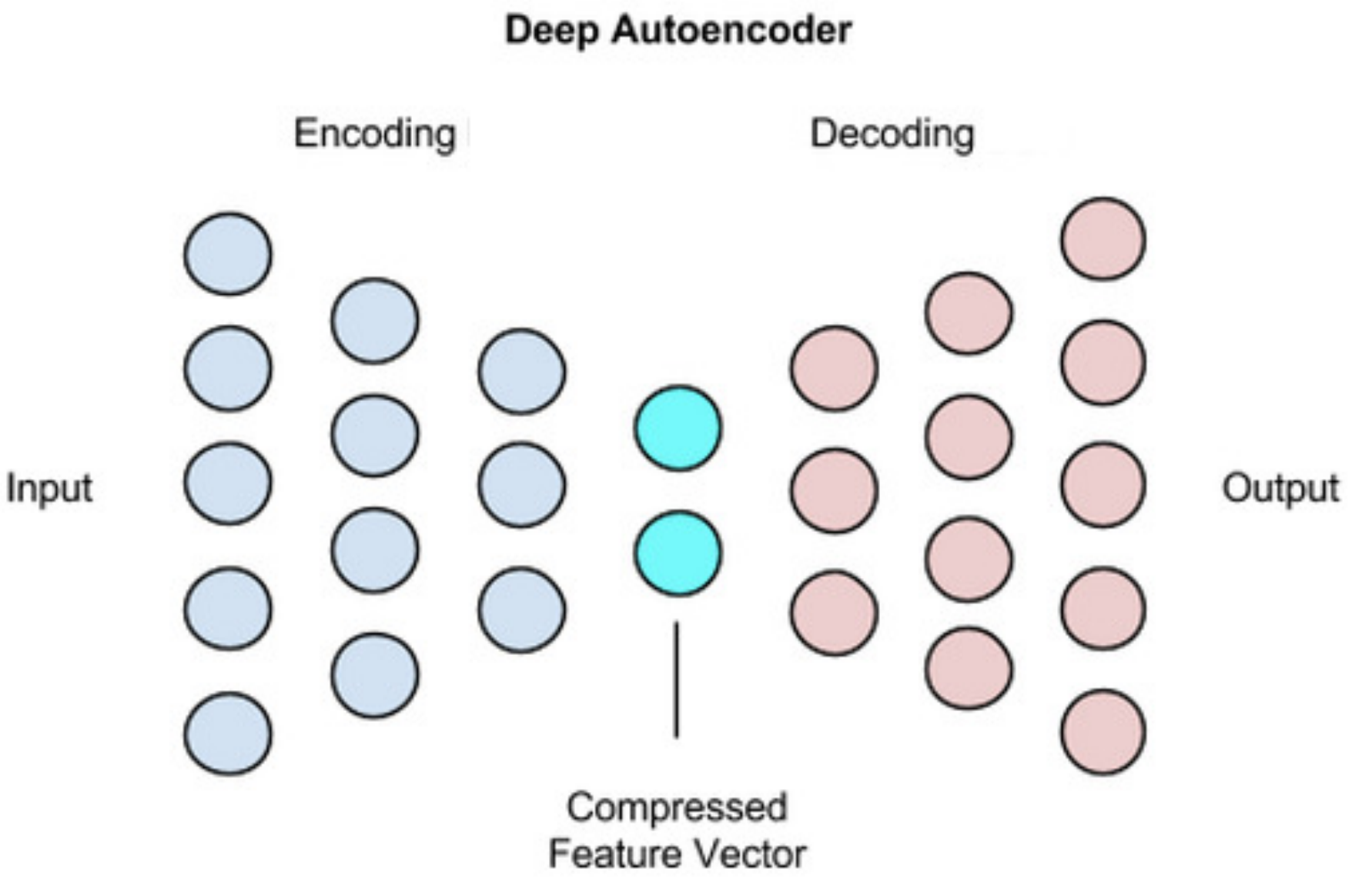}
\par\end{centering}
\caption[Structure of a typical Auto-Encoder Network.]{\label{fig:autoencoders}Structure of a typical Auto-Encoder Network.
Figure reproduced from \cite{buduma2017}.}
\end{figure}

Neural networks relying on stacked RBMs were first successfully deployed
as a method for dimensionality reduction with an application to face
recognition \cite{HintonSalakhutdinov2006b}, where they were used
as a type of auto-encoder. Loosely speaking, auto-encoders can be
defined as multilayer neural networks that are made of two main parts:
First, an encoder transforms the input data to a feature vector; second,
a decoder maps the generated feature vector back to the input space;
see, Figure \ref{fig:autoencoders}. The parameters of the auto-encoder
are learned by minimizing a reconstruction error between the input
and it's reconstructed version.\vspace{-15pt}

Beyond RBM based auto-encoders, several types of auto-encoders were
later proposed. Each auto-encoder introduced a different regularization
method that prevents the network from learning trivial solutions even
while enforcing different invariances. Examples include Sparse Auto-Encoders
(SAE) \cite{Bengio2007}, Denoising Auto-Encoders (DAE) \cite{Vincent2008,Vincent2010}
and Contractive Auto-Encoders (CAE) \cite{Rifai2011}. Sparse Auto-Encoders
\cite{Bengio2007} allow the intermediate representation's size (\ie
as generated by the encoder part) to be larger than the input's size
while enforcing sparsity by penalizing negative outputs. In contrast,
Denoising Auto-Encoders \cite{Vincent2008,Vincent2010} alter the
objective of the reconstruction itself by trying to reconstruct a
clean input from an artificially corrupted version, with the goal
being to learn a robust representation. Similarly, Contractive Auto-Encoders
\cite{Rifai2011} build on denoising auto-encoders by further penalizing
the units that are most sensitive to the injected noise. More detailed
reviews of various types of auto-encoders can be found elsewhere \cite{Bengio2013}.

\subsection{Recurrent neural networks}

\begin{figure}
\begin{centering}
\includegraphics[width=0.6\textwidth]{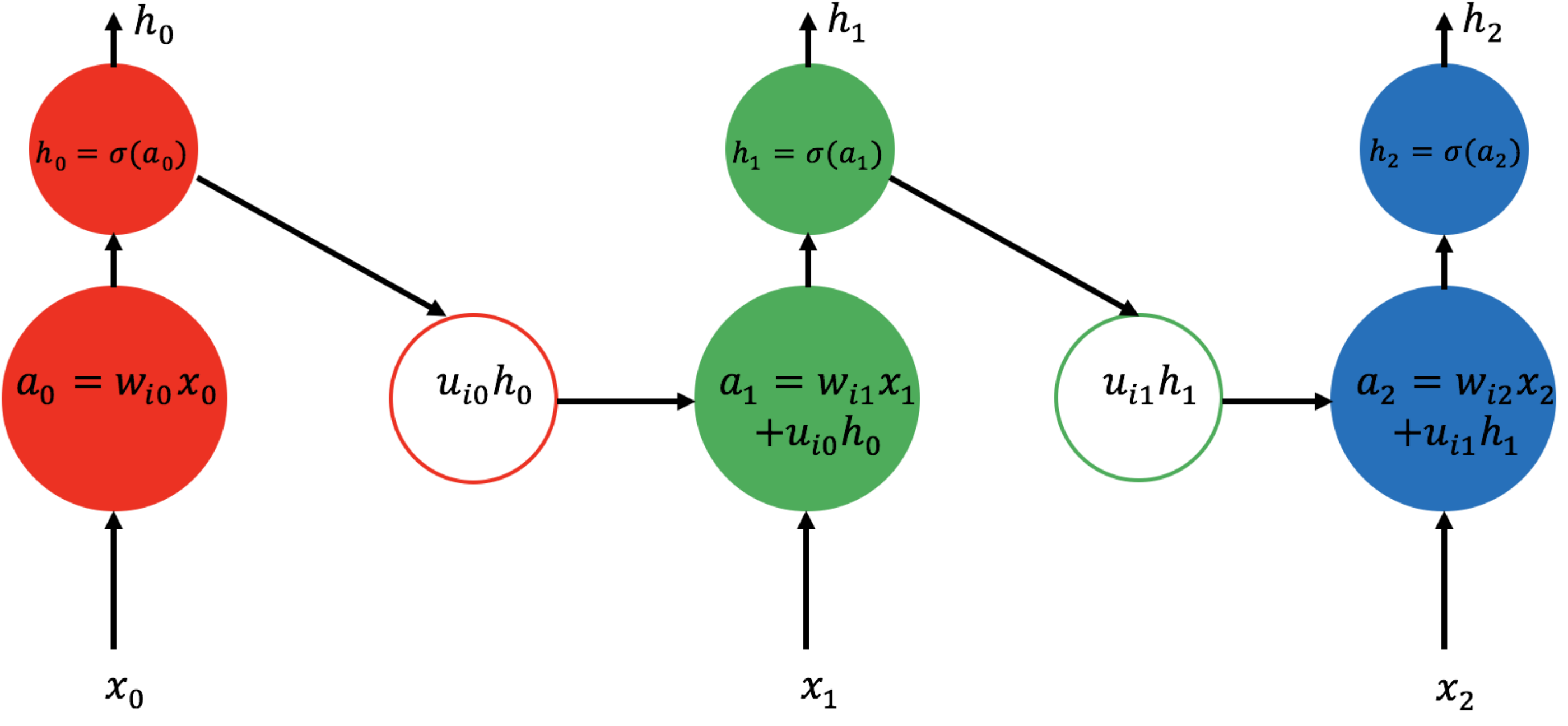}
\par\end{centering}
\caption[Illustration of the operations of a standard Recurrent Neural Network.]{\label{fig:rnn}Illustration of the operations of a standard Recurrent
Neural Network. Each RNN unit takes new input at the current time
frame, $x_{t}$, and from a previous time step, $h_{t-1}$ and the
new output of the unit is calculated according to \eqref{eq:rnn}
and can be fed to another layer of processing in a multilayer RNN.}
\end{figure}

When considering tasks that rely on sequential inputs, one of the
most successful multilayer architectures is the Recurrent Neural Network
(RNN) \cite{Bengio1994}. RNNs, illustrated in Figure \ref{fig:rnn},
can be seen as a special type of neural network where each hidden
unit takes input from the the data it observes at the current time
step as well as its state at a previous time step. The output of an
RNN is defined as

\begin{equation}
h_{t}=\sigma(w_{i}x_{t}+u_{i}h_{t-1})\label{eq:rnn}
\end{equation}
where $\sigma$ is some nonlinear squashing function and $w_{i}$
and \textbf{$u_{i}$} are the network parameters that control the
relative importance of the present and past information.\textbf{ }

Although RNNs are seemingly powerful architectures, one of their major
problems is their limited ability to model long term dependencies.
This limitation is attributed to training difficulties due to exploding
or vanishing gradient that can occur when propagating the error back
through multiple time steps \cite{Bengio1994}. In particular, during
training the back propagated gradient is multiplied with the network's
weights from the current time step all the way back to the initial
time step. Therefore, because of this multiplicative accumulation,
the weights can have a non-trivial effect on the propagated gradient.
If weights are small the gradient vanishes, whereas larger weights
lead to a gradient that explodes. To correct for this difficulty,
Long Short Term Memories (LSTM) were introduced \cite{Hochreiter1997}. 

\begin{figure}[H]
\begin{centering}
\includegraphics[width=0.5\textwidth]{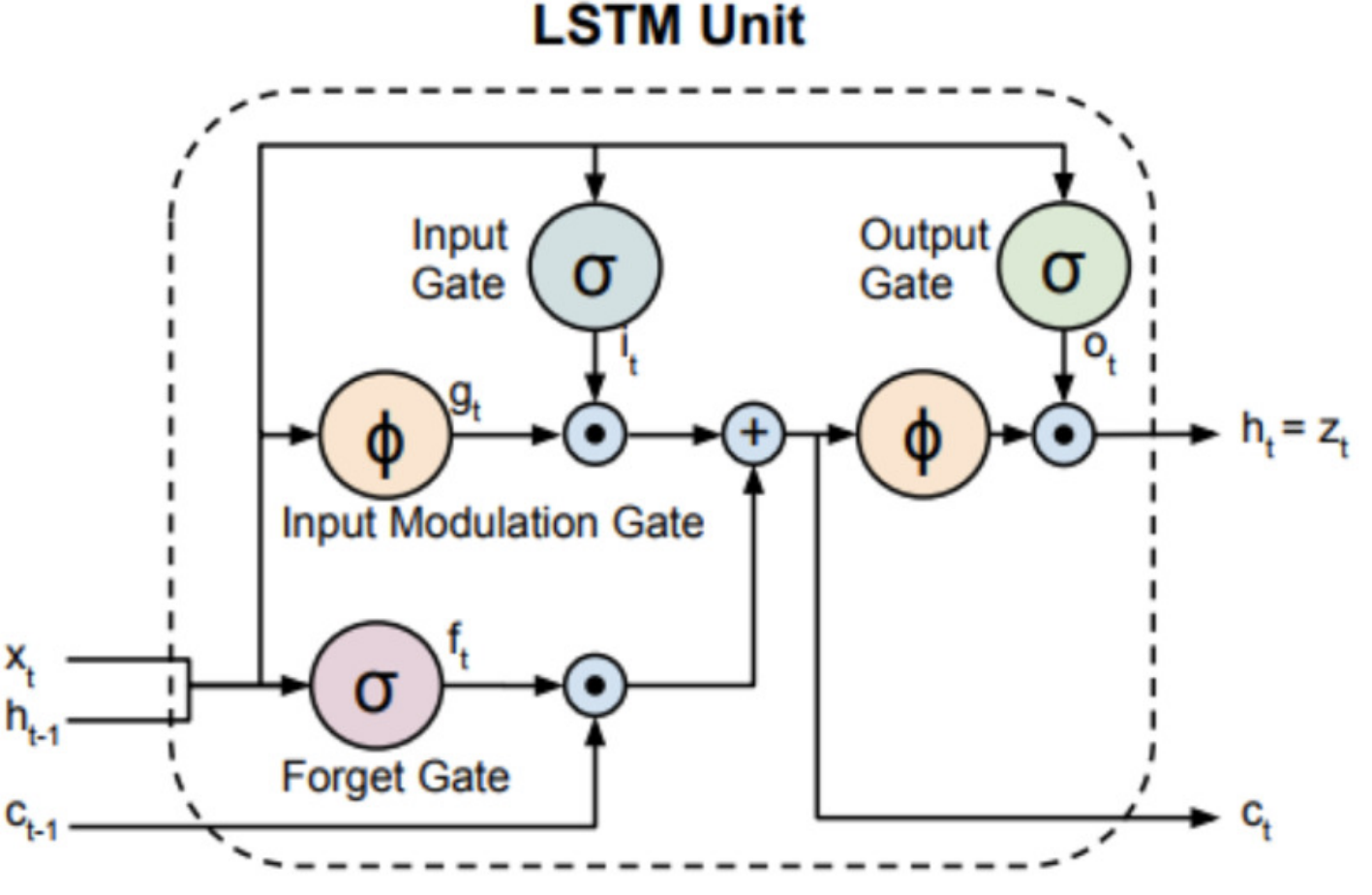}
\par\end{centering}
\caption[Illustration of a typical LSTM unit.]{\label{fig:lstm}Illustration of a typical LSTM unit. The unit takes
input at the current time, $x_{t}$, and from a previous time, $h_{t-1}$,
and it returns an output to be fed into the next time, $h_{t}$. The
final output of the LSTM unit is controlled by the input gate, $i_{t}$,
the forget gate, $f_{t}$, and the output gate, $o_{t}$, as well
as the memory cell state, $c_{t}$, which are defined in \eqref{eq:input_gate},
\eqref{eq:forget_gate}, \eqref{eq:output_gate1} and \eqref{eq:cell_state},
respectively. Figure reproduced from \cite{Donahue17}.}
\end{figure}

LSTMs are recurrent networks that are further equipped with a storage
or memory component, illustrated in Figure \ref{fig:lstm}, that accumulates
information over time. An LSTM's memory cell is gated such that it
allows information to be read from it or written to it. Notably, LSTMs
also contain a forget gate that allows the network to erase information
when it is not needed anymore. LSTMs are controlled by three different
gates (the input gate, $i_{t}$, the forget gate, $f_{t}$, and the
output gate, $o_{t}$), as well as the memory cell state, $c_{t}$.
The input gate is controlled by the current input, $x_{t}$, and the
previous state, $h_{t-1}$, and it is defined as 

\begin{equation}
i_{t}=\sigma(w_{i}x_{t}+u_{i}h_{t-1}+b_{i}),\label{eq:input_gate}
\end{equation}
where, $w_{i}$, $u_{i}$, $b_{i}$ represent the weights and bias
controlling the connections to the input gate and $\sigma$ is usually
a sigmoid function. The forget gate is similarly defined as 
\begin{equation}
f_{t}=\sigma(w_{f}x_{t}+u_{f}h_{t-1}+b_{f}),\label{eq:forget_gate}
\end{equation}
and it is controlled by its corresponding weights and bias, $w_{f}$,
$u_{f}$, $b_{f}$. Arguably, the most important aspect of an LSTM
is that it copes with the challenge of vanishing and exploding gradients.
This ability is achieved through additive combination of the forget
and input gate states in determining the memory cell's state, which,
in turn, controls whether information is passed on to another cell
via the output gate. Specifically, the cell state is computed in two
steps. First, a candidate cell state is estimated according to

\begin{equation}
g_{t}=\mathbf{\phi}(w_{c}x_{t}+u_{c}h_{t-1}+b_{c}),\label{eq:estimated_state}
\end{equation}
where $\phi$ is usually a hyperbolic tangent. Second, the final cell
state is finally controlled by the current estimated cell state, $g_{t}$,
and the previous cell state, $c_{t-1}$, modulated by the input and
forget gate according to

\begin{equation}
c_{t}=i_{t}g_{t}+f_{t}c_{t-1}.\label{eq:cell_state}
\end{equation}
Finally, using the cell's state and the current and previous inputs,
the value of the output gate and the output of the LSTM cell are estimated
according to

\begin{equation}
o_{t}=\sigma(w_{o}x_{t}+u_{o}h_{t-1}+b_{o}),\label{eq:output_gate1}
\end{equation}
where

\begin{equation}
h_{t}=\phi(c_{t})\:o_{t}.\label{eq:lstm_output}
\end{equation}

\subsection{Convolutional networks\label{subsec:Convolutional-Networks}}

\begin{figure}[H]
\begin{centering}
\includegraphics[width=0.8\textwidth]{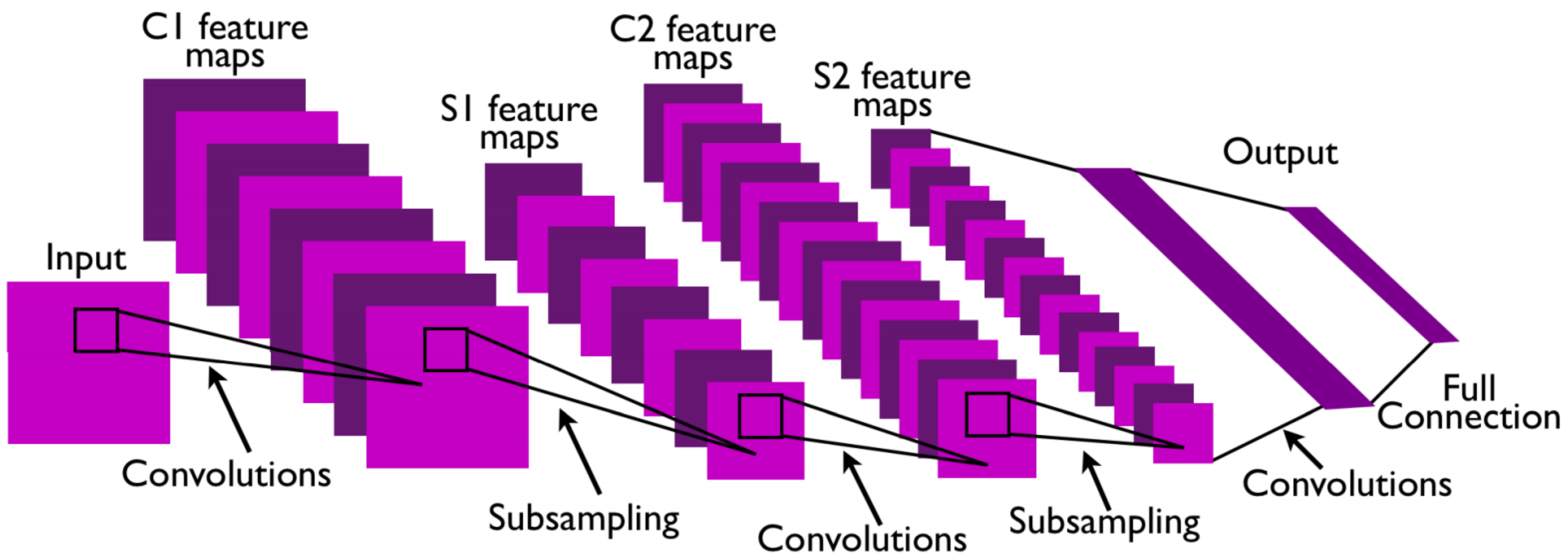}
\par\end{centering}
\caption[Illustration of the structure of a standard Convolutional Network.]{\label{fig:cnn}Illustration of the structure of a standard Convolutional
Network. Figure reproduced from \cite{LeCun2010}.}
\end{figure}

Convolutional networks (ConvNets) are a special type of neural network
that are especially well adapted to computer vision applications because
of their ability to hierarchically abstract representations with local
operations. There are two key design ideas driving the success of
convolutional architectures in computer vision. First, ConvNets take
advantage of the 2D structure of images and the fact that pixels within
a neighborhood are usually highly correlated. Therefore, ConvNets
eschew the use of one-to-one connections between all pixel units (\ie
as is the case of most neural networks) in favor of using grouped
local connections. Further, ConvNet architectures rely on feature
sharing and each channel (or output feature map) is thereby generated
from convolution with the same filter at all locations as depicted
in Figure \ref{fig:cnn}. This important characteristic of ConvNets
leads to an architecture that relies on far fewer parameters compared
to standard Neural Networks. Second, ConvNets also introduce a pooling
step that provides a degree of translation invariance making the architecture
less affected by small variations in position. Notably, pooling also
allows the network to gradually see larger portions of the input thanks
to an increased size of the network's receptive field. The increase
in receptive field size (coupled with a decrease in the input's resolution)
allows the network to represent more abstract characteristics of the
input as the network's depth increase. For example, for the task of
object recognition, it is advocated that ConvNets layers start by
focusing on edges to parts of the object to finally cover the entire
object at higher layers in the hierarchy.

\begin{figure}[H]
\begin{centering}
\includegraphics[width=0.65\textwidth]{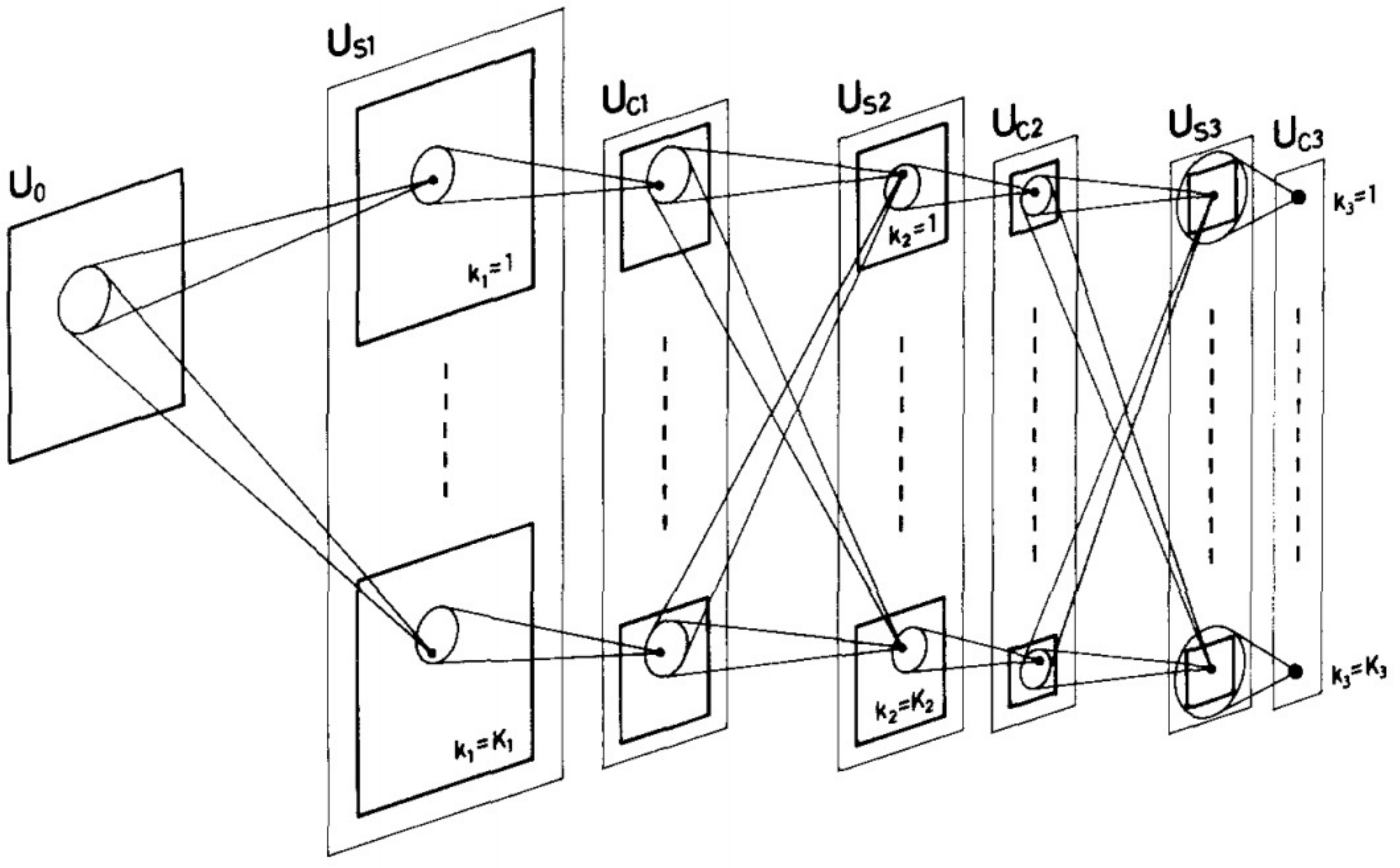}
\par\end{centering}
\caption[Illustration of the structure of the Neocognitron.]{\label{fig:neocognitron}Illustration of the structure of the Neocognitron.
Figure reproduced from \cite{Fukushima1980}.}
\end{figure}

The architecture of convolutional networks is heavily inspired by
the processing that takes place in the visual cortex as described
in the seminal work of Hubel and Wiesel \cite{hubel1962} (further
discussed in Chapter 3). In fact, it appears that the earliest instantiation
of Convolutional Networks is Fukushima's Neocognitron \cite{Fukushima1980},
which also relied on local connections and in which each feature map
responds maximally to only a specific feature type. The Neocognitron
is composed of a cascade of $K$ layers where each layer alternates
S-cell units, $U_{sl}$, and complex cell units, $U_{cl}$, that loosely
mimic the processing that takes place in the biological simple and
complex cells, respectively, as depicted in Figure \ref{fig:neocognitron}.
The simple cell units perform operations similar to local convolutions
followed by a Rectified Linear Unit (ReLU) nonlinearity, $\varphi(x)=\left\{ \begin{array}{cc}
x; & if\,x\geq0\\
0; & x<0
\end{array}\right.$,while the complex cells perform operations similar to average pooling.
The model also included a divisive nonlinearity to accomplish something
akin to normalization in contemporary ConvNets.

As opposed to most standard ConvNet architectures (\eg \cite{LeCun1998,Krizhevsky2012})
the Neocognitron does not need labeled data for learning as it is
designed based on self organizing maps that learn the local connections
between consecutive layers via repetitive presentations of a set of
stimulus images. In particular, the Neocognitron is trained to learn
the connections between an input feature map and a simple cell layer
(connections between a simple cells layer and complex cells layer
are pre-fixed) and the learning procedure can be broadly summarized
in two steps. First, each time a new stimulus is presented at the
the input, the simple cells that respond to it maximally are chosen
as a representative cell for that stimulus type. Second, the connections
between the input and those representative cells are reinforced each
time they respond to the same input type. Notably, simple cells layers
are organized in different groups or planes such that each plane responds
only to one stimulus type (\ie similar to feature maps in a modern
ConvNet architecture). Subsequent extensions to the Neocognitron included
allowances for supervised learning \cite{Fukushima1991} as well as
top-down attentional mechanisms \cite{Fukushima87}.

Most ConvNets architectures deployed in recent computer vision applications
are inspired by the successful architecture proposed by LeCun in 1998,
now known as LeNet, for handwriting recognition \cite{LeCun1998}.
As described in key literature \cite{Jarret2009,LeCun2010}, a classical
convolutional network is made of four basic layers of processing:
(i) a convolution layer, (ii) a nonlinearity or rectification layer,
(iii) a normalization layer and (iv) a pooling layer. As noted above,
these components were largely present in the Neocognitron. A key addition
in LeNet was the incorporation of back propagation for relatively
efficient learning of the convolutional parameters.

Although, ConvNets allow for an optimized architecture that requires
far fewer parameters compared to their fully connected neural network
counterpart, their main shortcoming remains their heavy reliance on
learning and labeled data. This data dependence is probably one of
the main reasons why ConvNets were not widely used until 2012 when
the availability of the large ImageNet dataset \cite{ImageNet} and
concomitant computational resources made it possible to revive interest
in ConvNets \cite{Krizhevsky2012}. The success of ConvNets on ImageNet
led to a spurt of various ConvNet architectures and most contributions
in this field are merely based on different variations of the basic
building blocks of ConvNets, as will be discussed later in Section
\ref{sec:Spatial-Convolutional-Networks}. 

\subsection{Generative adversarial networks}

Generative Adversarial Networks (GANs) are relatively new models taking
advantage of the strong representational power of multilayer architectures.
GANs were first introduced in 2014 \cite{Goodfellow2014} and although
they did not present a different architecture per se (\ie in terms
of novel network building blocks for example), they entail some peculiarities,
which make them a slightly different class of multilayer architectures.
A key challenge being responded to by GANs is the introduction of
an unsupervised learning approach that requires no labeled data.

\begin{figure}
\begin{centering}
\includegraphics[width=0.5\textwidth]{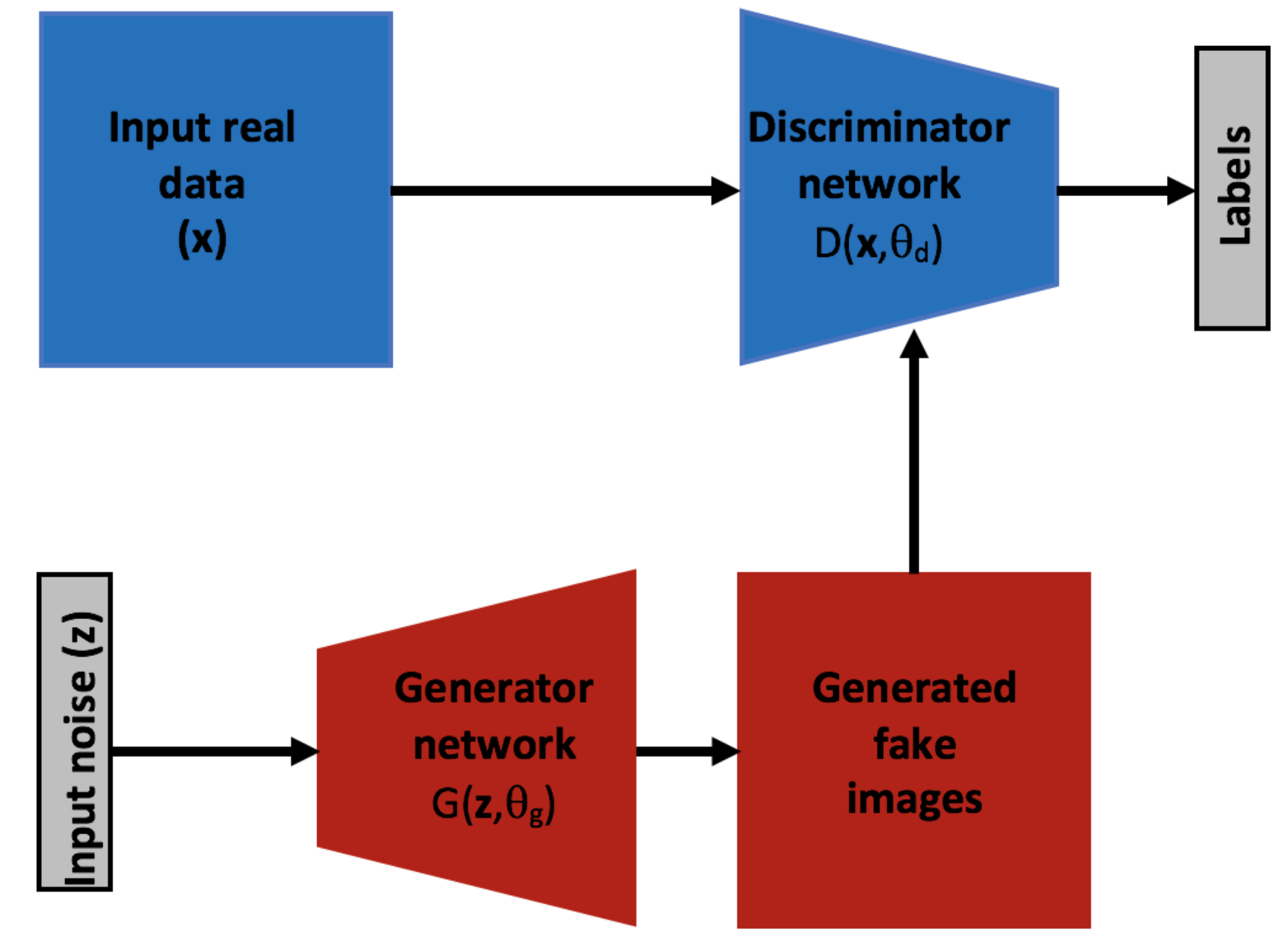}
\par\end{centering}
\caption[Illustration of the structure of a general purpose Generative Adverserial
Network (GAN).]{\label{fig:GAN}Illustration of the structure of a general purpose
Generative Adverserial Network (GAN).}
\end{figure}

A typical GAN is made of two competing blocks or sub-networks, as
shown in Figure \ref{fig:GAN}; a generator network, $G(\mathbf{z};\theta_{g})$,
and a discriminator network, $D(\mathbf{x};\theta_{d})$, where $\mathbf{z}$
is input random noise, $\mathbf{x}$ is real input data (\eg an image)
and $\theta_{g}$ and $\theta_{d}$ are the parameters of the two
blocks, respectively. Each block can be made of any of the previously
defined multilayer architectures. In the original paper both the generator
and discriminator were multilayer fully connected networks. The discriminator,
$D$, is trained to recognize the data coming from the generator and
assigning the label \textit{``fake''} with probability $p_{d}$
while assigning the label \textit{``real''} to true input data with
probability $1-p_{d}$. In complement, the generator network is optimized
to generate fake representations capable of fooling the discriminator.
The two blocks are trained alternately in several steps where the
ideal outcome of the training process is a discriminator that assigns
a probability of $50\%$ to both real and fake data. In other words,
after convergence the generator should be able to generate realistic
data from random input. 

Since the original paper, many contributions participated in enhancing
the capabilities of GANs via use of more powerful multilayer architectures
as the backbones of the network \cite{Radford2015} (\eg pretrained
convolutional networks for the discriminator and deconvolutional networks,
that learn upsampling filters for the generator). Some of the successful
applications of GANs include: text to image synthesis (where the input
to the network is a textual description of the image to be rendered
\cite{Reed2016}), image super resolution where the GAN generates
a realistic high resolution image from a lower resolution input \cite{Ledig2016a},
image inpainting where the role of GANs is to fill holes of missing
information from an input image \cite{yeh2017semantic} and texture
synthesis where GANs are used to synthesize realistic textures from
input noise \cite{bergmann17a}.

\subsection{Multilayer network training}

As discussed in the previous sections, the success of the various
multilayer architectures largely depends on the success of their learning
process. While neural networks usually rely on an unsupervised pretraining
step first, as described in Section \ref{subsec:Neural-Networks},
they are usually followed by the most widely used training strategy
for multilayer architectures, which is fully supervised. The training
procedure is usually based on error back propagation using gradient
descent. Gradient descent is widely used in training multilayer architectures
for its simplicity. It relies on minimizing a smooth error function,
$E(\boldsymbol{w})$, following an iterative procedure defined as

\begin{equation}
\mathbf{w}_{k}=\mathbf{w}_{k-1}-\alpha\frac{\partial E(\mathbf{w})}{\partial\mathbf{w}},\label{eq:backprop}
\end{equation}
where $\mathbf{\mathbf{w}}$ represents the network's parameters,
$\alpha$ is the learning rate that may control the speed of convergence
and $\mathbf{\frac{\partial E(\mathbf{w})}{\partial\mathbf{w}}}$
is the error gradient calculated over the training set. This simple
gradient descent method is especially suitable for training multilayer
networks thanks to the use of the chain rule for back propagating
and calculating the error derivative with respect to various network's
parameters at different layers. While back propagation dates back
a number of years \cite{BRYSON1963,Werbos74}, it was popularized
in the context of multilayer architectures \cite{Rumelhart1986}.
In practice, stochastic gradient descent is used \cite{Amari1993},
which consists of approximating the error gradient over the entire
training set from successive relatively small subsets. 

One of the main problems of the gradient descent algorithm is the
choice of the learning rate, $\alpha$. A learning rate that is too
small leads to slow convergence, while a large learning rate can lead
to overshooting or fluctuation around the optimum. Therefore, several
approaches were proposed to further improve the simple stochastic
gradient descent optimization method. The simplest method, referred
to as stochastic gradient descent with momentum \cite{sutton1986},
keeps track of the update amount from one iteration to another and
gives momentum to the learning process by pushing the update further
if the gradient keeps pointing to the same direction from one time
step to another as defined in, 

\begin{equation}
\mathbf{w}_{k}=\mathbf{w}_{k-1}-\alpha\frac{\partial E(\mathbf{w})}{\partial\mathbf{w}}-\gamma(\frac{\partial E(\mathbf{w})}{\partial\mathbf{w}})_{t-1},\label{eq:sgd_momentum}
\end{equation}
with $\gamma$ controlling the momentum. Another simple method involves
setting the learning rate in a decreasing fashion according to a fixed
schedule, but this is far from ideal given that this schedule has
to be pre-set ahead of the training process and is completely independent
from the data. Other more involved methods (\eg Adagrad \cite{Duchi2011},
Adadelta \cite{zeiler2012}, Adam \cite{kingma2015}) suggest adapting
the learning rate during training to each parameter, $w_{i}$, being
updated, by performing smaller updates on frequently changing parameters
and larger updates on infrequent ones. A detailed comparison between
the different versions of these algorithms can be found elsewhere
\cite{Ruder2016}.

The major shortcoming of training using gradient descent, as well
as its variants, is the need for large amounts of labeled data. One
way to deal with this difficulty is to resort to unsupervised learning.
A popular unsupervised method used in training some shallow ConvNet
architectures is based on the Predictive Sparse Decomposition (PSD)
method \cite{koray-psd-08}. Predictive Sparse Decomposition learns
an overcomplete set of filters whose combination can be used to reconstruct
an image. This method is especially suitable for learning the parameters
of a convolutional architecture, as the algorithm is designed to learn
basis functions that reconstruct an image patchwise. Specifically,
Predictive Sparse Decomposition (PSD) builds on sparse coding algorithms
that attempts to find an efficient representation, Y, of an input
signal, X, via a linear combination with a basis set, B. Formally,
the problem of sparse coding is broadly formulated as a minimization
problem defined as,

\begin{equation}
L(X,Y;B)=||X-BY||_{2}^{2}.\label{eq:SD}
\end{equation}

PSD adapts the idea of sparse coding in a convolutional framework
by minimizing a reconstruction error defined as,

\begin{equation}
L(X,Y;B)=||X-BY||_{2}^{2}+\lambda||Y||_{1}+\alpha||Y-F(X;G,W,D)||_{2}^{2}\label{eq:PSD}
\end{equation}
where $F(X;G,W,D)=G\:\tanh(WX+D)$ and $W$, $D$ and $G$ are weights,
biases and gains (or normalization factors ) of the network, respectively.
By minimizing the loss function defined in equation \ref{eq:PSD},
the algorithm learns a representation, $Y$, that reconstructs the
input patch, $X$, while being similar to the predicted representation
$F$. The learned representation will also be sparse owing to the
second term of the equation. In practice, the error is minimized in
two alternating steps where parameters, $(B,G,W,D)$, are fixed and
minimization is performed over $Y$. Then, the representation $Y$
is fixed while minimizing over the other parameters. Notably, PSD
is applied in a patchwise procedure where each set of parameters,
$(G,W,D)$, is learned from the reconstruction of a different patch
from an input image. In other words, a different set of kernels is
learned by focusing the reconstruction on different parts of the input
images. 

\subsection{A word on transfer learning}

One of the unexpected benefits of training multilayer architecture
is the surprising adaptability of the learned features across different
datasets and even different tasks. Examples include using networks
trained with ImageNet for recognition on: other object recognition
datasets such as Caltech-101 \cite{FeiFeiFergusPeronaPAMI} (\eg\cite{minlin14,Zeiler2014}),
other recognitions tasks such as texture recognition (\eg\cite{cimpoi14}),
other applications such as object detection (\eg\cite{girshick2014rcnn})
and even to video based tasks, such as video action recognition (\eg\cite{SimonyanTwoStream,Chris2017ACTIONS,liminwang2016}).

The adaptability of features extracted with multilayer architectures
across different datasets and tasks, can be attributed to their hierarchical
nature where the representations progress from being simple and local
to abstract and global. Thus, features extracted at lower levels of
the hierarchy tend to be common across different tasks thereby making
multilayer architectures more amenable to transfer learning. 

A systematic exploration of the intriguing transferability of features
across different networks and tasks revealed several good practices
to take into account in consideration of transfer learning \cite{Yosinski2014}.
First, it was shown that fine tuning higher layers only, led to systematically
better performance when compared to fine tuning the entire network.
Second, this research demonstrated that the more different the tasks
are the less efficient transfer learning becomes. Third, and more
surprisingly, it was found that even after fine tuning the network's
performance under the initial task is not particularly hampered.

Recently, several emerging efforts attempt to enforce a networks'
transfer learning capabilities even further by casting the learning
problem as a sequential two step procedure, \eg\cite{Andrychowicz2016,santoro2016}.
First, a so called rapid learning step is performed where a network
is optimized for a specific task as is usually done. Second, the network
parameters are further updated in a global learning step that attempts
to minimize an error across different tasks.

\section{Spatial convolutional networks\label{sec:Spatial-Convolutional-Networks}}

In theory, convolutional networks can be applied to data of arbitrary
dimensions. Their two dimensional instantiations are well suited to
the structure of single images and therefore have received considerable
attention in computer vision. With the availability of large scale
datasets and powerful computers for training, the vision community
has recently seen a surge in the use of ConvNets for various applications.
This section describes the most prominent 2D ConvNet architectures
that introduced relatively novel components to the original LeNet
described in Section \ref{subsec:Convolutional-Networks}.

\subsection{Key architectures in the recent evolution of ConvNets}

\begin{figure}[H]
\begin{centering}
\includegraphics[width=0.85\textwidth]{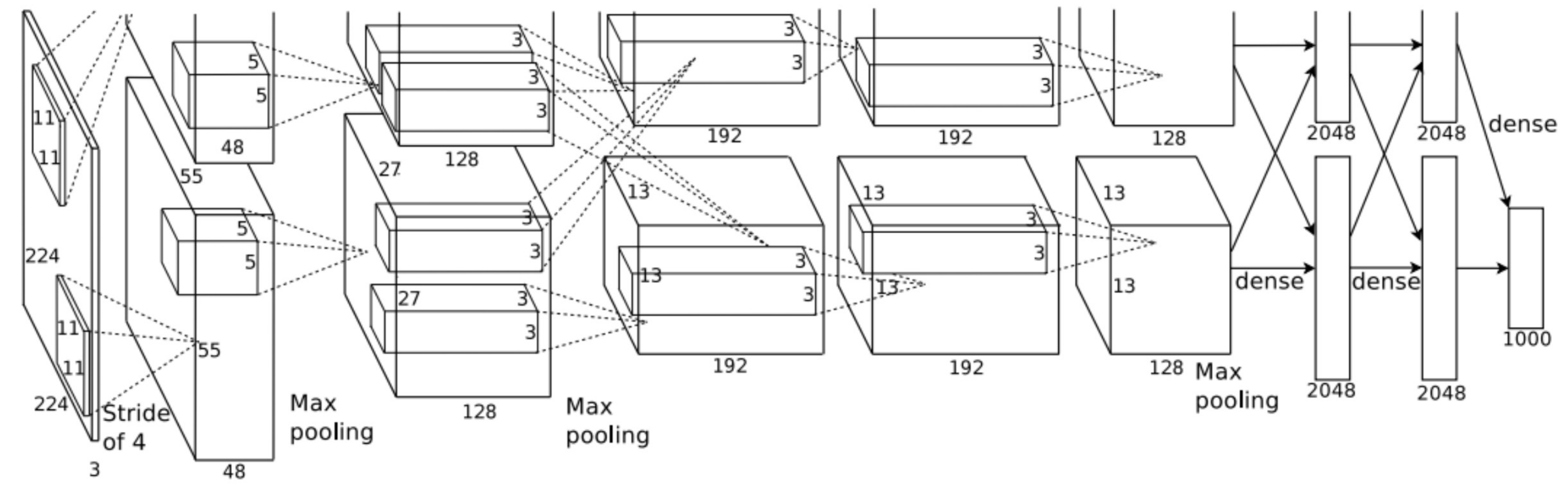}
\par\end{centering}
\caption[AlexNet architecture.]{\label{fig:alexnet}AlexNet architecture. Notably, although the depiction
suggests a two stream architecture, it was in fact a single stream
architecture and this depiction only reflects the fact that AlexNet
was trained in parallel on 2 different GPUs. Figure reproduced from
\cite{Krizhevsky2012}.}
\end{figure}

The work that rekindled interest in ConvNet architectures was Krishevsky's
AlexNet \cite{Krizhevsky2012}. AlexNet was able to achieve record
breaking object recognition results on the ImageNet dataset. It consisted
of eight layers in total, 5 convolutional and 3 fully connected, as
depicted in Figure \ref{fig:alexnet}. 

AlexNet introduced several architectural design decisions that allowed
for efficient training of the network using standard stochastic gradient
descent. In particular, four important contributions were key to the
success of AlexNet. First, AlexNet considered the use of the ReLU
nonlinearity instead of the saturating nonlinearites, such as sigmoids,
that were used in previous state-of-the-art ConvNet architectures
(\eg LeNet \cite{LeCun1998}). The use of the ReLU diminished the
problem of vanishing gradient and led to faster training. Second,
noting the fact that the last fully connected layers in a network
contain the largest number of parameters, AlexNet used dropout, first
introduced in the context of neural networks \cite{Srivastava2014dropout},
to reduce the problem of overfitting. Dropout, as implemented in AlexNet,
consists in randomly dropping (\ie setting to zero) a given percentage
of a layer's parameters. This technique allows for training a slightly
different architecture at each pass and artificially reducing the
number of parameters to be learned at each pass, which ultimately
helps break correlations between units and thereby combats overfitting.
Third, AlexNet relied on data augmentation to improve the network's
ability to learn invariant representations. For example, the network
was trained not only on the original images in the training set, but
also on variations generated by randomly shifting and reflecting the
training images. Finally, AlexNet also relied on several techniques
to make the training process converge faster, such as the use momentum
and a scheduled learning rate decrease whereby the learning rate is
decreased every time the learning stagnates.

The advent of AlexNet led to a spurt in the number of papers trying
to understand what the network is learning either via visualization,
as done in the so called DeConvNet \cite{Zeiler2014}, or via systematic
explorations of various architectures \cite{Chatfield11,chatfield2014}.
One of the direct results of these explorations was the realization
that deeper networks can achieve even better results as first demonstrated
in the 19 layer deep VGG-Net \cite{Simonyan14c}. VGG-Net achieves
its depth by simply stacking more layers while following the standard
practices introduced with AlexNet (\eg reliance on the ReLU nonlinearity
and data augmentation techniques for better training). The main novelty
presented in VGG-Net was the use of filters with smaller spatial extent
(\ie $3\times3$ filters throughout the network instead of \eg $11\times11$
filters used in AlexNet), which allowed for an increase in depth without
dramatically increasing the number of parameters that the network
needs to learn. Notably, while using smaller filters, VGG-Net required
far more filters per layer. 

\begin{figure}[H]
\begin{centering}
\begin{tabular}{c}
\includegraphics[width=0.5\textwidth]{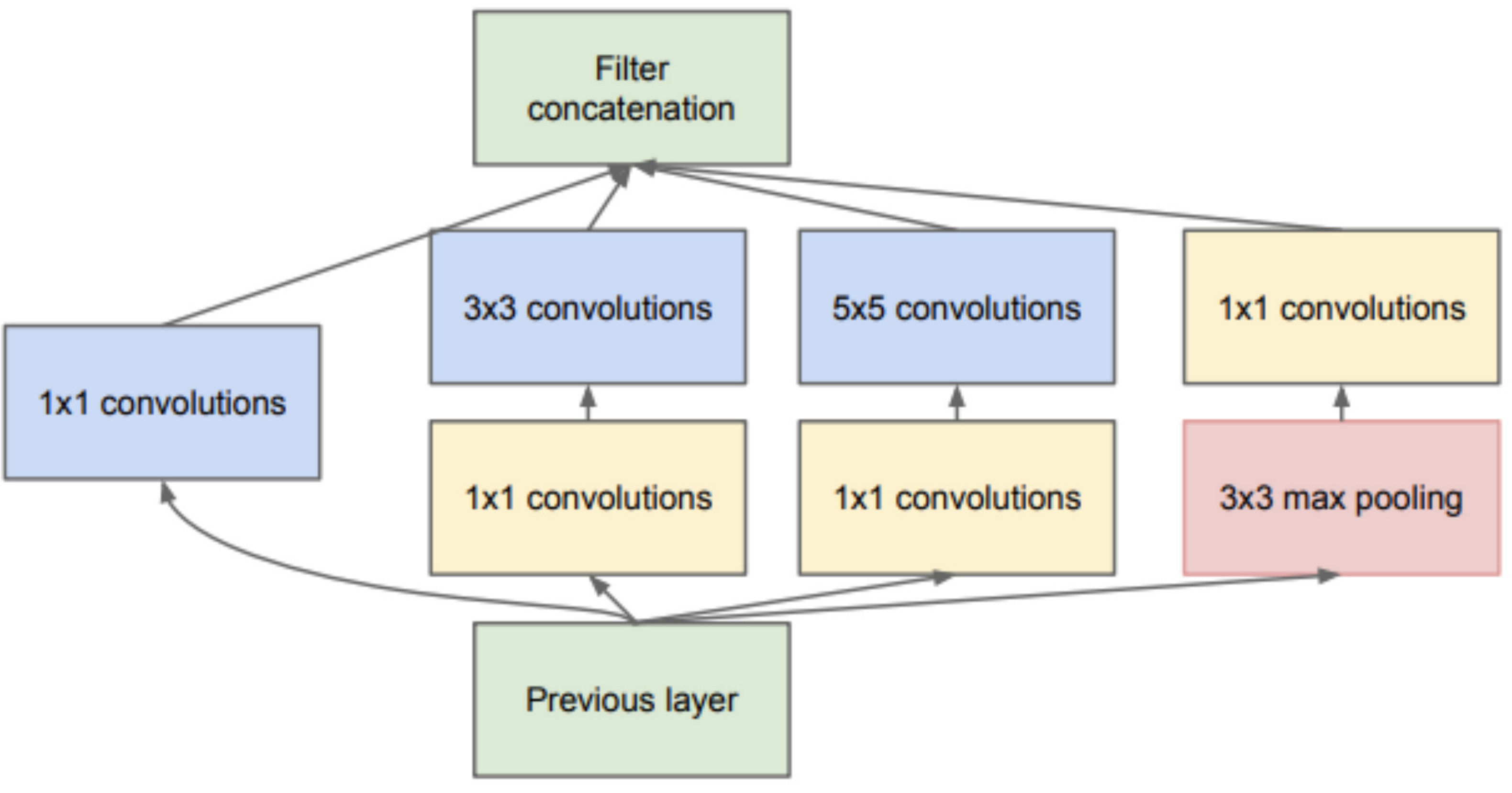}\tabularnewline
(a)\tabularnewline
\includegraphics[width=0.95\textwidth]{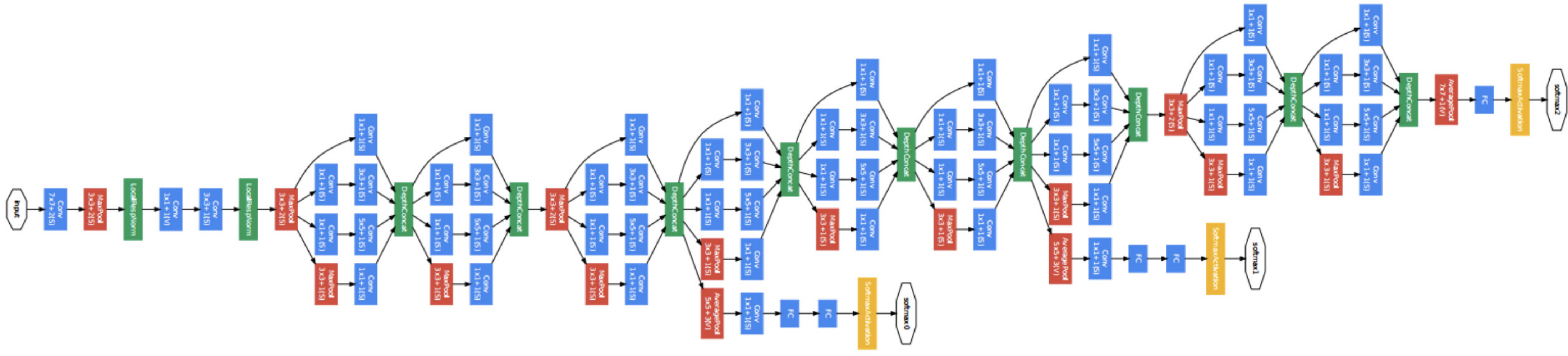}\tabularnewline
(b)\tabularnewline
\end{tabular}
\par\end{centering}
\caption[GoogLeNet architecture.]{\label{fig:inception}GoogLeNet architecture. (a) A typical inception
module showing operations that happen sequentially and in parallel.
(b) Illustration of a typical ``inception'' architecture that consists
of stacking many inception modules. Figure reproduced from \cite{Szegedy_2015_CVPR}}

\end{figure}

VGG-Net was the first and simplest of many deep ConvNet architectures
that followed AlexNet. A deeper architecture, commonly known as GoogLeNet,
with 22 layers was proposed later \cite{Szegedy_2015_CVPR}. While
being deeper than VGG-Net, GoogLeNet requires far fewer parameters
thanks to the use of the so called inception module, shown in Figure
\ref{fig:inception}(a), as a building block. In an inception module
convolution operations at various scales and spatial pooling happen
in parallel. The module is also augmented with $1\times1$ convolutions
(\ie cross-channel pooling) that serve the purpose of dimensionality
reduction to avoid or attenuate redundant filters, while keeping the
network's size manageable. This cross-channel pooling idea was motivated
by the findings of a previous work known as the Network in Network
(NiN) \cite{minlin14}, which disucssed the large redundancies in
the learned networks. Stacking many inception modules led to the now
widely used GoogLeNet architecture depicted in Figure \ref{fig:inception}(b). 

\begin{figure}[H]
\begin{centering}
\begin{tabular}{c}
\includegraphics[width=0.5\textwidth]{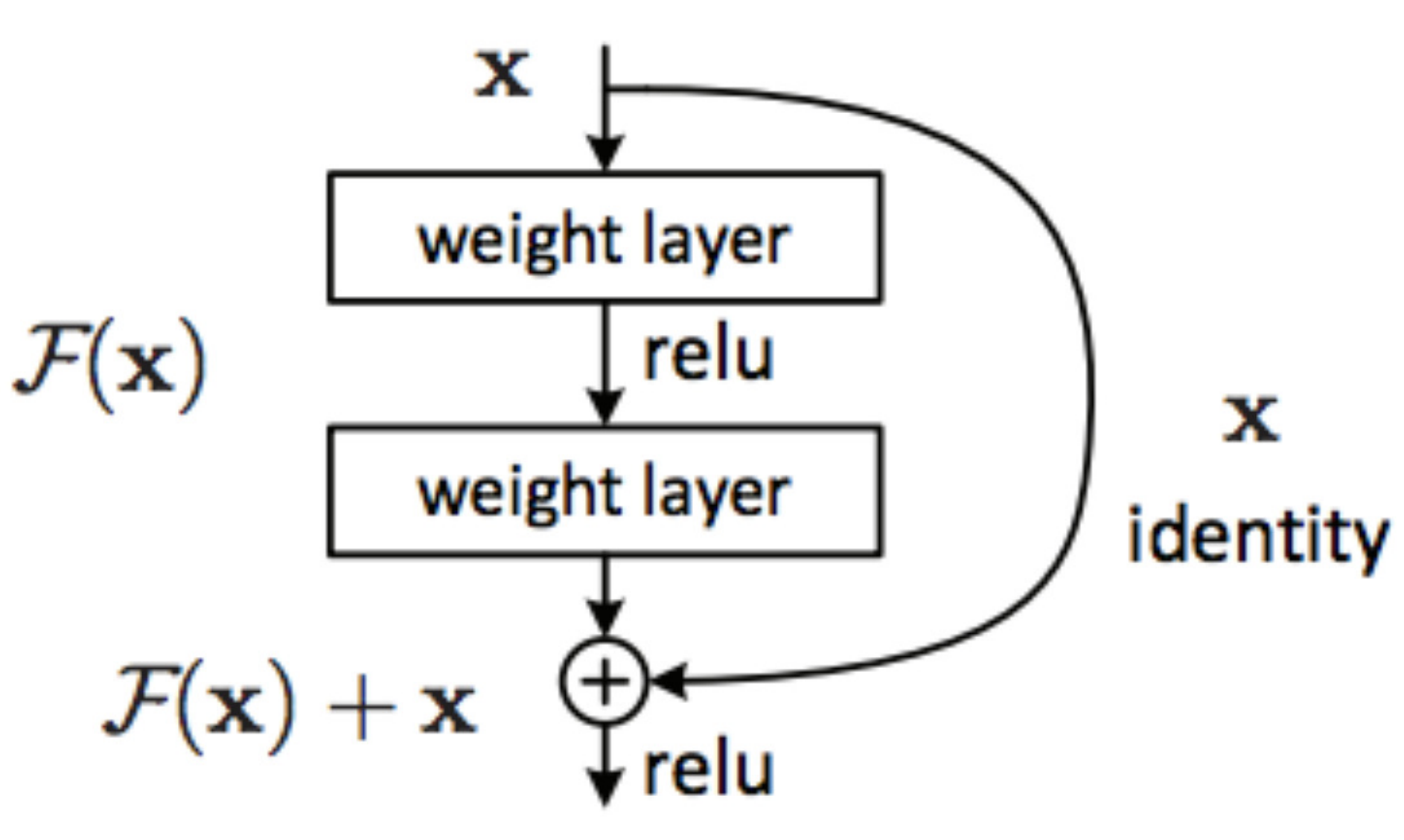}\tabularnewline
(a)\tabularnewline
\includegraphics[width=0.95\textwidth]{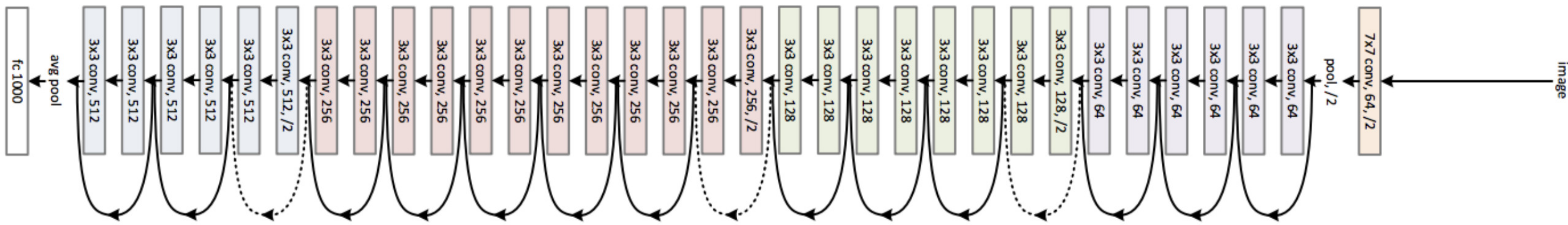}\tabularnewline
(b)\tabularnewline
\end{tabular}
\par\end{centering}
\caption[ResNet architecture.]{\label{fig:resnet}ResNet architecture. (a) A residual module. (b)
Illustration of a typical ResNet architecture that consists of stacking
many residual modules. Figure reproduced from \cite{He2016}.}
\end{figure}

GoogLeNet was the first network to stray away from the strategy of
simply stacking convolutional and pooling layers and it was soon followed
by one of the deepest architectures to date, known as ResNet \cite{He2016},
that also proposed a novel architecture with over 150 layers. ResNet
stands for Residual Network where the main contribution lies in its
reliance on residual learning. In particular, ResNet is built such
that each layer learns an incremental transformation, $F(x)$, on
top of the input, $x$, according to 

\begin{equation}
H(x)=F(x)+x,\label{resnet}
\end{equation}
instead of learning the transformation $H(x)$ directly as done in
other standard ConvNet architectures. This residual learning is achieved
via use of skip connections, illustrated in Figure \ref{fig:resnet}(a),
that connect components of different layers with an identity mapping.
Direct propagation of the signal, $x$, combats the vanishing gradient
problem during back propagation and thereby enables the training of
very deep architectures.

\begin{figure}[H]
\begin{centering}
\begin{tabular}{c}
\includegraphics[width=0.5\textwidth]{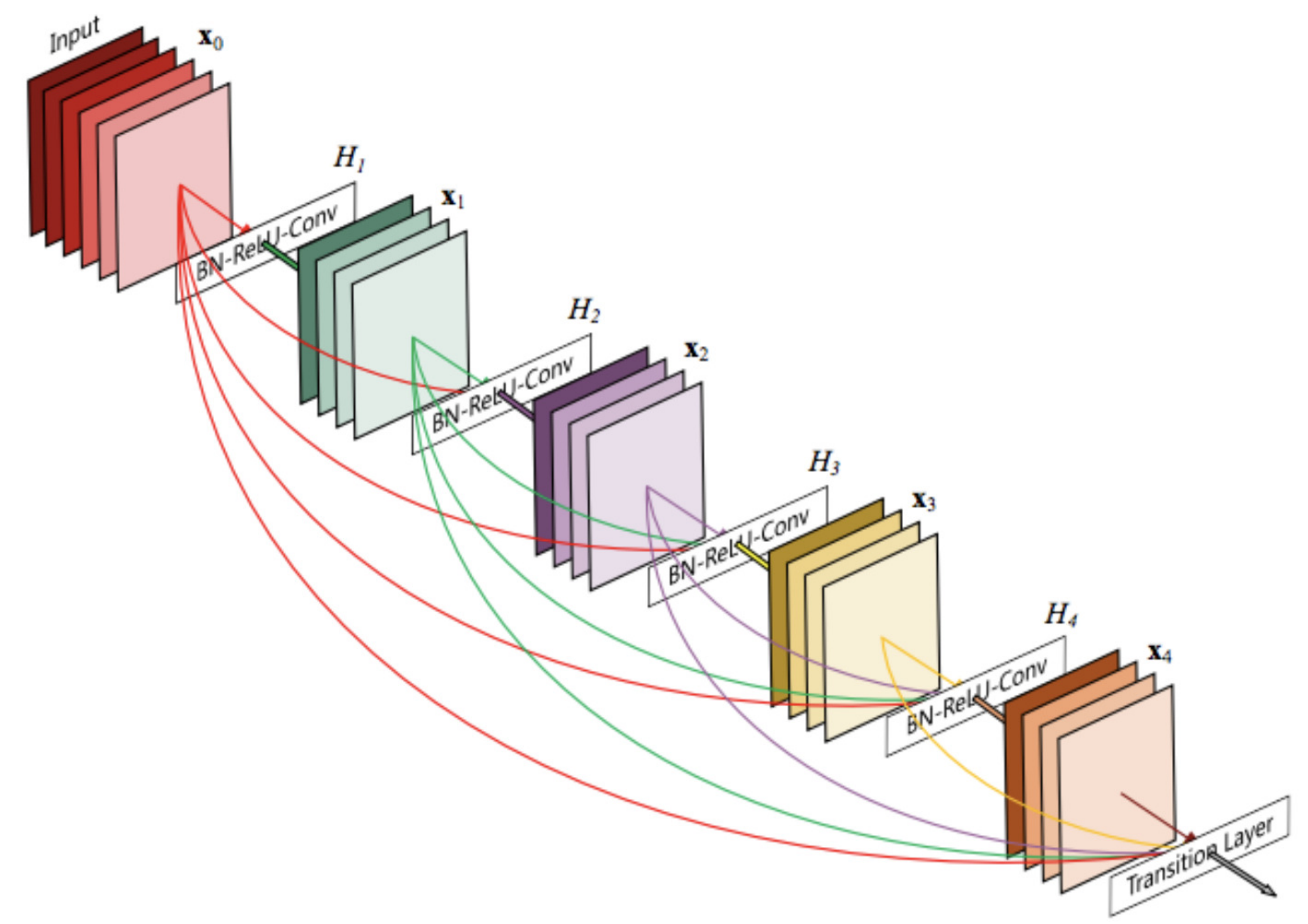}\tabularnewline
(a)\tabularnewline
\includegraphics[width=0.95\textwidth]{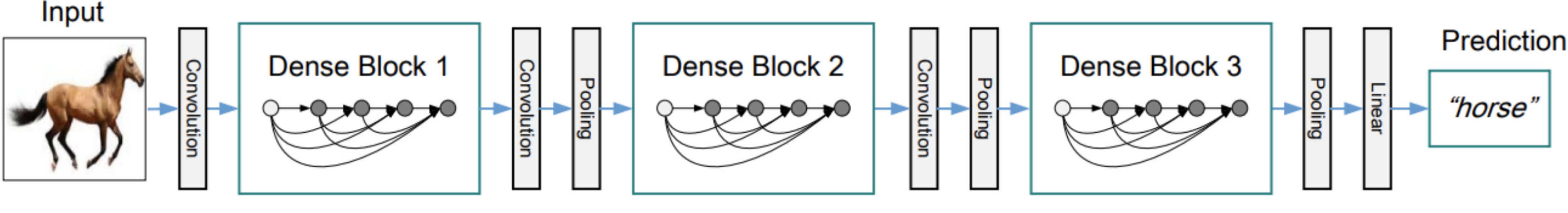}\tabularnewline
(b)\tabularnewline
\end{tabular}
\par\end{centering}
\caption[DenseNet architecture.]{\label{fig:densenet}DenseNet architecture. (a) A dense module. (b)
Illustration of a typical DenseNet architecture that consists of stacking
many dense modules. Figure reproduced from \cite{HuangLMW17}.}
\end{figure}

A recent, closely related network building on the success of ResNet
is the so called DenseNet \cite{HuangLMW17}, which pushes the idea
of residual connections even further. In DenseNet, every layer is
connected, via skip connections, to all subsequent layers of a dense
block as illustrated in Figure \ref{fig:densenet}. Specifically,
a dense block connects all layers with feature maps of the same size
(\ie blocks between spatial pooling layers). Different from ResNet,
DenseNet does not add feature maps from a previous layer, \eqref{resnet},
but instead concatenates features maps such that the network learns
a new representation according to

\begin{equation}
H(x_{l})=F(x_{l-1},...,x_{1},x_{0}).
\end{equation}

The authors claim that this strategy allows DenseNet to use fewer
filters at each layer since possible redundant information is avoided
by pushing features extracted at one layer to other layers higher
up in the hierarchy. Importantly, these deep skip connections allow
for better gradient flow given that lower layers have more direct
access to the loss function. Using this simple idea allowed DenseNet
to compete with other deep architectures, such as ResNet, while requiring
fewer parameters and incurring less overfitting.

\subsection{Toward ConvNet invariance}

One of the challenges of using ConvNets is the requirement of very
large datasets to learn all the underlying parameters. Even large
scale datasets such as ImageNet \cite{ImageNet}, with over a million
images, is considered too small for training certain deep architectures.
One way to cope with the large dataset requirement is to artificially
augment the dataset by altering the images via random flipping, rotation
and jittering, for example. The major advantage of these augmentations
is that the resulting networks become more invariant to various transformations.
In fact, this technique was one of the main reasons behind the large
success of AlexNet. Therefore, beyond methods altering the network's
architecture for easier training, as discussed in the previous section,
other work aims at introducing novel building blocks that yield better
training. Specifically, networks discussed under this section introduce
novel blocks that incorporate learning invariant representation directly
from the raw data. 

\begin{figure}[H]
\begin{centering}
\begin{tabular}{cc}
\includegraphics[width=0.62\textwidth]{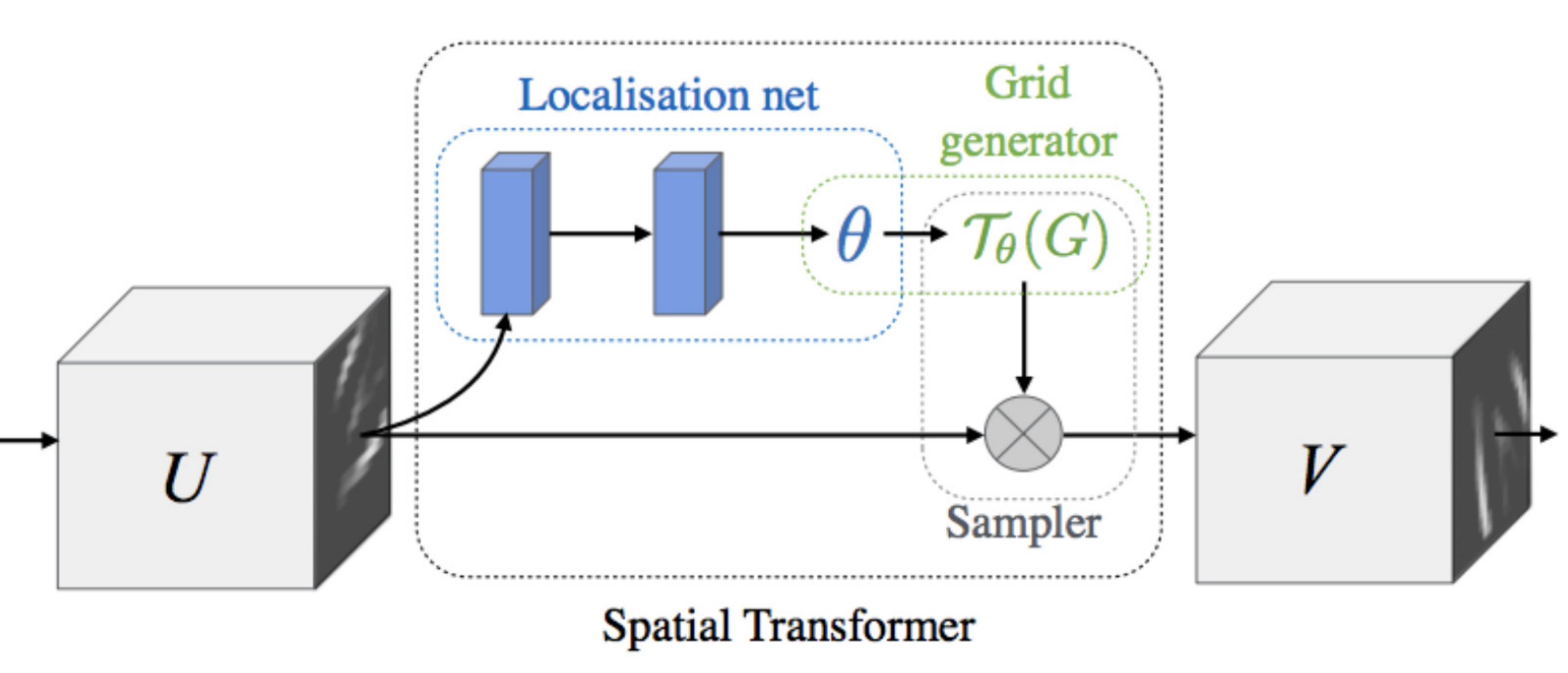} & \includegraphics[width=0.32\textwidth]{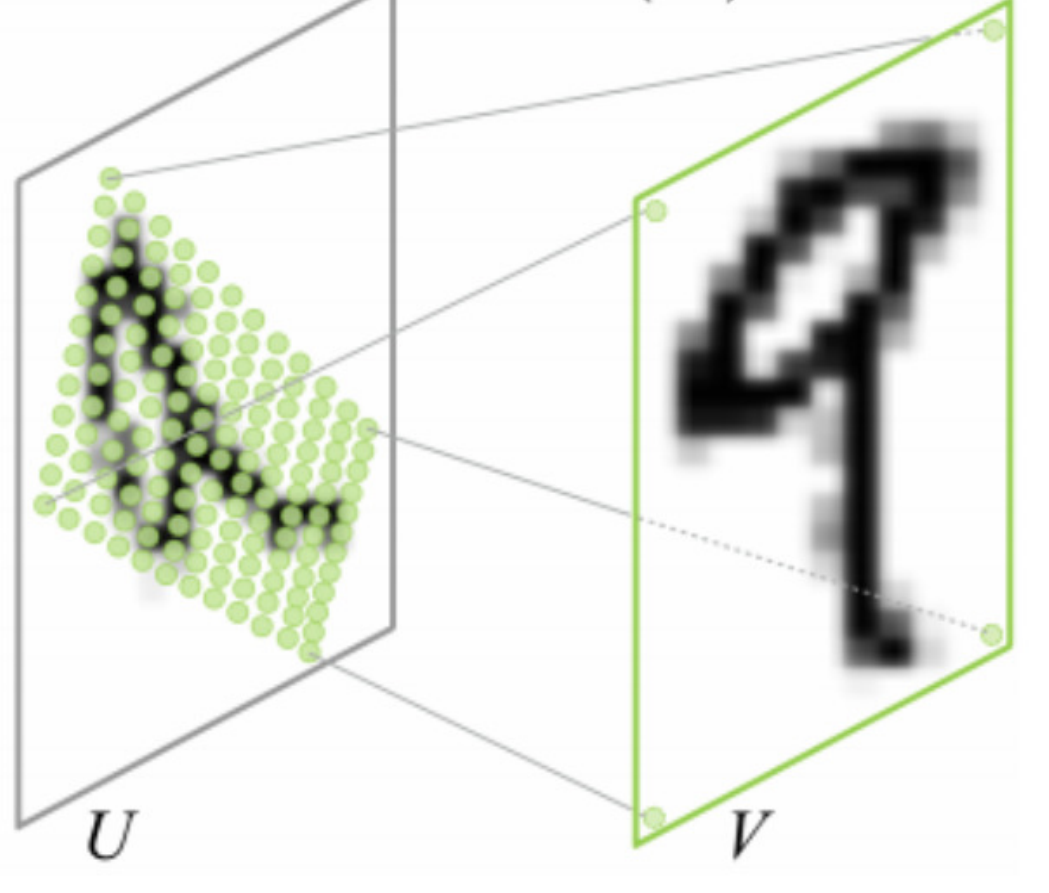}\tabularnewline
(a) & (b)\tabularnewline
\end{tabular}
\par\end{centering}
\caption[Spatial Transformer Networks operations. ]{\label{fig:stn}Spatial Transformer Networks operations. (a) Depictions
of a Spatial Transformer module, a typical transformation operation
is illustrated in (b). Figure reproduced from \cite{Jaderberg2016}.}

\end{figure}

A prominent ConvNet that explicitly tackles invariance maximization
is the Spatial Transformer Network (STN) \cite{Jaderberg2016}. In
particular, this network makes use of a novel learned module that
increased invariance to unimportant spatial transformations, \eg
those that result from varying viewpoint during object recognition.
The module is comprised of three submodules: A localization net, a
grid generator and a sampler, as shown in Figure \ref{fig:stn}(a).
The operations performed can be summarized in three steps. First,
the localization net, which is usually a small 2 layer neural network,
takes a feature map, $U$, as input and learns transformation parameters,
$\theta$, from this input. For example, the transformation, $\mathcal{T}_{\theta}$,
can be defined as a general affine transformation allowing the network
to learn translations, scalings, rotations and shears. Second, given
the transformation parameters and an output grid of pre-defined size,
$H\times W$, the grid generator calculates for each output coordinate,
$(x_{i}^{t},y_{i}^{t})$, the corresponding coordinates, $(x_{i}^{s},y_{i}^{s})$,
that should be sampled from the input, $U$, according to

\begin{equation}
\left(\begin{array}{c}
x_{i}^{s}\\
y_{i}^{s}
\end{array}\right)=\left[\begin{array}{ccc}
\theta_{11} & \theta_{12} & \theta_{13}\\
\theta_{21} & \theta_{22} & \theta_{23}
\end{array}\right]\left(\begin{array}{c}
x_{i}^{t}\\
y_{i}^{t}\\
1
\end{array}\right).\label{eq:STN}
\end{equation}
Finally, the sampler takes the feature map, $U$, and the sampled
grid and interpolates the pixels values, $(x_{i}^{s},y_{i}^{s})$,
to populate the output feature map, $V$, at locations $(x_{i}^{t},y_{i}^{t})$
as illustrated in Figure \ref{fig:stn}(b). Adding such modules at
each layer of any ConvNet architecture allows it to learn various
transformations adaptively from the input to increase its invariance
and thereby improve its accuracy. 

\begin{figure}[H]
\begin{centering}
\includegraphics[width=0.7\textwidth]{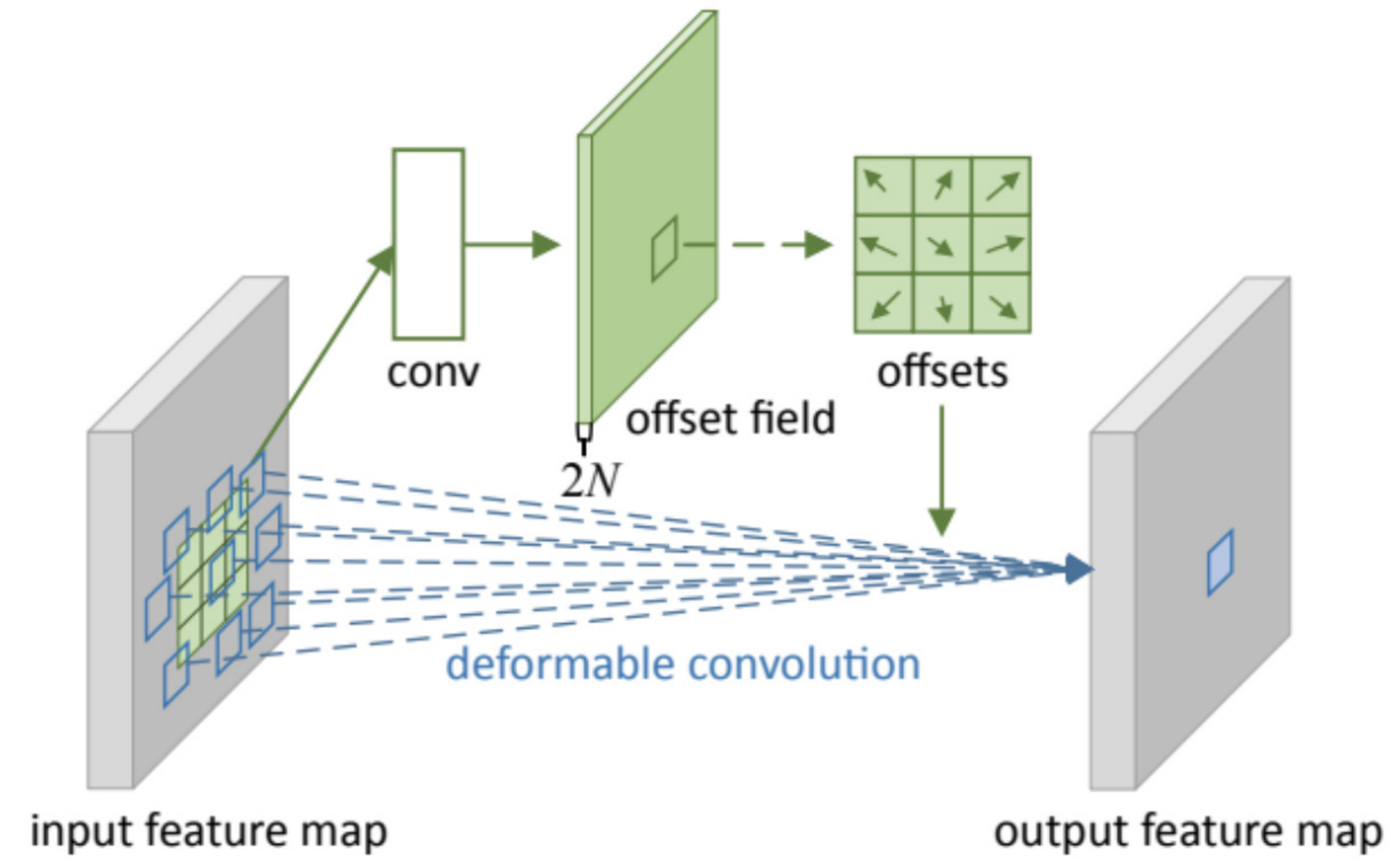}
\par\end{centering}
\caption[Deformable or active convolution.]{\label{fig:deformablecnn}Deformable or active convolution. Starting
from a fixed window size the network learns offsets via a small subnetwork
(shown in the top part of the figure in green) and finally performs
convolution on a deformed window. Figure reproduced from \cite{dai17dcn}.}

\end{figure}

With the same goal of enhancing the geometric transformation modeling
capability of ConvNets, two contemporary approaches, known as Deformable
ConvNet \cite{dai17dcn} and Active ConvNet \cite{Jeon2017}, introduce
a flexible convolutional block. The basic idea in these approaches
is to eschew the use of rigid windows during convolution in favor
of learning Regions of Interest (RoI) over which convolutions are
performed. This idea is akin to what is done by the localization network
and the grid generator of a Spatial Transformer module. To determine
the RoIs at each layer, the convolutional block is modified such that
it learns offsets from the initial rigid convolution window. Specifically,
starting from the standard definition of a convolution operation over
a rigid window given by

\begin{equation}
y(p)=\sum_{p_{n}\in R}w(p_{n})\,x(p-p_{n}),\label{eq:conv}
\end{equation}
where $R$ is the region over which convolution is performed, $p_{n}$
are the pixel locations within the region $R$ and $w(p_{n})$ are
the corresponding filter weights, an new term is added to include
offsets according to

\begin{equation}
y(p)=\sum_{p_{n}\in R}w(p_{n})\,x(p-p_{n}-\triangle p_{n}),\label{deformanble_conv}
\end{equation}
where $\triangle p_{n}$ are the offsets and now the final convolution
step will be performed over a deformed window instead of the traditional
rigid $n\times n$ window. To learn the offsets, $\triangle p_{n}$,
the convolutional block of Deformable ConvNets is modified such that
it includes a new submodule whose role is to learn the offsets as
depicted in Figure \ref{fig:deformablecnn}. Different from Spatial
Transformer Networks that alternately learn the submodule parameters
and the network weights, Deformable ConvNets learn the weights and
offsets concurrently, thus making it faster and easier to deploy in
various architectures. 

\subsection{Toward ConvNet localization}

\begin{figure}[H]
\begin{centering}
\includegraphics[width=0.5\textwidth]{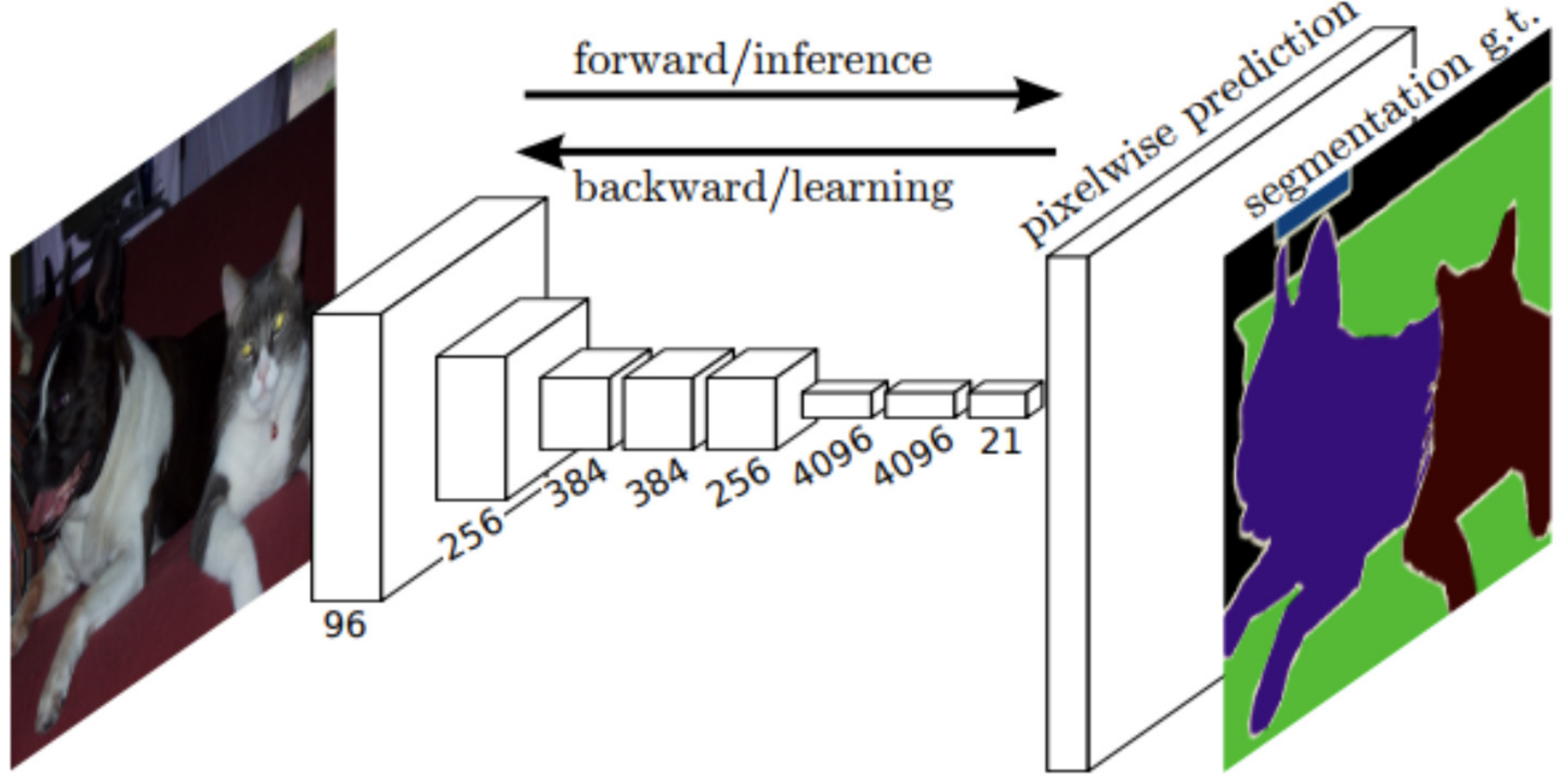}
\par\end{centering}
\caption[Fully Convolutional Network.]{\label{fig:fcn}Fully Convolutional Network. After upsampling to
recover the image full resolution at the last layer, each pixel is
classified using a softmax to finally generate the segments. Figure
reproduced from \cite{Long2015}.}
\end{figure}

Beyond simple classification tasks, such as object recognition, recently
ConvNets have been excelling at tasks that require accurate localization
as well, such as semantic segmentation and object detection. Among
the most successful networks for semantic segmentation is the so called
Fully Convolutional Network (FCN) \cite{Long2015}. As the name implies,
FCN does not make use of fully connected layers explicitly but instead
casts them as convolutional layers whose receptive fields cover the
entire underlying feature map. Importantly, the network learns an
upsampling or deconvolution filter that recovers the full resolution
of the image at the last layer as depicted in Figure \ref{fig:fcn}.
In FCN, the segmentation is achieved by casting the problem as a dense
pixelwise classification. In other words, a softmax layer is attached
to each pixel and segmentation is achieved by grouping pixels that
belong to the same class. Notably, it was reported in this work that
using features from lower layers of the architecture in the upsampling
step plays an important role. It allowed for more accurate segmentation
given that lower layer features tend to capture finer grained details,
which are far more important for a segmentation task compared to classification.
An alternative to learning a deconvolution filter, relies on using
atrou or dilated convolutions \cite{chen14deeplab}, \ie upsampled
sparse filters, which helps recovering higher resolution feature maps
while keeping the number of parameters to be learned manageable. 

When it comes to object localization, one of the earliest approaches
within the ConvNet framework is known as Region CNN or R-CNN. This
network combined a region proposal method with a ConvNet architecture
\cite{girshick2014rcnn}. Although R-CNN was built around simple ideas,
it yielded state-of-the-art object detection results. In particular,
R-CNN first uses an off-the-shelf algorithm for region proposals (\eg
selective search \cite{Uijlings2013}) to detect potential regions
that may contain an object. These regions are then warped to match
the default input size of the employed ConvNet and fed into a ConvNet
for feature extraction. Finally, each region's features are classified
with an SVM and refined in a post processing step via non-maximum
suppression. 

In its naive version, R-CNN simply used ConvNets as a feature extractor.
However, its ground breaking results led to improvements that take
more advantage of ConvNets' powerful representation. Examples include,
Fast R-CNN \cite{girshick15fastrcnn}, Faster R-CNN \cite{HeRenfaster}
and Mask R-CNN \cite{He_2017_ICCV}. Fast R-CNN, proposes propagating
the independently computed region proposals through the network to
extract their corresponding regions in the last feature map layer.
This technique, avoids costly passes through the network for each
region extracted from the image. In addition, Fast R-CNN avoids heavy
post-processing steps by changing the last layer of the network such
that it learns both object classes and refined bounding box coordinates.
Importantly, in both R-CNN and Fast R-CNN the detection bottleneck
lies in the region proposal step that is done outside of the ConvNet
paradigm. 

\begin{figure}[H]
\begin{centering}
\begin{tabular}{cc}
\includegraphics[width=0.4\textwidth]{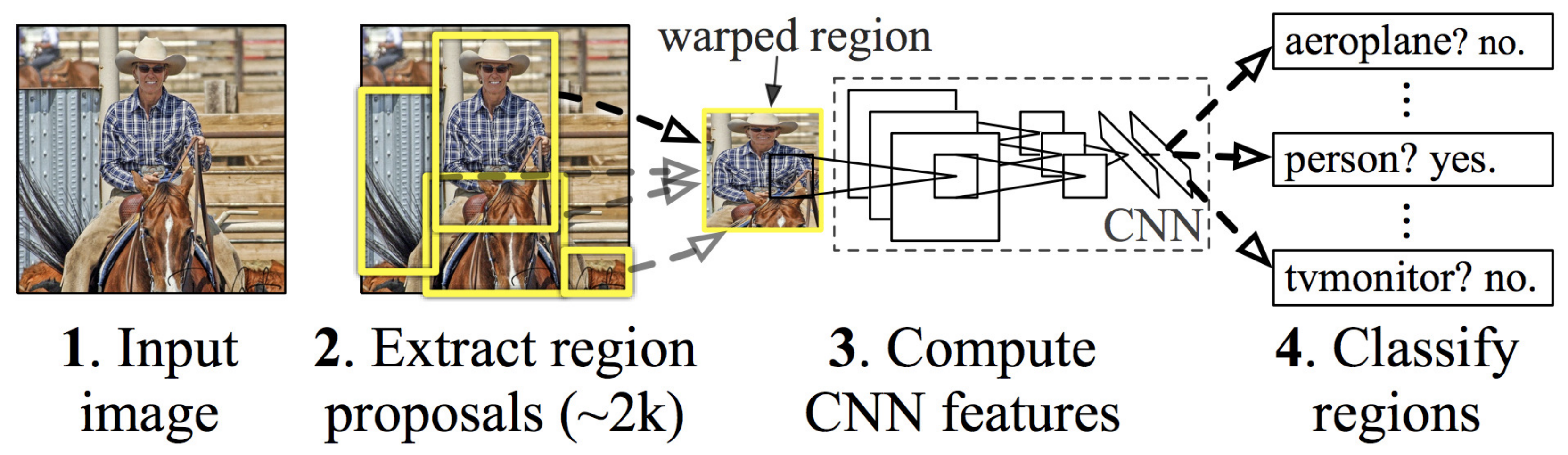} & \includegraphics[width=0.2\textwidth]{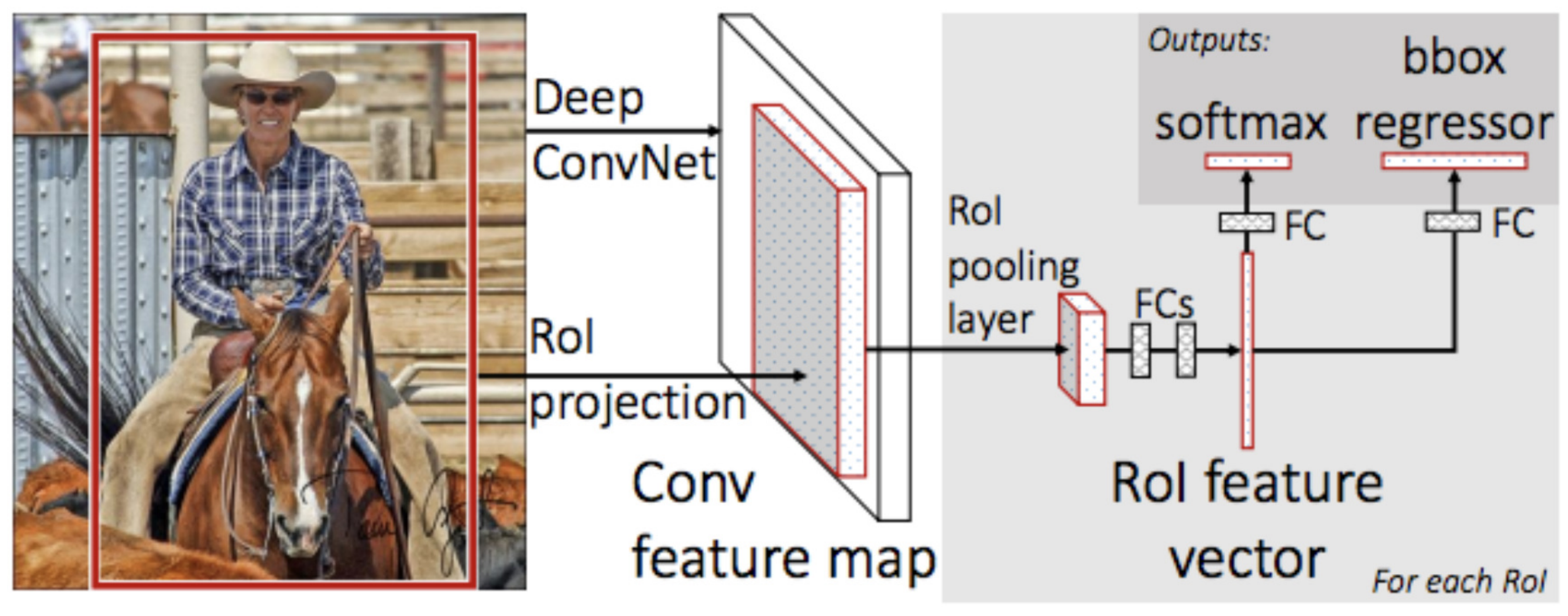}\tabularnewline
(a) & (b)\tabularnewline
\includegraphics[width=0.2\textwidth]{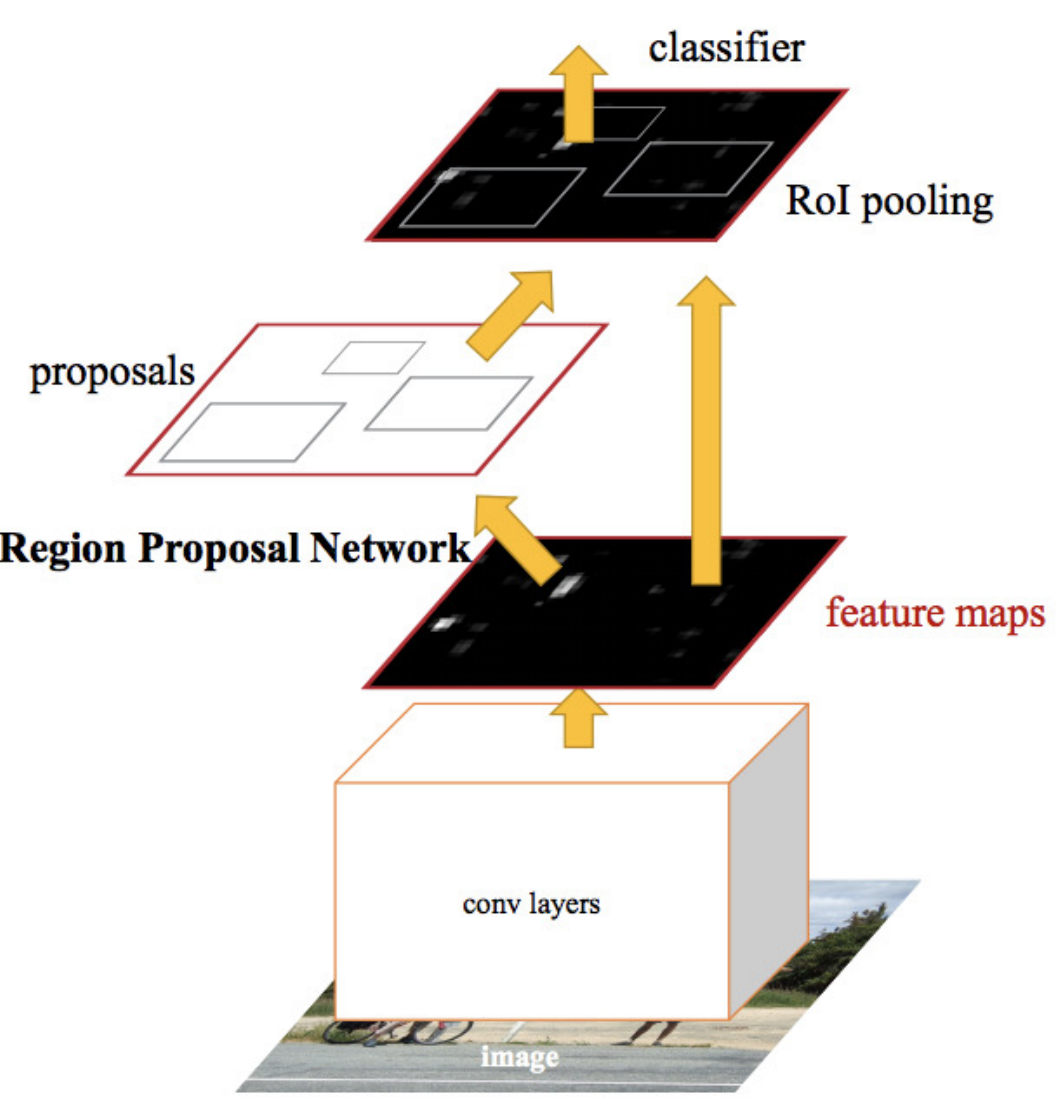} & \includegraphics[width=0.2\textwidth]{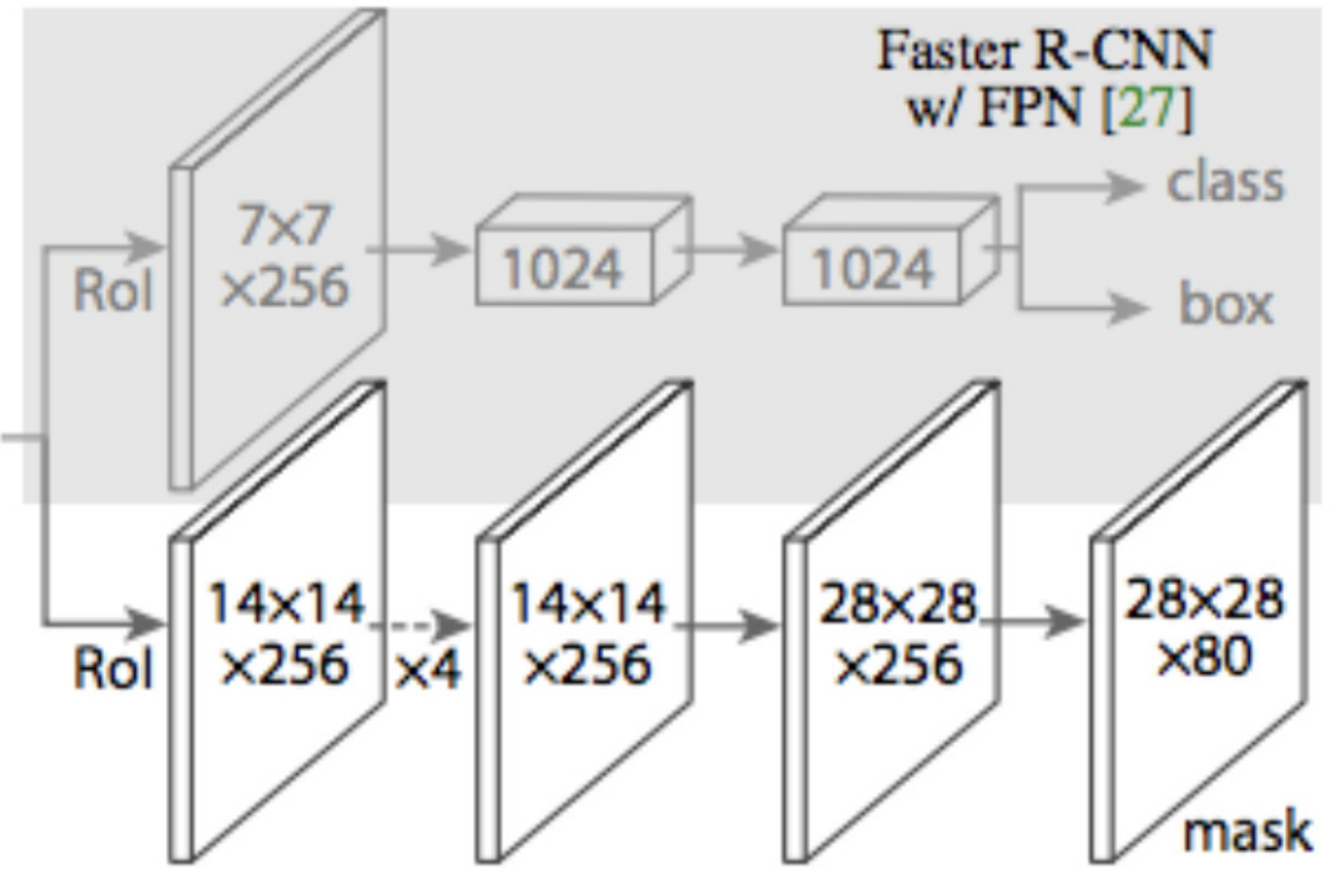}\tabularnewline
(c) & (d)\tabularnewline
\end{tabular}
\par\end{centering}
\caption[Progress of prominent region proposal networks. ]{\label{fig:rcnn}Progress of prominent region proposal networks. (a)
Structure of the original R-CNN. Figure reproduced from \cite{girshick2014rcnn}.
(b) Structure of Fast R-CNN. Figure reproduced from \cite{girshick15fastrcnn}.
(c) Structure of Faster R-CNN. Figure reproduced from \cite{HeRenfaster}.(d)
Structure of Mask R-CNN. Figure reproduced from \cite{He_2017_ICCV}.}

\end{figure}

Faster R-CNN pushes the use of ConvNets even further by adding a sub-module
(or sub-network), called Region Proposal Network (RPN), after the
last convolutional layer of a ConvNet. An RPN module enables the network
to learn the region proposals as part of the network optimization.
Specifically, RPN is designed as a small ConvNet consisting of a convolutional
layer and a small fully connected layer and two outputs that return
potential object positions and objectness scores (\ie probability
of belonging to an object class). The entire network's training is
achieved following an iterative two step procedure. First, the network
is optimized for region proposal extraction using the RPN unit. Second,
keeping the extracted region proposals fixed, the network is finetuned
for object classification and final object bounding box position.
More recently, mask R-CNN was introduced to augment faster R-CNN with
the ability to segment the detected regions yielding tight masks around
the detected objects. To this end, mask R-CNN adds a segmentation
branch to the classification and bounding box regression branches
of faster R-CNN. In particular, the new branch is implemented as a
small FCN that is optimized for classifying pixels in any bounding
box to one of two classes; foreground or background. Figure \ref{fig:rcnn}
illustrates the differences and progress from simple R-CNN to mask
R-CNN. 

\section{Spatiotemporal convolutional networks}

The significant performance boost brought to various image based applications
via use of ConvNets, as discussed in Section \ref{sec:Spatial-Convolutional-Networks},
sparked interest in extending 2D spatial ConvNets to 3D spatiotemporal
ConvNets for video analysis. Generally, the various spatiotemporal
architectures proposed in the literature have simply tried to extend
2D architectures from the spatial domain, $(x,y)$, into the temporal
domain, $(x,y,t)$. In the realm of training based spatiotemporal
ConvNets, there are three different architectural design decisions
that stand out: LSTM based (\eg\cite{NgHVVMT15,Donahue17}), 3D (\eg\cite{Tran2015,Karpathy2014})
and Two-Stream ConvNets (\eg\cite{SimonyanTwoStream,Chris2016cvpr}),
which will be described in this section.

\subsection{LSTM based spatiotemporal ConvNet}

\begin{figure}[H]
\begin{centering}
\includegraphics[width=0.45\textwidth]{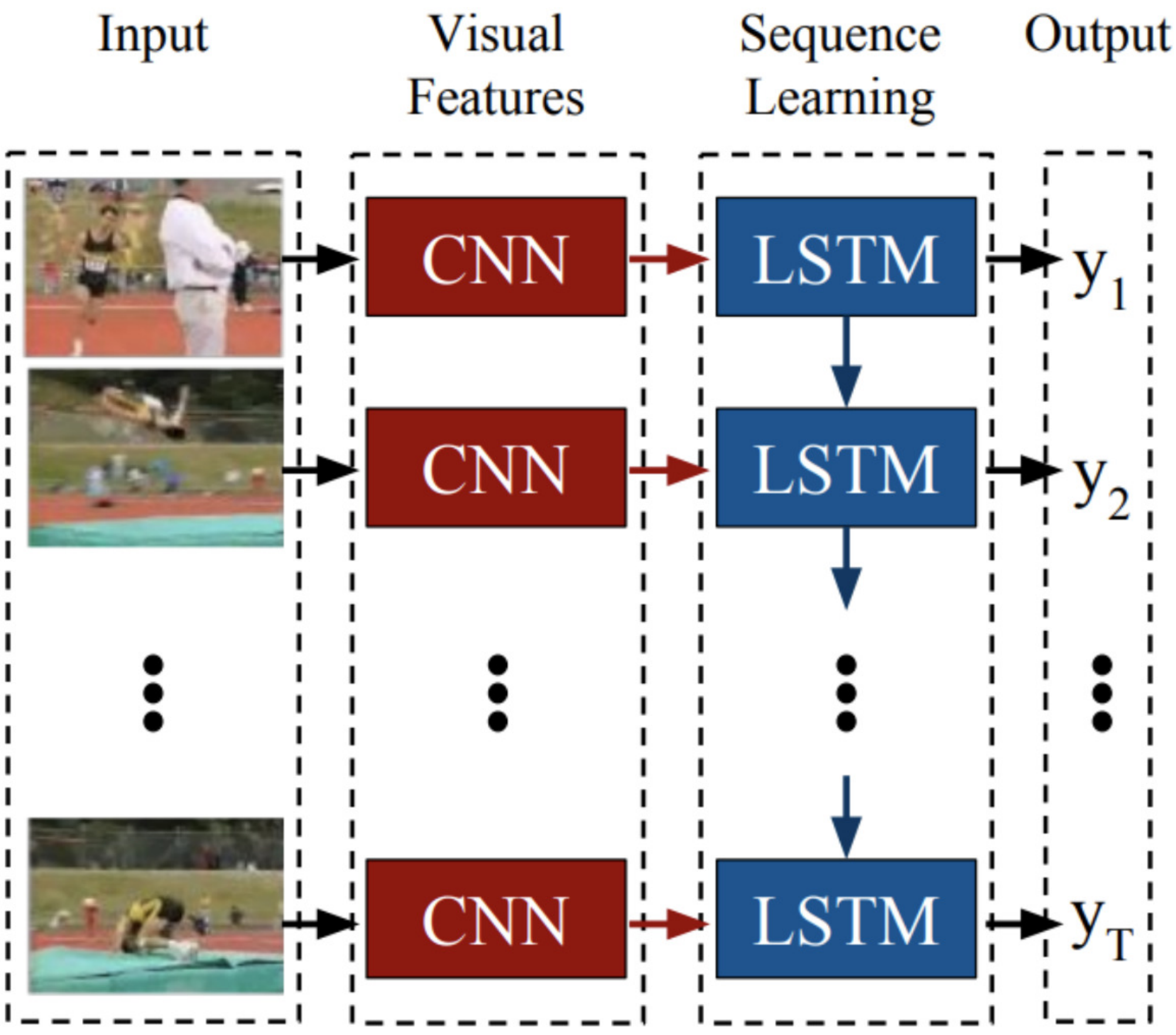}
\par\end{centering}
\caption[Sample LSTM based spatiotemporal ConvNet.]{\label{fig:lstm_cnn}Sample LSTM based spatiotemporal ConvNet. In
this network the input consists of consecutive frames from a video
stream. Figure reproduced from \cite{Donahue17}.}
\end{figure}

LSTM based spatiotemporal ConvNets, \eg \cite{NgHVVMT15,Donahue17},
were some of the early attempts to extend 2D networks to spacetime
processing. Their operations can be summarized in three steps as shown
in Figure \ref{fig:lstm_cnn}. First, each frame is processed with
a 2D network and feature vectors are extracted from their last layer.
Second, these features, from different time steps, are then used as
input to LSTMs that produce temporal outcomes, $y_{t}$. Third, these
outcomes are then either averaged or linearly combined and passed
to a softmax classifier for final prediction. 

The goal of LSTM based ConvNets is to progressively integrate temporal
information while not being restricted to a strict input size (temporally).
One of the benefits of such an architecture is equipping the network
with the ability to produce variable size text descriptions (\ie
a task at which LSTMs excel), as done in \cite{Donahue17}. However,
while LSTMs can capture global motion relationships, they may fail
at capturing finer grained motion patterns. In addition, these models
are usually larger, need more data and are therefore hard to train.
To date, excepting cases where video and text analysis are being integrated
(\eg \cite{Donahue17}), LSTMs generally have seen limited success
in spatiotemporal image analysis.

\subsection{3D ConvNet}

The second prominent type of spatiotemporal networks provides the
most straightforward generalization of standard 2D ConvNet processing
to image spacetime. It works directly with temporal streams of RGB
images and operates on these images via application of learned 3D,
$(x,y,t)$, convolutional filters. Some of the early attempts at this
form of generalization use filters that extend into the temporal domain
with very shallow networks \cite{Ji2013a} or only at the first convolutional
layer \cite{Karpathy2014}. When using 3D convolutions at the first
layer only, small tap spatiotemporal filters are applied on each 3
or 4 consecutive frames. To capture longer range motions multiple
such streams are used in parallel and the hierarchy that results from
stacking such streams increases the network's temporal receptive field.
However, because spatiotemporal filtering is limited to the first
layer only, this approach did not yield a dramatic improvement over
a naive frame based application of 2D ConvNets. A stronger generalization
is provided by the now widely used C3D network, that uses 3D convolution
and pooling operations at all layers \cite{Tran2015}. The direct
generalization of C3D from a 2D to a 3D architecture entails a great
increase in the number of parameters to be learned, which is compensated
for by using very limited spacetime support at all layers (\ie $3\times3\times3$
convolutions). A recent, slightly different, approach proposes integration
of temporal filtering by modifying the ResNet architecture \cite{He2016}
to become a Temporal ResNet (T-ResNet) \cite{Chris2017SCENES}. In
particular, T-ResNet augments the residual units (shown in Figure
\ref{fig:resnet}(a)) with a $1\times1\times T$ filter that applies
one dimensional learned filtering operations along the temporal dimension. 

Ultimately, the goal of such 3D ConvNet architectures is to directly
integrate spacetime filtering throughout the model in order to capture
both appearance and motion information at the same time. The main
downside of these approaches is the entailed increase in the number
of their parameters.

\subsection{Two-Stream ConvNet}

\begin{figure}[H]
\begin{centering}
\includegraphics[width=0.8\textwidth]{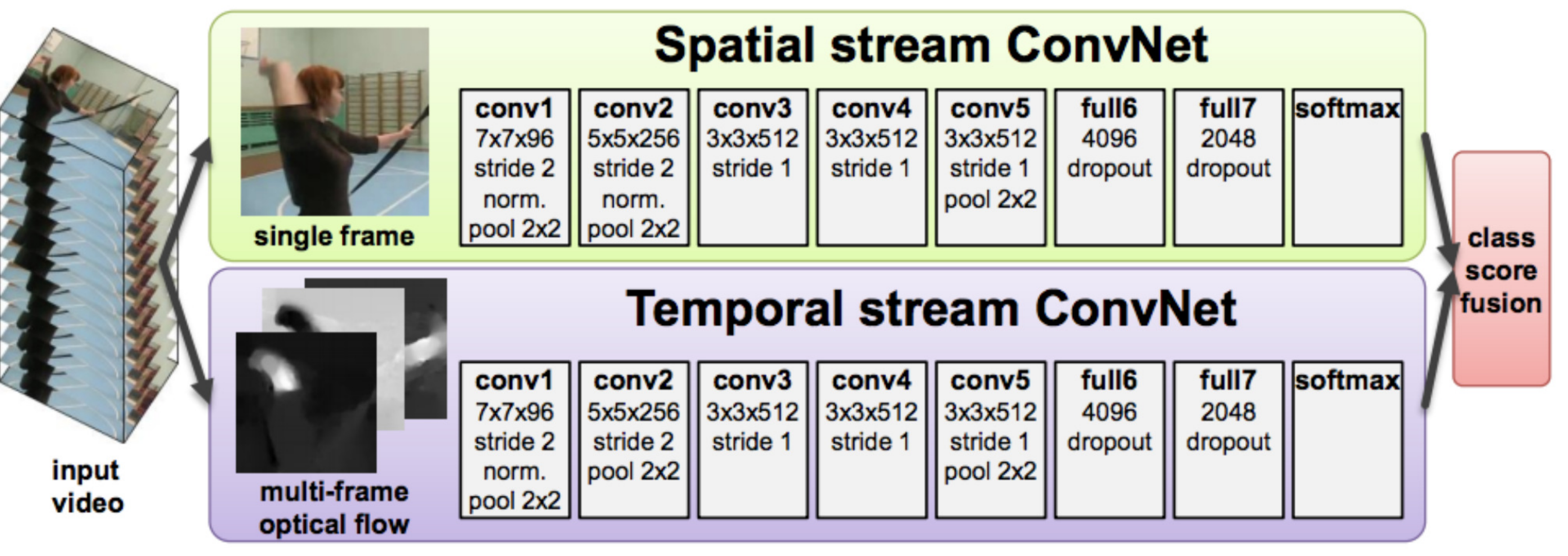}
\par\end{centering}
\caption[The original Two-Stream Network.]{\label{fig:twostream}The original Two-Stream Network. The network
takes as input RGB frames and stacks of optical flow. Figure reproduced
from \cite{SimonyanTwoStream}.}
\end{figure}

The third type of spatiotemporal architecture relies on a two-stream
design. The standard Two-Stream architecture \cite{SimonyanTwoStream},
depicted in Figure \ref{fig:twostream}, operates in two parallel
pathways, one for processing appearance and the other for motion by
analogy with the two-stream hypothesis in the study of biological
vision systems \cite{GOODALE1992}. Input to the appearance pathway
are RGB images; input to the motion path are stacks of optical flow
fields. Essentially, each stream is processed separately with fairly
standard 2D ConvNet architectures. Separate classification is performed
by each pathway, with late fusion used to achieve the final result.
The various improvements over the original two stream network follow
from the same underlying idea while using various baseline architectures
for the individual streams (\eg\cite{Chris2016cvpr,liminwang2015,liminwang2016})
or proposing different ways of connecting the two streams (\eg \cite{Chris2016cvpr,Chris2016nips,Chris2017ACTIONS}).
Notably, recent work known as I3D \cite{carreira2017}, proposes use
of both 3D filtering and Two-Stream architectures via use of 3D convolutions
on both streams. However, the authors do not present compelling arguments
to support the need for a redundant optical flow stream in addition
to 3D filtering, beyond the fact that the network achieves slightly
better results on benchmark action recognition datasets.

Overall, Two-Stream ConvNets support the separation of appearance
and motion information for understanding spatiotemporal content. Significantly,
this architecture seems to be the most popular among spatiotemporal
ConvNets as its variations led to state-of-the-art results on various
action recognition benchmarks (\eg\cite{Chris2016cvpr,Chris2016nips,Chris2017ACTIONS,liminwang2016}). 

\section{Overall discussion }

Multilayer representations have always played an important role in
computer vision. In fact, even standard widely used hand crafted features
such as SIFT \cite{lowe04} can be seen as a shallow multilayer representation,
which loosely speaking consists of a convolutional layer followed
by pooling operations. Moreover, pre-ConvNet state-of-the-art recognition
systems typically followed hand-crafted feature extraction with (learned)
encodings followed by spatially organized pooling and a learned classifier
(\eg \cite{chris2015}), which also is a multilayer representational
approach. Modern multilayer architectures push the idea of hierarchical
data representation deeper while typically eschewing hand designed
features in favor of learning based approaches. When it comes to computer
vision applications, the specific architecture of ConvNets makes them
one of the most attractive architectures. 

Overall, while the literature tackling multilayer networks is very
large where each faction advocates the benefits of one architecture
over another, some common ``best practices'' have emerged. Prominent
examples include: the reliance of most architectures on four common
building blocks (\ie convolution, rectification, normalization and
pooling), the importance of deep architectures with small support
convolutional kernels to enable abstraction with a manageable number
of parameters, residual connections to combat challenges in error
gradient propagation during learning. More generally, the literature
agrees on the key point that good representations of input data are
hierarchical, as previously noted in several contributions \cite{Rodriguez2015}.

Importantly, while these networks achieve competitive results in many
computer vision applications, their main shortcomings remain: the
limited understanding of the exact nature of the learned representation,
the reliance on massive training datasets, the lack of ability to
support precise performance bounds and the lack of clarity regarding
the choice of the networks hyper parameters. These choices include
the filters sizes, choice of nonlinearities, pooling functions and
parameters as well as the number of layers and architectures themselves.
Motivations behind several of these choices, in the context of ConvNets'
building block, are discussed in the next chapter. 

\chapter{Understanding ConvNets Building Blocks \label{cha:chapter3}}

In the light of the plethora of unanswered questions in the ConvNets
area, this chapter investigates the role and significance of each
layer of processing in a typical convolutional network. Toward this
end, the most prominent efforts tackling these questions are reviewed.
In particular, the modeling of the various ConvNet components will
be presented both from theoretical and biological perspectives. The
presentation of each component ends with a discussion that summarizes
our current level of understanding.

\section{The convolutional layer\label{sec:The-Convolutional-Layer}}

The convolutional layer is, arguably, one of the most important steps
in ConvNet architectures. Basically, convolution is a linear, shift
invariant operation that consists of performing local weighted combination
across the input signal. Depending on the set of weights chosen (\ie
the chosen point spread function) different properties of the input
signal are revealed. In the frequency domain, the correlate of the
point spread function is the modulation function that tells how the
frequency components of the input are modified through scaling and
phase shifting. Therefore, it is of paramount importance to select
the right kernels to capture the most salient and important information
contained in the input signal that allows for making strong inferences
about the content of the signal. This section discusses some of the
different ways to approach the kernel selection step.

\subsection{Biological perspective}

Neurophysiological evidence for hierarchical processing in the mamalian
visual cortex provides an underlying inspiration for spatial and spatiotemporal
ConvNets. In particular, research that hypothesized a cascade of simple
and complex cells that progressively extract more abstract attributes
of the visual input \cite{hubel1962} has been of particular importance.
At the very earliest stages of processing in the visual cortex, the
simple cells were shown capable of detecting primitive features such
as oriented gratings, bars and edges, with more complicated tunings
emerging at subsequent stages.

A popular choice for modeling the described properties of cortical
simple cells is a set of oriented Gabor filters or Gaussian derivatives
at various scales. More generally, filters selected at this level
of processing typically are oriented bandpass filters. Many decades
later most biological models still rely on the same set of simple
cells at the initial layers of the hierarchy \cite{Riesenhuber99,serre2005,Serre2007,Jhuang2007,baker2001,FreemanMovshon2013}.
In fact, these same Gabor kernels are also extended to the chromatic
\cite{ZhangSerr2012} and temporal \cite{Jhuang2007} domains to account
for color and motion sensitive neurons, respectively. 

\begin{figure}
\begin{centering}
\includegraphics[width=0.5\textwidth]{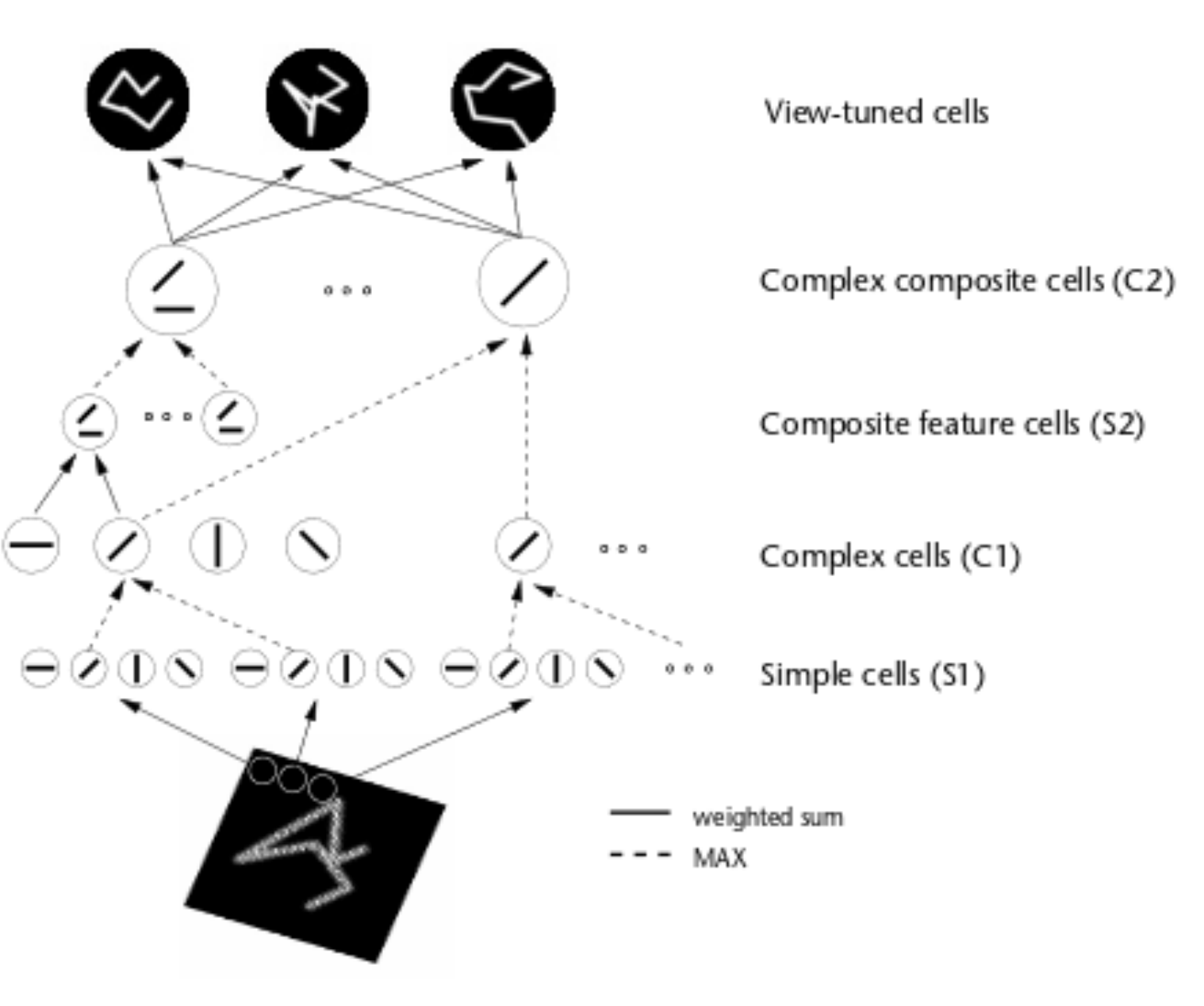}
\par\end{centering}
\caption[Illustration of the HMAX Model.]{\label{fig:hmax_model}Illustration of the HMAX Model. This model
consists of a hierarchy of cells with alternating simple ($S$) and
complex ($C$) cells. Filtering operations happen at the level of
the $S$ cells. It is shown in this figure that simple cells at the
initial layer ($S1$) detect simple oriented bars (\ie through the
use of oriented Gabor filters). On the other hand, simple cells at
higher layers ($S2$) respond to filtering with templates that are
combinations of filters used at the previous ($S1$) layer such that
cells at higher layers in the hierarchy detect more complex shapes
than oriented bars. Complex composite cells (C1, C2) intervene between
the layers of simple cells to aggregate similarly tuned cells across
spatial position and thereby achieve a degree of shift invariance.
Figure reproduced from \cite{Riesenhuber99}.}
\end{figure}

Matters become more subtle, however, when it comes to representing
cells at higher areas of the visual cortex and most contributions
building on Hubel and Wiesel's work strive to find an appropriate
representation for these areas. The HMAX model is one of the most
well known models tackling this issue \cite{Riesenhuber99}. The main
idea of the HMAX model is that filters at higher layers of the hierarchy
are obtained through the combination of filters from previous layers
such that neurons at higher layers respond to co-activations of previous
neurons. This method ultimately should allow the model to respond
to more and more complex patterns at higher layers as illustrated
in Figure \ref{fig:hmax_model}. This approach relates nicely to the
Hebbian theory stating that ``cells that fire together, wire together''
\cite{hebb1949}.

\begin{figure}[H]
\begin{centering}
\includegraphics[width=0.7\textwidth]{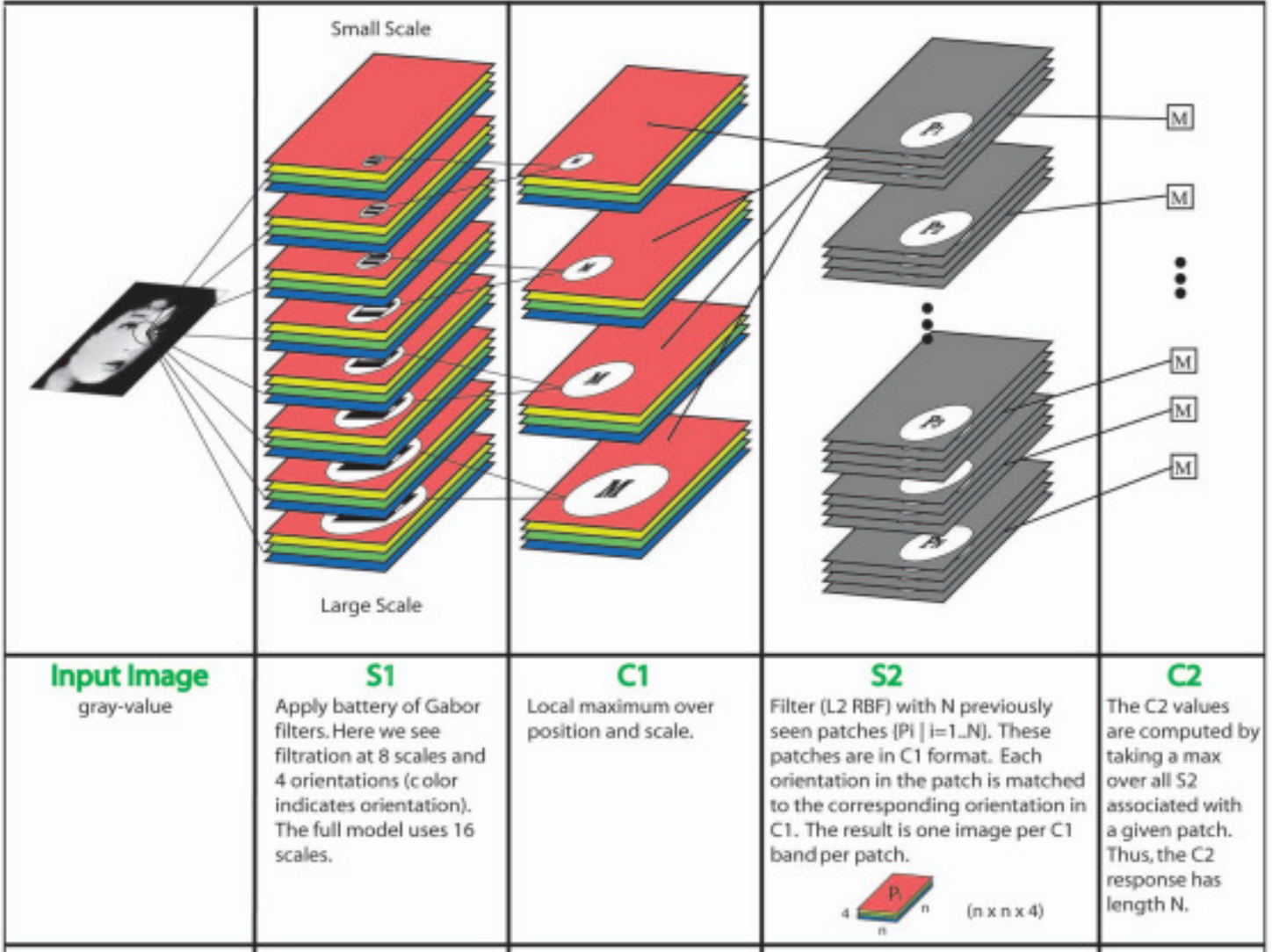}
\par\end{centering}
\caption[The Network Architecture Proposed by Serre \etal.]{\label{fig:serrenetwork}The Network Architecture Proposed by Serre
\etal. Similarly to the HMAX model \cite{Riesenhuber99}, it consists
of alternating simple and complex cells such that the overall architecture
of the proposed networks can be summarized as $S1\rightarrow C1\rightarrow S2\rightarrow C2$.
However, as opposed to the HMAX model, templates used at the level
of the S2 cells are explicitly learned from a training set such that
this layer detects complex objects (\ie when trained with an object
recognition dataset). The details of the process are summarized in
the second row of the figure. Figure reproduced from \cite{Serre2007}.}
\end{figure}

Another hallmark of the HMAX model is the assumption that learning
comes into play in order to recognize across various viewpoints of
similar visual sequences. Direct extensions of this work thereafter
explicitly introduce learning to model filters at higher layers. Among
the most successful such approaches is the biologically motivated
network introduced by Serre \etal \cite{Serre2007} that attempts
to model the processes taking place at the initial layers of the visual
cortex with a network made of $4$ layers where simple ($S$) and
complex ($C$) cells alternate as illustrated in Figure \ref{fig:serrenetwork}.
It is seen that each simple cell is directly followed by a complex
cell such that the overall structure of the network can be summarizes
as $S1\rightarrow C1\rightarrow S2\rightarrow C2$. In this network
convolutions take place at the level of the $S1$ and $S2$ units.
While the $S1$ units rely on $2D$ oriented Gabor filters, the kernels
used at the second layer are based on a learning component. This choice
is motivated by biological evidence suggesting that learning occurs
at the higher layers of the cortex \cite{serre2005}, although there
also is evidence that learning plays a role at earlier layers of the
visual cortex \cite{blakemore1970}. In this case, the learning process
corresponds to selecting a random set of $n\times n\times l$ patches,
$\mathbf{P}_{i}$, from a training set at the $C1$ layer, where $n$
is the spatial extent of the patch and $l$ corresponds to the number
of orientations. The $S2$ layer feature maps are obtained by performing
template matching between the $C1$ features in each scale and the
set of learned patches $\mathbf{P}_{i}$ at all orientations simultaneously. 

A direct extension to this work exists for video processing \cite{Jhuang2007}.
The kernels used for video processing are designed to mimic the behavior
of cells in the dorsal stream. In this case, $S1$ units involve convolutions
with 3D oriented filters. In particular, third order Gaussian derivative
filters are used owing to their nice separability properties and a
similar learning process is adopted to select convolutional kernels
for the $S2$ and $S3$ units. 

Many variations of the above underlying ideas have been proposed,
including various learning strategies at higher layers \cite{Woodbeck2008,Wei2014},
wavelet based filters \cite{Hong2011}, different feature sparsification
strategies \cite{Huang2008,Woodbeck2008,MutchLowe2006} and optimizations
of filter parameters \cite{Woodbeck2008,Mishra2010}. 

Another related, although somewhat different, train of thoughts suggest
that there exist more complex cells at higher levels of the hierarchy
that are dedicated to capturing intermediate shape representation,
\eg curvatures \cite{john2011,john2012}. While the HMAX class of
models propose modeling shapes via compositions of feature types from
previous layers, these investigations propose an approach that directly
models hypercomplex cells (also referred to as endstopped cells) without
resorting to learning. In particular, models falling within this paradigm
model hypercomplex cells via combination of simple and complex cells
to generate new cells that are able to maximally respond to curvatures
of different degrees and signs as well as different shapes at different
locations. In suggesting that hypercomplex cells subserve curvature
calculations, this work builds on earlier work suggesting similar
functionality, \eg \cite{dobbins1987}.

Yet another body of research, advocates that the hierarchical processing
(termed $Filter\rightarrow Rectify\rightarrow Filter$) that takes
place in the visual cortex deals progressively with higher-order image
structures \cite{baker2001,FreemanMovshon2013,Movshon2014}. It is
therefore advocated that the same set of kernels present at the first
layer (\ie oriented bandpass filters) are repeated at higher layers.
However, the processing at each layer reveals different properties
of the input signal given that the same set of kernels now operate
on different input obtained from a previous layer. Therefore, features
extracted at successive layers progress from simple and local to abstract
and global while capturing higher order statistics. In addition, joint
statistics are also accounted for through the combination of layerwise
responses across various scales and orientations.

\subsubsection*{Discussion}

The ability of human visual cortex in recognizing the world while
being invariant to various changes has been the driving force of many
researchers in this field. Although, several approaches and theories
have been proposed to model the different layers of the visual cortex,
a common thread across these efforts is the presence of hierarchical
processing that splits the vision task into smaller pieces. However,
while most models agree on the choice of the set of kernels at the
initial layers, motivated by the seminal work of Hubel and Wiesel
\cite{hubel1962}, modeling areas responsible for recognizing more
abstract features seems to be more intricate and controversial. Also,
these biologically plausible models, typically leave open critical
questions regarding the theoretical basis of their design decisions.
This shortcoming applies to more theoretically driven models as well,
as will be discussed in the next section.

{} 

\subsection{Theoretical perspective\label{subsec:Conv-Theoretical-perspective}}

More theoretically driven approaches are usually inspired from biology
but strive to inject more theoretical justifications into their models.
These methods usually vary depending on their kernel selection strategy.

\begin{figure}[H]
\begin{centering}
\includegraphics[width=0.9\textwidth]{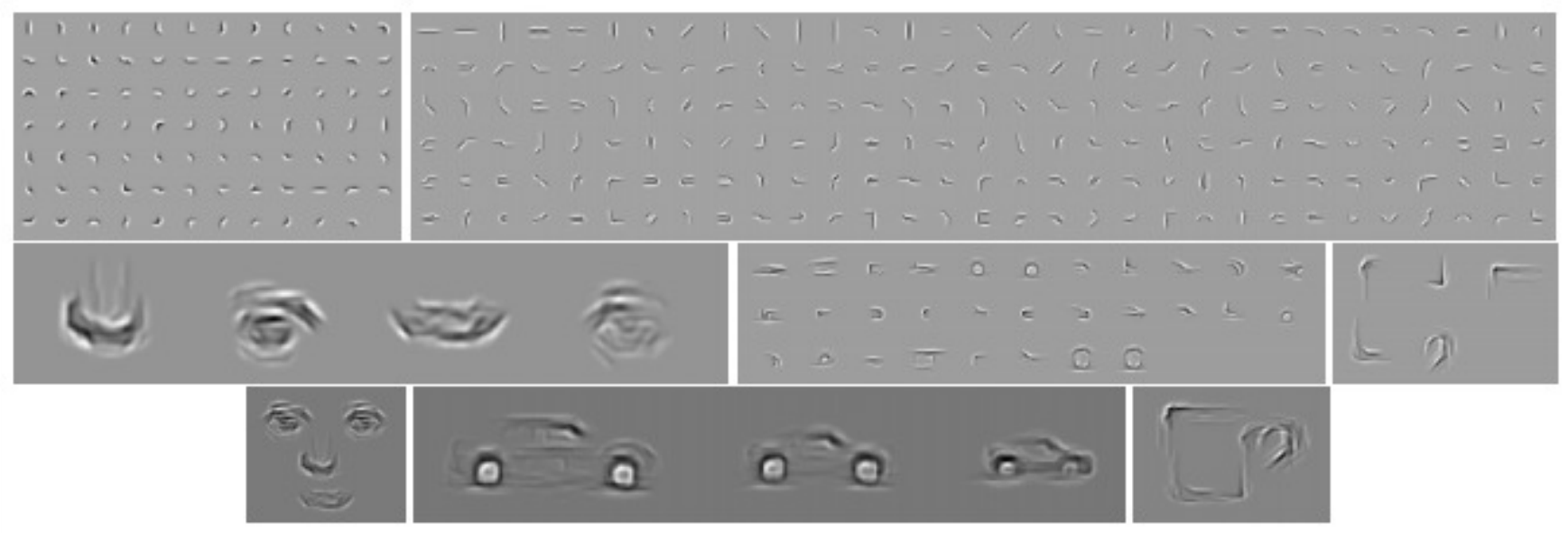}
\par\end{centering}
\caption[Sample Parts Learned by the Multilayer Architecture Proposed by Fidler
\etal.]{\label{fig:fidlernetwork}Sample Parts Learned by the Multilayer Architecture
Proposed by Fidler \etal. \textbf{1st row (left-to-right): }Layer
2 and layer 3 sample parts\textbf{. 2nd and 3rd rows: }Layer 4 and
layer 5 parts learned using faces, cars, and mugs. Figure reproduced
from \cite{Fidler2007}.}
\end{figure}

One way of looking at the kernel selection problem is to consider
that objects in the natural world are a collection of a set of primitive
shapes and thereby adopt a shape based solution %
{} \cite{Fidler2007,Fidler2006,Fidler2008}. In this case, the proposed
algorithms start by finding the most primitive shapes in an image
(\ie oriented edges) using a bank of oriented Gabor filters. Using
these edges, or more generally parts, the algorithm proceeds by finding
potential combinations of parts in the next layers by looking at increasingly
bigger neighborhoods around each part. Basically, every time a new
image is presented to the network, votes are collected about the presence
of other part types in the direct neighborhood of a given part in
the previous layer. After all images present in the training set are
seen by the network, each layer of the network is constructed using
combinations of parts from the previous layer. The choice of the combinations
is based on the probabilities learned during the unsupervised training.
In reality, such a shape based approach is more of a proof of concept
where only lower layers of the hierarchy can be learned in such an
unsupervised way, whereas higher layers are learned using category
specific images as illustrated in Figure \ref{fig:fidlernetwork}.
Therefore, a good representation of an object can be obtained in higher
layers only if the network saw examples from that object class alone.
However, because of this constraint, such an algorithm cannot be reasonably
deployed on more challenging datasets with objects from different
categories that it had not previously seen.

\begin{figure}[H]
\begin{centering}
\includegraphics[width=0.65\textwidth,height=0.2\paperheight]{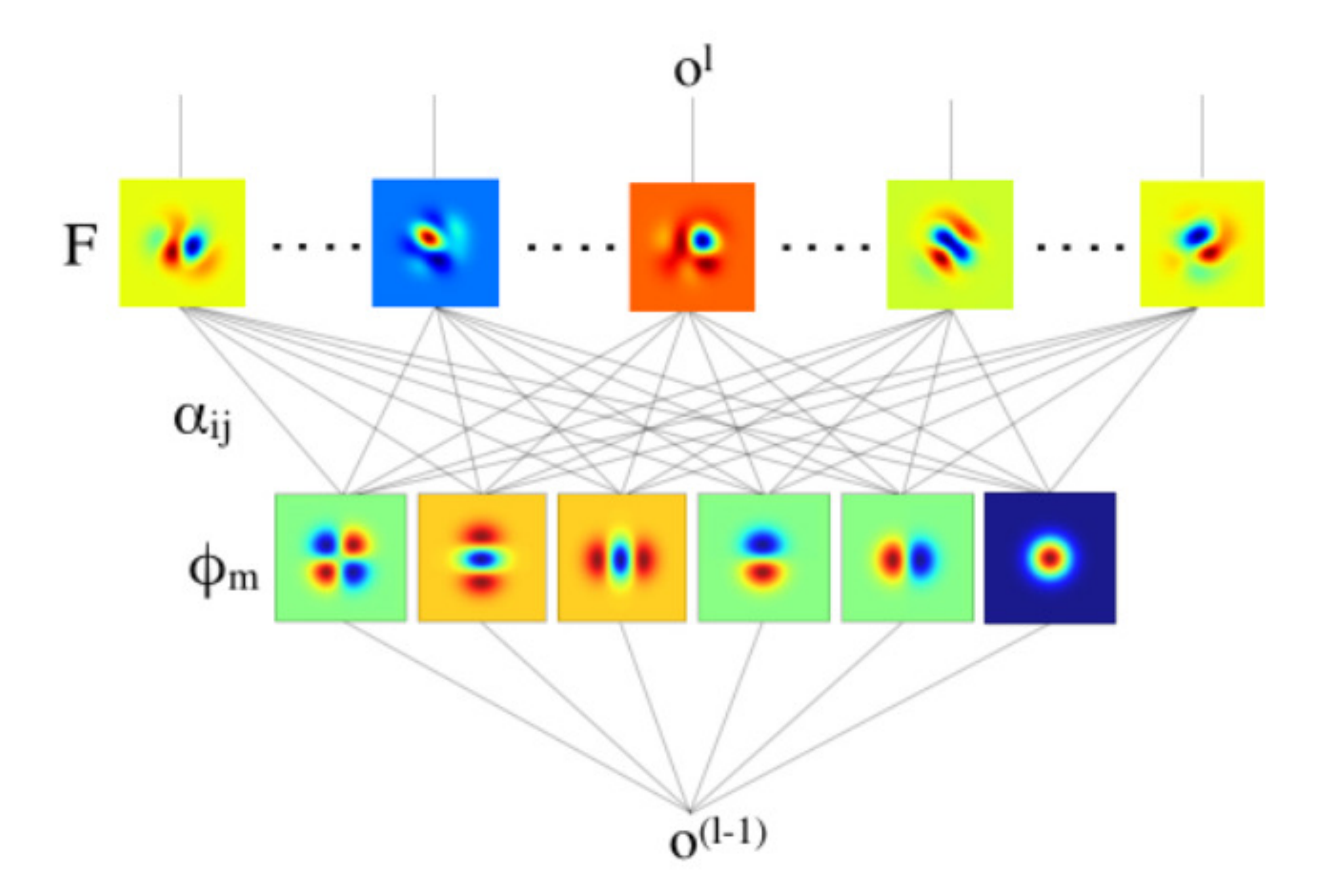}
\par\end{centering}
\caption[An Illustration the Receptive Fields CNN (also known as RFNN).]{\label{fig:jacobsennetwork}An Illustration the Receptive Fields CNN
(also known as RFNN). In this network, the filters used at all layers
are built (via learning) as a linear combination of the basis filter
set $\phi_{m}$, which is a set $n^{th}$ order Gaussian derivatives.
Instead of learning the kernel parameters of the filters, this network
learns the parameters $\alpha_{ij}$ used to linearly combine the
filters in the basis set. Figure reproduced from \cite{Jacobsen16}.}
\end{figure}

Another outlook on the kernel selection process is based on the observation
that many training based convolutional networks learn redundant filters.
Moreover, many of the learned filters at the first few layers of those
networks resemble oriented band pass filters; \eg see Figure \ref{fig:CReU-1}.
Therefore, several recent investigations aim at injecting priors into
their network design with a specific focus on the convolutional filter
selection. One approach proposes learning layerwise filters over a
basis set of 2D derivative operators \cite{Jacobsen16} as illustrated
in Figure \ref{fig:jacobsennetwork}. While this method uses a fixed
basis set of filters, it relies on supervised learning to linearly
combine the filters in the basis at each layer to yield the effective
layerwise filters and it is therefore dataset dependent. Nonetheless,
using a basis set of filters and learning combinations aligns well
with biological models, such as HMAX \cite{Riesenhuber99} and its
successors (\eg \cite{Serre2007,Jhuang2007}), and simplifies the
networks' architecture, while maintaining interpretability. Also,
as learning is one of the bottlenecks of modern ConvNets, using a
basis set also eases this process by tremendously decreasing the number
of parameters to be learned. For these reasons such approaches are
gaining popularity in the most recent literature \cite{Jacobsen16,Cohen2017,Worral2017,Luan2017,Zhou2017ORN}.

Interestingly, a common thread across these recent efforts is the
aim of reducing redundant kernels with a particular focus on modeling
rotational invariance (although it is not necessarily a property of
biological vision). The focus on rotation is motivated by the observation
that, often, learned filters are rotated versions of one another.
For example, one effort targeted learning of rotational equivariance
by training over a set of circular harmonics \cite{Worral2017}. Alternatively,
other approaches attempt to hard encode rotation invariance by changing
the network structure itself such that for each learned filter a set
of corresponding rotated versions are automatically generated either
directly based on a predefined set of orientations, \eg \cite{Zhou2017ORN},
or by convolving each learned filter with a basis set of oriented
Gabor filters \cite{Luan2017}.

Other approaches push the idea of injecting priors into their network
design even further by fully hand crafting their network via casting
the kernel selection problem as an invariance maximization problem
based on group theory, \eg \cite{Bruna2013,Oyallon2015,Cohen2017}.
For example, kernels can be chosen such that they maximize invariances
to small deformations and translations for texture recognition \cite{Bruna2013}
or to maximize rotation invariance for object recognition \cite{Oyallon2015}.

\begin{figure}[H]
\begin{centering}
\includegraphics[width=0.75\textwidth,height=0.25\paperheight]{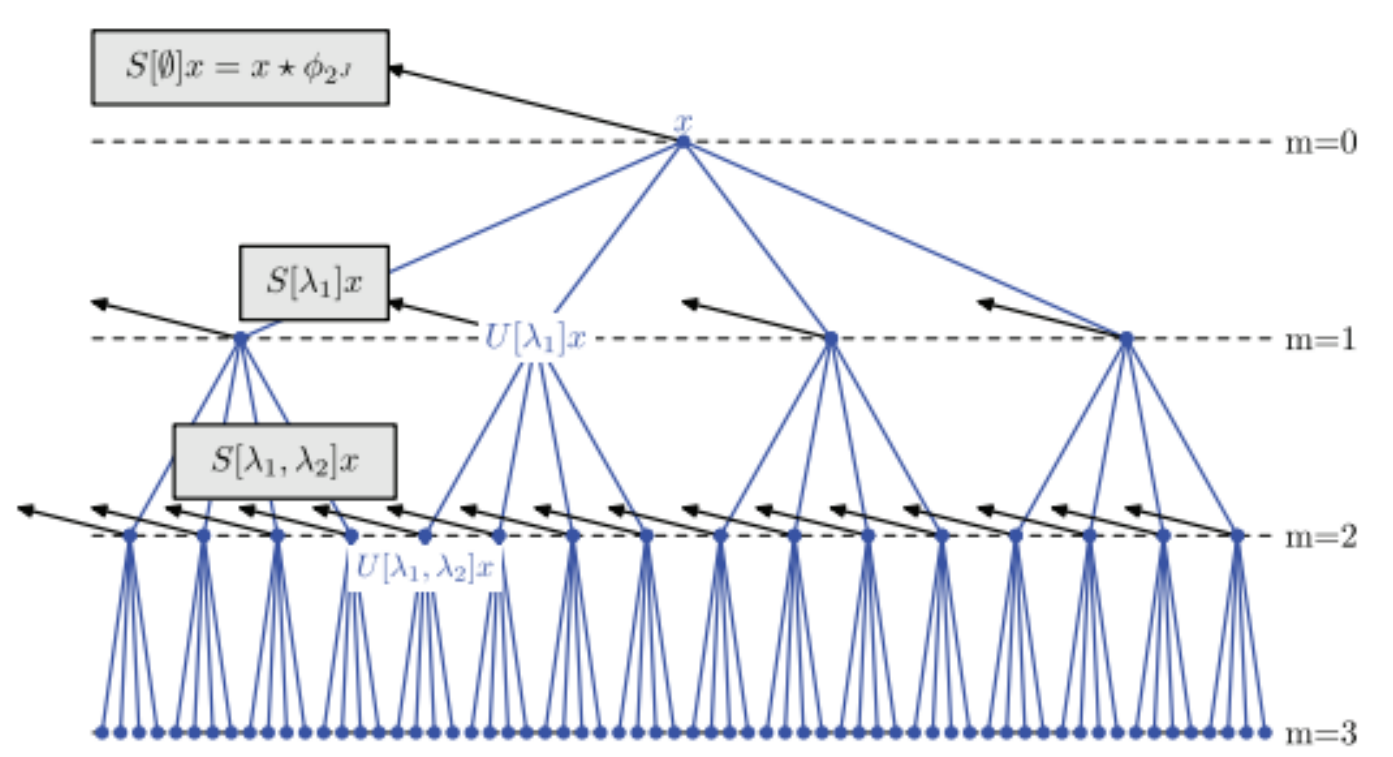}
\par\end{centering}
\caption[Scattering Transform Network.]{\label{fig:mallatnetwork}Scattering Transform Network. In this network,
the scattering transform $S[\lambda]x$ proposed in \cite{Bruna2013}
is applied repeatedly at each layer $m$ on all the outputs $U[\lambda_{i}]x$
from a previous layer. Essentially, the outputs of each layer go through
the same transform over and over again, however, the transform targets
a different effective frequency at each layer and thereby extracts
novel information at each layer. In this figure, an instantiation
of the network with $m=3$ layers is depicted as an illustration.
Figure reproduced from \cite{Bruna2013}.}
\end{figure}

Arguably, the scattering transform network (ScatNet) has one of the
most rigorous mathematical definitions to date \cite{Bruna2013}.
The construction of scattering transforms starts from the assertion
that a good image representation should be invariant to small, local
deformations and various transformation groups depending on the task
at hand. The kernels used in this method are a set of dilated and
rotated wavelets $\psi_{\lambda}$ where $\lambda$ is the frequency
location of the wavelet and it is defined as $\lambda=2^{-j}r$ where
$2^{-j}$ represents the dilation and $r$ represents the rotation.
The network is constructed by a hierarchy of convolutions using various
wavelets centered around different frequencies, as well as various
nonlinearities as discussed in the next section. The frequency locations
of the employed kernels are chosen to be smaller at each layer. The
entire process is summarized in Figure \ref{fig:mallatnetwork}. 

\begin{figure}
\begin{centering}
\includegraphics[width=0.45\textwidth]{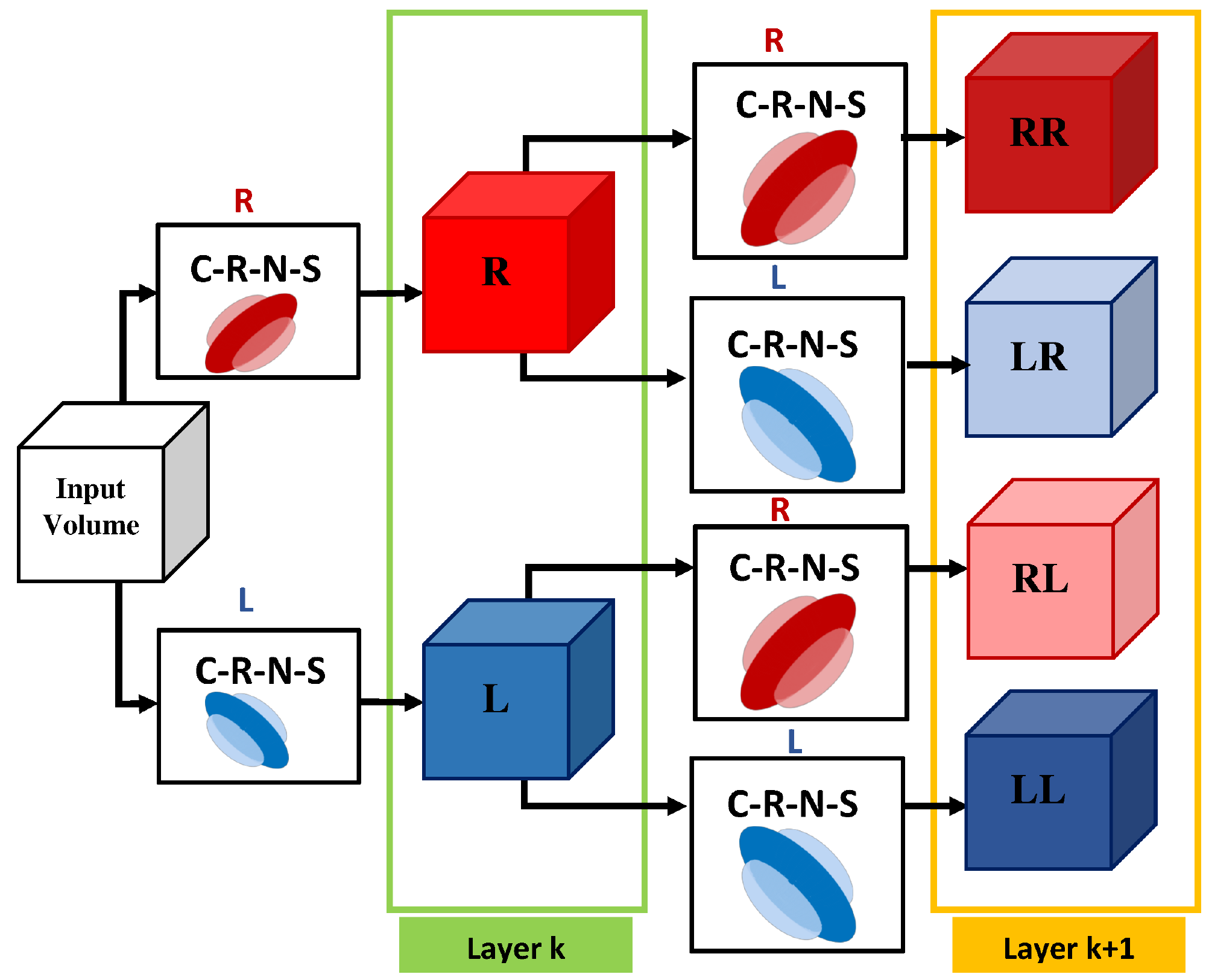}
\par\end{centering}
\caption[Unfolded SOE-Net Architecture.]{\label{fig:soenet}SOE-Net Architecture. Local spatiotemporal features
at various orientations are extracted with an initial processing layer,
$\mathcal{L}_{k}$. $\sf{C}$-$\sf{R}$-$\sf{N}$-$\sf{S}$ indicate Convolution,
Rectification, Normalization and Spatiotemporal pooling, while R and
L indicate rightward vs. leftward filtered data, resp., with symbol
strings (\eg LR) indicating multiple filterings. A network with only
2 filters (\ie 2 orientations) is shown for illustration. Each of
the feature maps at layer $\mathcal{L}_{k}$ is treated as a new separate
signal and fed back to layer $\mathcal{L}_{k+1}$ to be convolved
with the same set of filters but at a different effective resolution
due to spatiotemporal pooling. Figure reproduced from \cite{ISMA2017}.}

\end{figure}

A related ConvNet, dubbed SOE-Net, was proposed for spacetime image
analysis \cite{ISMA2017}. SOE-Net relies on a vocabulary of theory
motivated, analytically defined filters. In particular, its convolutional
block relies on a basis set of 3D oriented Gaussian derivative filters
that are repeatedly applied while following a frequency decreasing
path similar to ScatNet as illustrated in Figure \ref{fig:soenet}.
In this case, however, the network design is cast in terms of spatiotemporal
orientation analysis and invariance is enforced via a multiscale instantiation
of the used basis set.

Loosely speaking both SOE-Net and ScatNet fall under the $Filter\rightarrow Rectify\rightarrow Filter$
paradigm advocated by some biologically based models \cite{baker2001}.
Because these network are based on a rigorous mathematical analysis,
they also take into account the frequency content of the signal as
it is processed in each layer. One of the direct results of this design
is the ability to make theory driven decisions regarding the number
of layers used in the network. In particular, given that outputs of
the different layers of the network are calculated using a frequency
decreasing path, the signal eventually decays. Hence, the iterations
are stopped once there is little energy left in the signal. Further,
through its choice of filters that admit a finite basis set (Gaussian
derivatives), SOE-Net can analytically specify the number of orientations
required.

Another simple, yet powerful, outlook on the kernel selection process
relies on pre-fixed filters learned using PCA \cite{PCAnet2015}.
In this approach, it is argued that PCA can be viewed as the simplest
class of auto-encoders that minimize reconstruction error. The filters
are simply learned using PCA on the entire training dataset. In particular,
for each pixel in each image $X_{i}$, a patch of size $k_{1}\times k_{2}$
is taken and subjected to a de-meaning operation to yield a set of
patches $\bar{X_{i}}$. A collection of such overlapped patches from
each image is stacked together to form the volume $X=[\bar{X}_{1},\bar{X}_{2},...,\bar{X}_{N}]$.
The filters used correspond to the first $L_{1}$ principal eigenvectors
of $XX^{T}$. These vectors are reshaped to form kernels $W_{l}$
of size $k_{1}\times k_{2}$ and convolved with each input image $X_{i}$
to obtain feature maps $I_{i}^{l}$. The same procedure is repeated
for higher layers of the network. 

Compared to ScatNet \cite{Bruna2013} and SOE-Net \cite{ISMA2017},
the PCA approach work is much less mathematically involved and relies
more on learning. However, it is worth highlighting that the most
basic form of auto-encoder was able to achieve respectable results
on several tasks including face recognition, texture recognition and
object recognition. A closely related approach also relies on unsupervised
kernel selection as learned via k-means clustering \cite{DUNDAR2016}.
Once again, although such an approach does not yield state-of-the
art results compared to standard learning based architectures it is
worthy of note that it still is competitive even on heavily researched
datasets such as MNIST \cite{LeCun1998}. More generally, the effectiveness
of such purely unsupervised approaches suggest that there is non-trivial
information that can be leveraged simply from the inherent statistics
of the data.

\subsubsection{Optimal number of kernels}

As previously mentioned, the biggest bottleneck of multilayer architectures
is the learning process that requires massive amounts of training
data mainly due to the large number of parameters to be learned. Therefore,
it is of paramount importance to carefully design the network's architecture
and decide on the number of kernels at each layer. Unfortunately,
even hand-crafted ConvNets usually resort to a random selection of
the number of kernels (\eg\cite{Bruna2013,Oyallon2015,Serre2007,Jhuang2007,PCAnet2015,Fidler2006}).
One exception among the previously discussed analytically defined
ConvNets is SOE-Net, which as previously mentioned, specifies the
number of filters analytically owing to its used of a finite basis
set (\ie oriented Gaussian derivatives).

The recent methods that suggest the use of basis sets to reduce the
number of kernels at each layer \cite{Jacobsen16,Cohen2017} offer
an elegant way of tackling this issue although the choice of the set
of filters and the number of filters in the set is largely based on
empirical considerations. The other most prominent approaches tackling
this issue aim at optimizing the network architecture during the training
process. A simple approach to deal with this optimization problem,
referred to as optimal brain damage \cite{LeCun1990OBD}, is to start
from a reasonable architecture and progressively delete small magnitude
parameters whose deletion does not negatively affect the training
process. A more sophisticated approach \cite{Feng2015} is based on
the Indian Buffet Process \cite{Griffiths05}. The optimal number
of filters is determined by training a network to minimize a loss
function $\mathcal{L}$ that is a combination of three objectives

\begin{equation}
\mathcal{L}=\sum_{l=1}^{L}\mathcal{L}_{task}(.)+\sum_{l=1}^{L_{conv}}\mathcal{L}_{conv}(.)+\sum_{l=L_{conv}+1}^{L}\mathcal{L}_{fc}(.),\label{eq:4-1}
\end{equation}
where $L_{conv}$ is the number of convolutional layers and $L$ is
the total number of layers. In \eqref{eq:4-1}, $\mathcal{L}_{fc}$
and $\mathcal{L}_{conv}$ are the unsupervised loss functions of the
fully connected and convolutional layers, respectively. Their role
is to minimize reconstruction errors and are trained using unlabeled
data. In contrast, $\mathcal{L}_{task}$ is a supervised loss function
designed for the target task and is trained to maximize classification
accuracy using labeled training data. Therefore, the number of filters
$K$ in each layer is tuned by minimizing both a reconstruction error
and a task related loss function. This approach allows the proposed
network to use both labeled and unlabeled data.

In practice, the three loss functions are minimized alternatively.
First, the filter parameters $W^{(l)}$ are fixed and the number of
filters $K^{(l)}$ is learned with a Grow-And-Prune (GAP) algorithm
using all available training data (i.e. labeled and unlabeled). Second,
the filter parameters are updated by minimizing the task specific
loss function using the labeled training data. The GAP algorithm can
be described as a two way greedy algorithm. The forward pass increases
the number of filters. The backward pass reduces the network size
by removing redundant filters.

\subsubsection*{Discussion}

Overall, most theoretically driven approaches to convolutional kernel
selection aim at introducing priors into their hierarchical representations
with the ultimate goal of reducing the need for massive training.
In doing so, these methods either rely on maximizing invariances through
methods grounded in group theory or rely on combinations over basis
sets. Interestingly, similar to more biologically inspired instantiations,
it also is commonly observed that there is a pronounced tendency to
model early layers with filters that have the appearance of oriented
bandpass filters. However, the choice for higher layers' kernels remains
an open critical question.%

\section{Rectification\label{sec:Rectification}}

Multilayer networks are typically highly nonlinear and rectification
is, usually, the first stage of processing that introduces nonlinearities
to the model. Rectification refers to applying a pointwise nonlinearity
(also known as an activation function) to the output of the convolutional
layer. Use of this term borrows from signal processing, wherein rectification
refers to conversion from alternating to direct current. It is another
processing step that finds motivation both from biological and theoretical
point views. Computational neuroscientists introduce the rectification
step in an effort to find the appropriate models that explain best
the neuroscientific data at hand. On the other hand, machine learning
researchers use rectification to obtain models that learn faster and
better. Interestingly, both streams of research tend to agree, not
only on the need for rectification, but they are also converging to
the same type of rectification.

\subsection{Biological perspective}

From a biological perspective, rectification nonlinearities are usually
introduced into the computational models of neurons in order to explain
their firing rates as a function of the input \cite{Dayan2005}. A
fairly well accepted model for biological neuron's firing rate in
general is referred to as the \textit{Leaky Integrate and Fire} (LIF)
\cite{Dayan2005}. This model explains that the incoming signal to
any neuron has to exceed a certain threshold in order for the cell
to fire. Research investigating the cells in the visual cortex in
particular also relies on a similar model, referred to as half wave
rectification \cite{hubel1962,Movshon1978,Heeger91}.

Notably, Hubel and Wiesel's seminal work already presented evidence
that simple cells include nonlinear processing in terms of half wave
rectification following on linear filtering \cite{hubel1962}. As
previously mentioned in Section \ref{sec:The-Convolutional-Layer},
the linear operator itself can be considered as a convolution operation.
It is known that, depending on the input signal, convolution can give
rise to either positive or negative outputs. However, in reality cells'
firing rates are by definition positive. This is the reason why Hubel
and Wiesel suggested a nonlinearity in the form of a clipping operation
that only takes into account the positive responses. More in line
with the LIF model, other research suggested a slightly different
half wave rectification in which the clipping operation happens based
on a certain threshold (\ie other than zero)\cite{Movshon1978}.
Another more complete model also took into account the possible negative
responses that may arise from the filtering operation \cite{Heeger91,heeger1992}.
In this case, the author suggested a two-path half wave rectification
where the positive and negative incoming signals are clipped separately
and carried in two separate paths. Also, in order to deal with the
negative responses both signals are followed by a pointwise squaring
operation and the rectification is therefore dubbed half-squaring
(although biological neurons do not necessarily share this property).
In this model the cells are regarded as energy mechanisms of opposite
phases that encode both the positive and negative outputs. 

\subsubsection*{Discussion}

Notably, these biologically motivated models of neuronal activation
functions have become common practice in today's convolutional network
algorithms and are, in part, responsible for much of their success
as will be discussed next.

\subsection{Theoretical perspective}

\begin{figure}[H]
\begin{centering}
\begin{tabular}{ccc}
\includegraphics[width=0.25\textwidth]{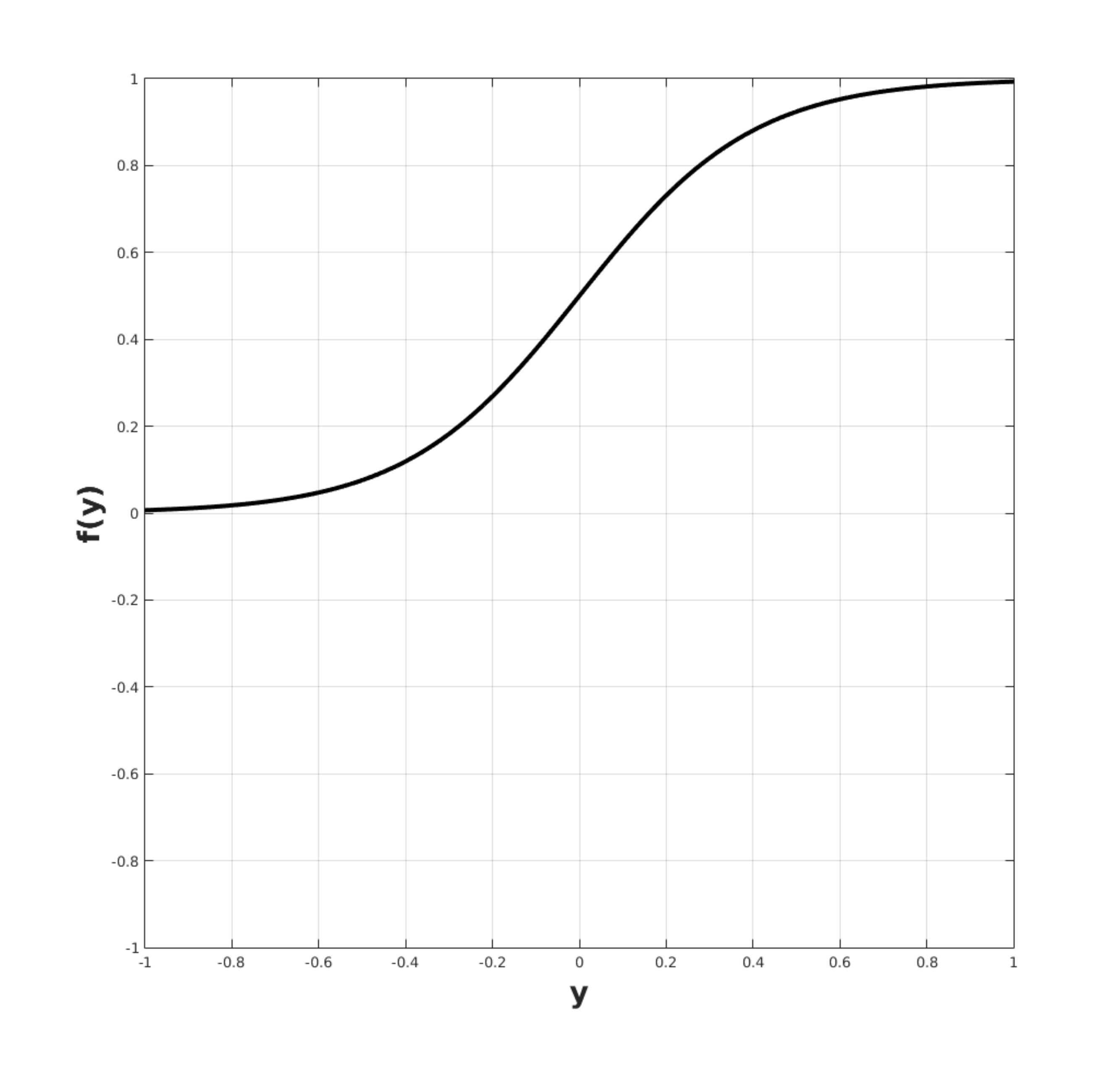} & \includegraphics[width=0.25\textwidth]{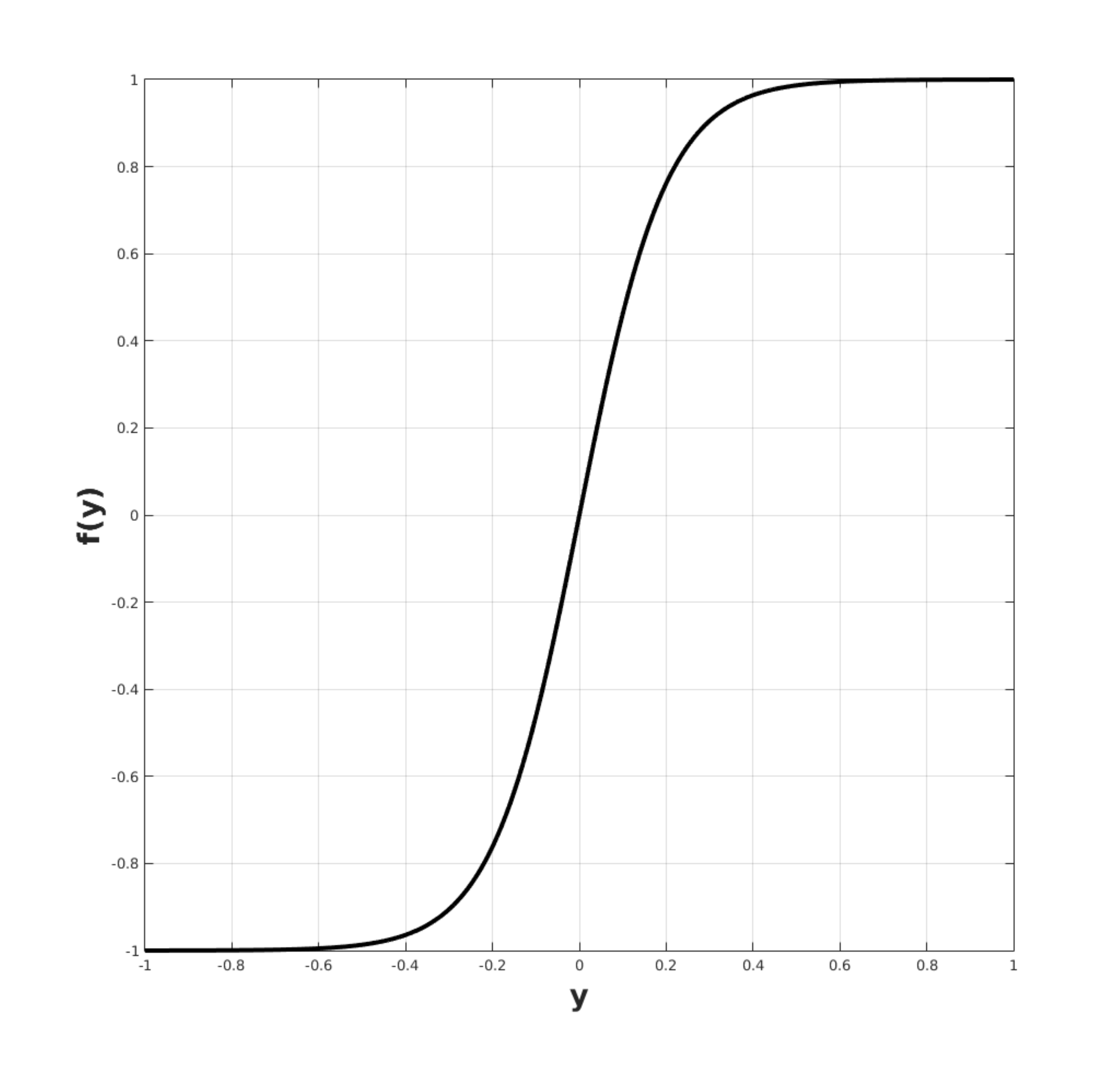} & \includegraphics[width=0.25\textwidth]{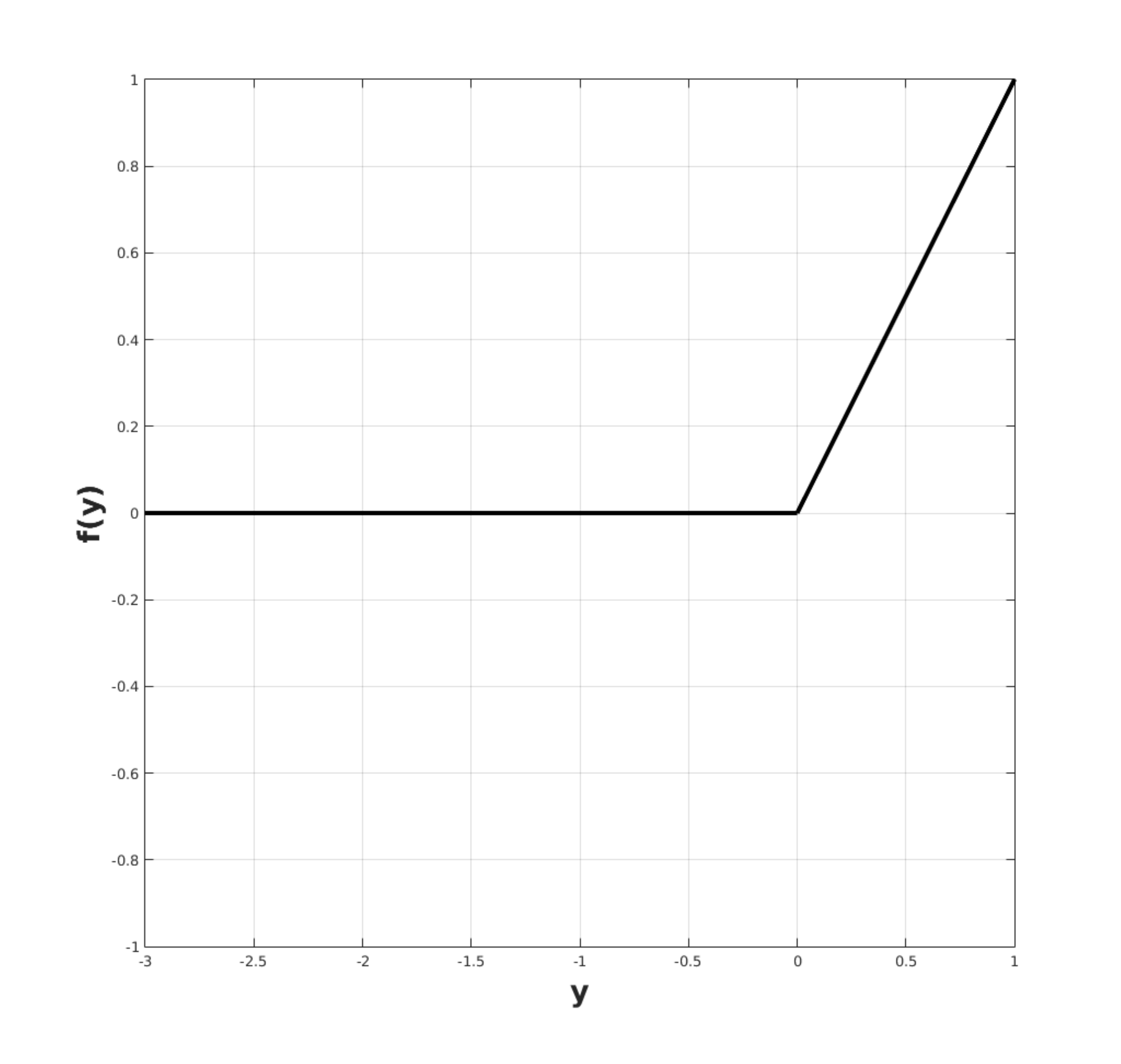}\tabularnewline
(a) Logistic & (b) tanh & (c) ReLU\tabularnewline
\includegraphics[width=0.25\textwidth]{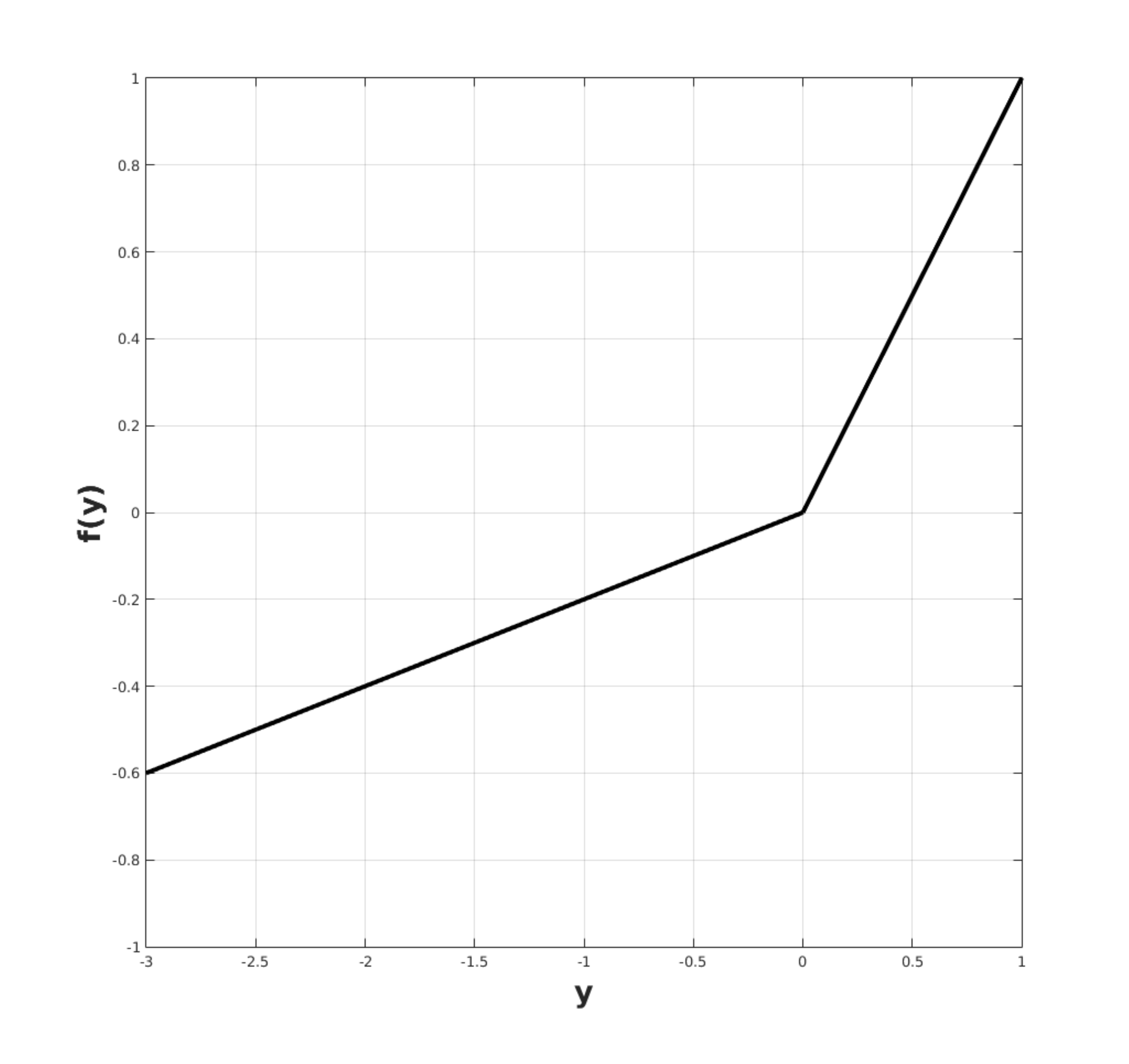} & \includegraphics[width=0.25\textwidth]{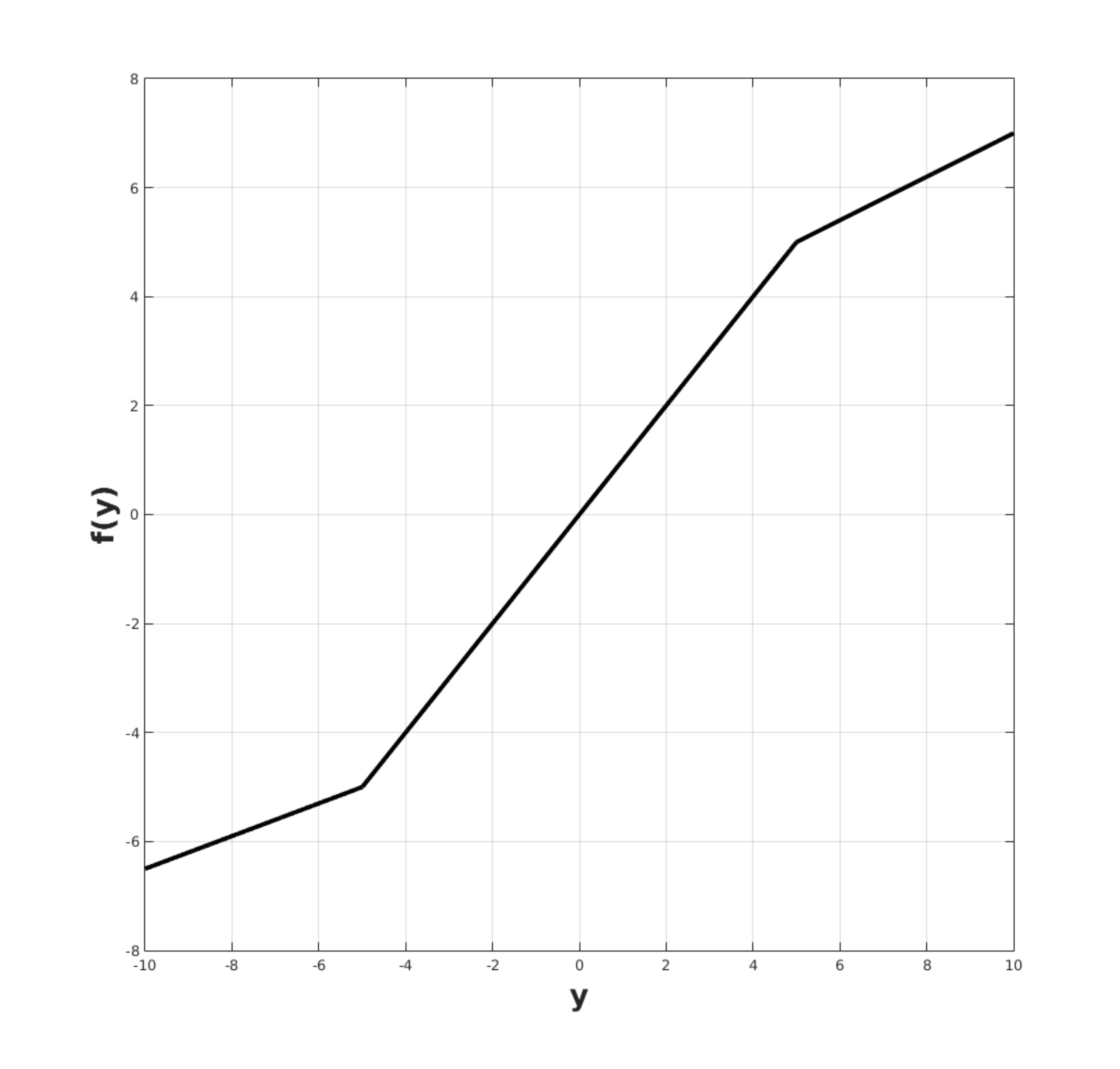} & \includegraphics[width=0.25\textwidth]{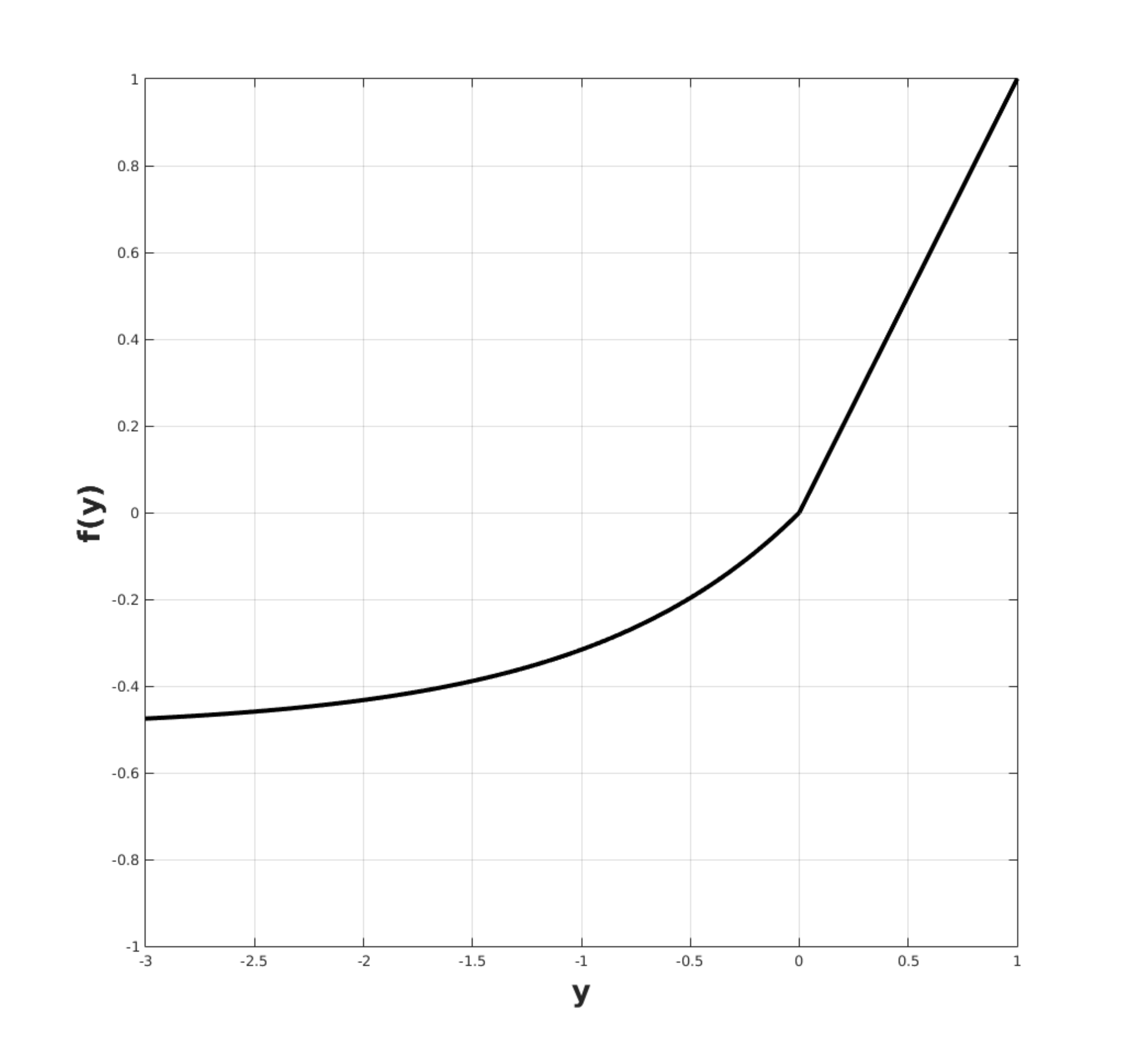}\tabularnewline
(d) LReLU/PReLU & (e) SReLU & (f) EReLU\tabularnewline
\end{tabular}
\par\end{centering}
\caption{\label{fig:rectifications}Nonlinear Rectification Functions Used
in the Multilayer Networks Literature.}
\end{figure}

From a theoretical perspective, rectification is usually introduced
by machine learning researchers for two main reasons. First, it is
used to increase the discriminating power of the extracted features
by allowing the network to learn more complex functions. Second, it
allows for controlling the numerical representation of the data for
faster learning. Historically, multilayer networks relied on pointwise
sigmoidal nonlinearities using either the logistic nonlinearity or
the hyperbolic tangent \cite{LeCun1998}. Although the logistic function
is more biologically plausible given that it does not have a negative
output, the hyperbolic tangent was more often used given that it has
better properties for learning such as a steady state around $0$
(See Figures \ref{fig:rectifications} (a) and (b), respectively).
To account for the negative parts of the hyperbolic tangent activation
function it is usually followed by a modulus operation (also referred
to as \textit{Absolute Value Rectification} AVR) \cite{Jarret2009}.
However, recently the \textit{Rectified Linear Unit} (ReLU), first
introduced by Nair \etal \cite{NairH10}, quickly became the default
rectification nonlinearity in many fields (\eg\cite{Maas2013}) and
particularly computer vision ever since its first successful application
on the ImageNet dataset \cite{Krizhevsky2012}. It was shown in \cite{Krizhevsky2012}
that the ReLU plays a key role against overfitting and expediting
the training procedure, even while leading to better performance compared
to traditional sigmoidal rectification functions.

Mathematically, ReLU is defined as follows,
\begin{equation}
f(y_{i})=max(0,y_{i})\label{eq:relu-1}
\end{equation}
and is depicted in Figure \ref{fig:rectifications} (c). The ReLU
operator has two main desirable properties for any learning based
network. First, ReLU does not saturate for positive input given that
its derivative is $1$ for positive input. This property makes ReLU
particularly attractive since it removes the problem of vanishing
gradients usually present in networks relying on sigmoidal nonlinearities.
Second, given that ReLU sets the output to $0$ when the input is
negative, it introduces sparsity, which has the benefit of faster
training and better classification accuracy. In fact, for improved
classification it is usually desirable to have linearly separable
features and sparse representations are usually more readily separable
\cite{Glorot2011}. However, the hard $0$ saturation on negative
input comes with its own risks. Here, there are two complementary
concerns. First, due to the hard zero activation some parts of the
network may never be trained if the paths to these parts were never
activated. Second, in a degenerate case where all units at a given
layer have a negative input, back propagation might fail and this
will lead to a situation that resembles the vanishing gradient problem.
Because of these potential issues many improvements to the ReLU nonlinearity
have been proposed to deal better with the case of negative outputs
while keeping the advantages of ReLU. 

Variations of the ReLU activation function include the \textit{Leaky
Rectified Linear Unit} (LReLU) \cite{Maas2013} and its closely related
\textit{Parametric Rectified Linear Unit} (PReLU) \cite{He2015} that
are mathematically defined as
\begin{equation}
f(y_{i})=max(0,y_{i})+a_{i}min(0,y_{i})\label{eq:lrelu-1}
\end{equation}
and depicted in Figure \ref{fig:rectifications} (d). In LRelu $a_{i}$
is a fixed value, whereas it is learned in PReLU. LReLU was initially
introduced to avoid zero gradients during back propagation but did
not improve the results of the tested networks significantly. Also,
it heavily relied on cross validation experimentation in the selection
of the parameter $a_{i}$. In contrast, PReLU optimizes the value
of this parameter during training, which leads to a bigger boost in
performance. Notably, one of the most important results of PReLU is
the fact that early layers in the network tended to learn a higher
value for the parameters $a_{i}$, whereas that number is almost negligible
for higher layers in the network's hierarchy. The authors speculate
that this result could be due to the nature of the filters learned
at different layers. In particular, since first layer kernels are
usually oriented bandpass like filters both parts of the response
are kept as they represent a potentially significant difference in
the incoming signal. On the other hand, kernels at higher layers are
tuned to detect specific objects and are trained to be more invariant. 

\begin{figure}[H]
\begin{centering}
\includegraphics[width=0.55\textwidth,height=0.1\paperheight]{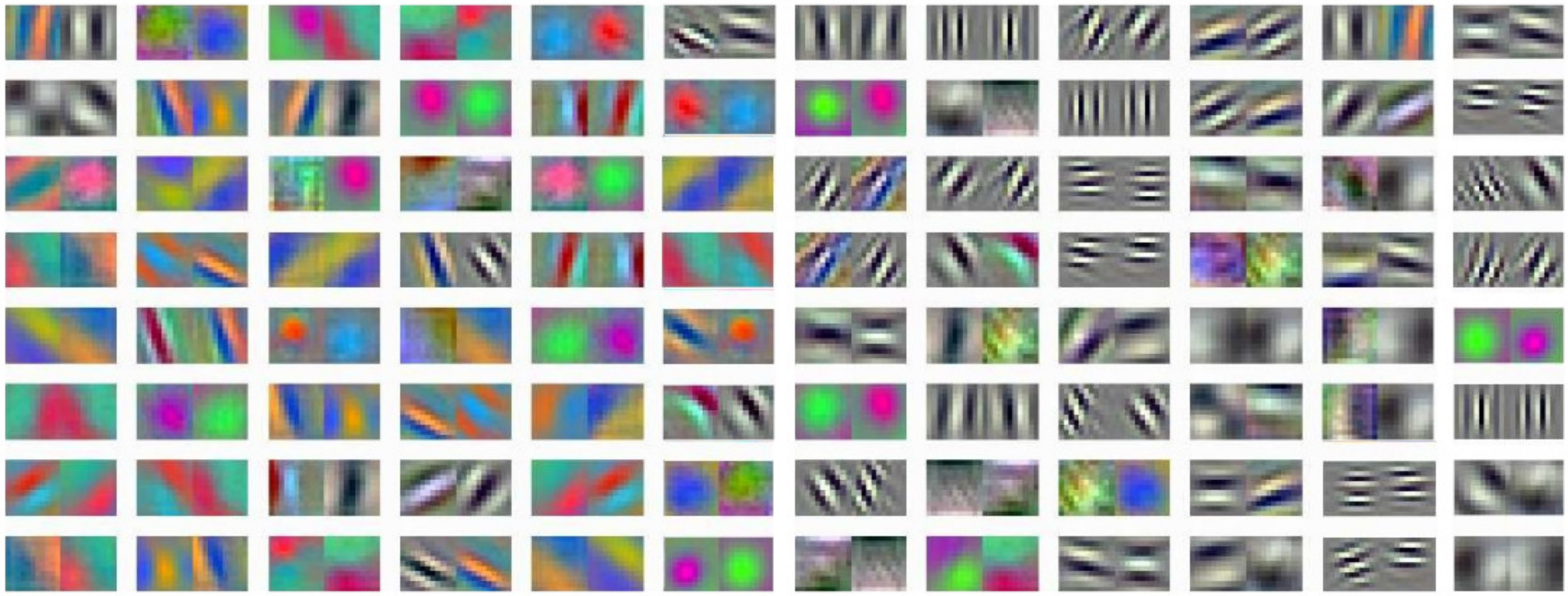}
\par\end{centering}
\caption[Visualization of Conv1 Filters Learned by AlexNet Trained on ImageNet
Dataset.]{\label{fig:CReU-1} Visualization of Conv1 Filters Learned by AlexNet
Trained on ImageNet Dataset. Figure reproduced from \cite{SHANG2016}.}
\end{figure}

Interestingly, another rectification function, dubbed \textit{Concatenated
Rectified Linear Unit} (CReLU), was proposed based on similar observations
\cite{SHANG2016}. In that case, the authors propose CReLU starting
from the observation that kernels learned at the initial layers of
most ConvNets tend to form negatively correlated pairs (\ie filters
that are $180$ degrees out of phase) as shown in Figure \ref{fig:CReU-1}.
This observation implies that the negative responses eliminated by
the ReLU nonlinearity are replaced by learning kernels of opposite
phase. By replacing ReLU with CReLU, the authors were able to demonstrate
that a network designed to encode a two path rectification leads to
a better performance, while reducing the number of parameters to be
learned through removing redundancies.

Other variation in the ReLU family include: the the \textit{S-shaped
Rectified Linear Unit} (SReLU) \cite{Jin2016}, defined as

\begin{equation}
f(y_{i})=\begin{cases}
t_{i}^{r}+a_{i}^{r}(y_{i}-t_{i}^{r}) & if\,y_{i}\geq t_{i}^{r}\\
y_{i} & if\,t_{i}^{r}>y_{i}>t_{i}^{l}\\
t_{i}^{l}+a_{i}^{l}(y_{i}-t_{i}^{l}) & if\,y_{i}\leq t_{i}^{l}
\end{cases}\label{eq:srelu-1}
\end{equation}
and depicted in Figure \ref{fig:rectifications} (e), which was introduced
to allow networks to learn more nonlinear transformations. It consists
of three piecewise linear functions with 4 learnable parameters. The
main downside of SReLU is that it introduces several parameters to
learn (\ie particularly if the parameters are not shared among several
channels), which makes learning more complex. This concern is especially
true given that a bad initialization of those parameters might impair
the learning. Yet another variant is the \textit{Exponential Linear
Unit} (ELU) \cite{Clevert2016}, defined as

\begin{equation}
f(y_{i})=\begin{cases}
y_{i} & if\,x>0\\
\alpha(exp(y_{i})-1) & if\,x\leq0
\end{cases}\label{eq:elu-1}
\end{equation}
and depicted in Figure \ref{fig:rectifications} (f), which is motivated
by a desire to increase invariance to noise by forcing the signal
to saturate to a value controlled by a variable $\alpha$ for negative
inputs. A common thread across all variations in the ReLU family is
that the negative input should be taken into account as well and dealt
with appropriately.

Another outlook on the choice of the rectification nonlinearity is
presented in the Scattering Network \cite{Bruna2013}. As previously
mentioned in Section \ref{sec:The-Convolutional-Layer}, ScatNet is
handcrafted with the main goal of increasing invariance of the representation
to various transformations. Since it broadly relies on Wavelets in
the convolutional layer it is invariant to small deformations; however,
it remains covariant to translation. Therefore, the authors rely on
an integral operation defined as
\begin{equation}
S[\lambda]x(u)=||x\star\psi_{\lambda}||_{1}=\int|x\star\psi_{\lambda}(u)|\,du\label{eq:3}
\end{equation}
and implemented as average pooling, to add a level of shift invariance.
Hence, in anticipation of the subsequent pooling operation that can
drive the response towards zero, \ie in the case where positive and
negative responses cancel each other, the $\mathbf{L}_{1}(\mathbb{\mathbb{R}}^{2})$
norm operator is used in the rectification step to make all responses
positive. Once again, it is worth noting here that traditional ConvNets
that relied on the hyperbolic tangent activation function also used
a similar AVR rectification to handle negative outputs \cite{LeCun1998,Jarret2009}.
Also, more biologically motivated models, such as the half-squaring
rectification \cite{Heeger91,heeger1992}, relied on pointwise squaring
of the signal to deal with negative responses. This squaring operation
also allows for reasoning about the responses in terms of energy mechanisms.
Interestingly, one of the recent more theory driven convolutional
networks \cite{ISMA2017} also proposed a a two path rectification
strategy defined as

\begin{equation}
\begin{array}{c}
E^{+}(\mathbf{x};\theta_{i},\sigma_{j})=\left(\max[C(\mathbf{x};{\displaystyle \theta_{i},\sigma_{j}),0]}\right)^{2}\\
E^{-}(\mathbf{x};\theta_{i},\sigma_{j})=\left(\min[C(\mathbf{x};{\displaystyle \theta_{i},\sigma_{j}),0]}\right)^{2}
\end{array},\label{eq:rectification}
\end{equation}
where $C(\mathbf{x};{\displaystyle \theta_{i},\sigma_{j})}$ is the
output from the convolution operation. This rectification strategy
combines the idea of keeping both phases of the filtered signal and
pointwise squaring and thereby allows for conservation the signals
magnitude and phase while considering the resulting signal in terms
of spectral energy.

\subsubsection*{Discussion}

Interestingly, the broad class of ReLU nonlinearities clearly became
the most popular choice for the rectification stage from a theoretical
perspective. Notably, the choice of completely neglecting the negative
inputs (\ie as done in ReLU) seems to be more questionable as evidenced
by the many contributions proposing alternatives to this choice \cite{Maas2013,He2015,Jin2016,Clevert2016,SHANG2016}.
It is also important to compare the behavior of the ReLU with AVR
rectification used in ScatNet \cite{Bruna2013} and older ConvNet
architectures \cite{Jarret2009}. While AVR preserves the energy information
but erases the phase information, ReLU on the other hand keeps the
phase information in some sense, by retaining the positive parts of
the signal only; however, it does not preserve the energy as it throws
away half of the signal. Significantly, methods that try to preserve
both (\eg CReLU \cite{SHANG2016} and the use of \eqref{eq:rectification}
in SOE-Net \cite{ISMA2017}) were able to achieve better performances
across several tasks, and such methods are also in consensus with
biological findings \cite{Heeger91}.

\section{Normalization\label{sec:Normalization}}

As previously mentioned, multilayer architectures are highly nonlinear
due to the cascade of nonlinear operations that take place in these
networks. In addition to the rectification nonlinearity discussed
in the previous section, normalization is another nonlinear block
of processing that plays a significant role in ConvNet architectures.
The most widely used form of normalization used in ConvNets is the
so called \textit{Divisive Normalization} or DN (also known as local
response normalization). This section sheds light on the role of the
normalization step and describes how it corrects for some of the shortcomings
of the previous two blocks of processing (\ie Convolution and Rectification).
Once again the role of normalization will be discussed both from biological
and theoretical perspectives.

\subsection{Biological perspective}

Normalization was proposed early on by neurophysiologists to explain
the phenomenon of light adaptation in the retina \cite{Boynton1970}
and was later extended to explain the nonlinear properties of neurons
in the mammalian visual cortex \cite{Heeger91}. Indeed, from a biological
point of view, the need for a normalization step stems from two main
observations \cite{heeger1992,Heeger91}. First, although cells responses
were proven to be stimulus specific \cite{hubel1962}, it was also
shown that cell responses can inhibit one another and that there exists
a phenomenon of cross-orientation suppression, where the response
of a neuron to its preferred stimuli is attenuated if it is superimposed
with another ineffective stimuli \cite{heeger1992,Brouwer2011,carandini-and-heeger-2012}.
Neither the linear models (\ie in the convolution step) nor the different
forms of rectification discussed in the previous section, such as
half-wave rectification proposed by computational neuroscientists,
explain this cross-orientation suppression and inhibition behavior.
Second, while cell responses are known to saturate at high contrast,
a model relying only on convolution and unbounded rectifiers, such
as ReLU, will have values that keep increasing with increasing contrast.
These two observations suggested the need for a step that discounts
the responses of other stimuli in order to keep the specificity of
each cell and make it contrast invariant while explaining other inhibition
behaviors of cells. 

One popular model to deal with these issues includes a divisive normalization
block described mathematically as follows

\begin{equation}
\bar{E_{i}}=\frac{E_{i}}{\sigma^{2}+\sum_{j}E_{j}},\label{eq:norm_bio}
\end{equation}
where $E_{i}$ is the output of a squared, half wave rectified convolution
operation, pooled over a set of orientations and scales $j$ and $\sigma^{2}$
is a saturation constant that can be chosen based on either one of
two adaptation mechanisms \cite{Heeger91}. In the first case, it
could be a different value for each cell learned from the cell's response
history. The second possibility is to derive it from the statistics
of the responses of all cells. This divisive normalization scheme
discards information about magnitude of the contrast in favor of encoding
the underlying image pattern in terms of relative contrast across
the input responses, $E_{j}$, in the normalization operation, \eqref{eq:norm_bio}.
Use of this model seemed to provide a good fit to neuron responses
of mammalian visual cortex \cite{heeger1992}. It was also shown that
it explains well the cross-orientation suppression phenomenon as well
\cite{Brouwer2011}.

\subsubsection*{Discussion}

Interestingly, most of the studies investigating the role of divisive
normalization show that neuronal models including it fit well the
recorded data (\eg \cite{Heeger91,heeger1992,Brouwer2011,carandini-and-heeger-2012}).
Indeed, more recent studies suggest that divisive normalization could
also explain the phenomenon of adaptation in IT cortex where the neural
responses decrease with stimulus repetition (\eg \cite{kaliukhovich2016}).
Moreover, the suggested prevalence of divisive normalization in several
areas of the cortex lead to the hypothesis that Divisive Normalization
can be seen as a canonical operation of the mammalian visual cortex
similar to the operation of convolution \cite{carandini-and-heeger-2012}.

\subsection{Theoretical perspective}

From a theoretical perspective, normalization has been explained as
being a method of achieving efficient coding when representing natural
images \cite{simoncelli2008}. In that work, the normalization step
was motivated by findings regarding the statistics of natural images
\cite{simoncelli2008} that are known to be highly correlated and
for containing very redundant information. In light of these findings,
the normalization step was introduced with the goal of finding a representation
that minimizes statistical dependencies in images. To achieve this
goal a popular derivation discussed thoroughly in \cite{simoncelli2008,Lyu2010}
starts by representing images using a statistical model based on a
Gaussian Scale Mixture. Using this model and an objective function
whose role is to minimize dependencies, a nonlinearity is derived
in the form of 

\begin{equation}
r_{i}=\frac{{\displaystyle x_{i}-}\sum_{j}a_{j}x_{j}}{\sqrt{b+\sum_{k}c_{j}(x_{j}-\sum_{k}a_{k}x_{k})^{2}}}\label{eq:div_norm}
\end{equation}
where $x_{i}$ and $r_{i}$ are the input and output images, respectively,
while $b$, $a_{i}$ and $c_{i}$ are parameters of the divisive normalization
that can be learned from a training set. Notably, there exists a direct
relationship between the definition of the divisive normalization
introduced to deal with redundancies and high order dependencies in
natural images, \eqref{eq:div_norm}, and that suggested to best fit
neuron responses in the visual cortex, \eqref{eq:norm_bio}. In particular,
with a change of variable where we set $y_{i}={\displaystyle x_{i}-\sum_{j}a_{j}x_{j}}$,
we see that the two equations are related, subject to the square root
difference, by an elementwise operation, (\ie squaring, with $E_{i}=y_{i}^{2}$),
and thereby both models achieve the goal of maximizing independence
while satisfying neuroscientific observations.

Another way of looking at normalization in ConvNets in particular
is to consider it as a way of enforcing local competition between
features \cite{Jarret2009,LeCun1998}, similar to the one taking place
in biological neurons. This competition can be enforced between adjacent
features within a feature map through subtractive normalization or
between feature maps through divisive normalization operating at the
same spatial locations across feature maps. Alternatively, divisive
normalization can be seen as a way of minimizing sensitivity to multiplicative
contrast variation \cite{ISMA2017}. It was also found, on deeper
network architectures, that divisive normalization was helpful in
increasing the generalization power of a network \cite{Krizhevsky2012}.

More recent ConvNets rely on what is referred to as \textit{batch
normalization} \cite{IoffeS15}. Batch normalization is another kind
of divisive normalization that takes into account a batch of the training
data to learn normalization parameters (\ie the mean and variance
in equation \eqref{eq:batch_norm}) and it also introduces new hyperparameters,
$\gamma^{(k)}$and $\beta^{(k)}$, to control the amount of normalization
needed at each layer.

Batch normalization can be summarized in two steps. First, at any
layer with a $d$-dimensional input $x=(x^{(1)}...\,x^{(d)})$, each
scalar feature is independently normalized according to

\begin{equation}
\widehat{x}^{(k)}=\frac{x^{(k)}-E[x^{(k)}]}{\sqrt{Var[x^{(k)}]}},\label{eq:batch_norm}
\end{equation}
with $E[x^{(k)}]$ being the mini-batch mean calculated as $E[x^{(k)}]=\frac{1}{m}\sum_{i=1}^{m}x_{i}$
over the $m$ samples of the mini-batch, and $Var[x^{(k)}]$ is the
variance of the same mini-batch calculated as $Var[x^{(k)}]=\frac{1}{m}\sum_{i=1}^{m}(x_{i}-E[x^{(k)}])^{2}$.
Second, the output of the normalization in equation \eqref{eq:batch_norm}
is subject to a linear transformation such that the final output of
the proposed batch normalization block is given by $y^{(k)}=\gamma^{(k)}\widehat{x}^{(k)}+\beta^{(k)}$,
with $\gamma^{(k)}$and $\beta^{(k)}$ being hyperparameters to be
learned during training. 

The first step in batch normalization aims at fixing the means and
variances of the inputs at each layer. However, since that normalization
strategy can change or limit what a layer can represent, the second
linear transformation step is included to maintain the network's representational
power. For example, if the original distribution at the input was
already optimal, then the network can restore it by learning an identity
mapping. Therefore, the normalized inputs, $\widehat{x}^{(k)}$, can
be thought of as being inputs to a linear block added at each layer
of a network.

Batch normalization was first introduced as an improvement to traditional
divisive normalization with the ultimate goal of reducing the problem
of \textit{internal covariate shift}, which refers to the continuous
change of the distribution of inputs at each layer \cite{IoffeS15}.
The changing scale and distribution of inputs at each layer implies
that the network has to significantly adapt its parameters at each
layer and thereby training has to be slow (\ie use of small learning
rate) for the loss to keep decreasing during training (\ie to avoid
divergence during training). Therefore, batch normalization was introduced
to guarantee more regular distributions at all inputs. 

This normalization strategy was inspired by general rules of thumb
established for efficient training of ConvNets. In particular, for
good generalization performance in ConvNets it is common practice
to enforce that all training and testing set samples have the same
distribution (\ie through normalization). For example, it has been
shown that networks converge faster when the input is always whitened
\cite{LeCun1998,Jarret2009}. Batch normalization builds on this idea
by considering that each layer can be considered as a shallow network.
Therefore, it would be advantageous to make sure that the inputs keep
the same distribution at each layer and this is enforced by learning
the distribution of the training data (using mini-batches) and using
the statistics of the training set to normalize each input. More generally,
it is also important to remember that, from a machine learning perspective,
such a normalization scheme can also make features easier to classify.
For example, if two different inputs induce two different outputs,
they are more easily separable by a classifier if the responses lie
within the same range and it is therefore important to process the
data to satisfy this condition. 

Similar to divisive normalization, batch normalization also proved
to play an important role in ConvNets. In particular, it has been
shown that batch normalization not only speeds up training, but it
also plays a non-trivial role in generalization where it was able
to outperform previous state-of-the-art on image classification (on
ImageNet in particular) while removing the need for Dropout regularization
\cite{Krizhevsky2012}.

In comparison, batch normalization is somewhat similar to divisive
normalization in the sense that they both make the scale of the inputs
at each layer similar. However, Divisive Normalization normalizes
the values for each input by dividing it by all other inputs at the
same location within the same layer. Batch normalization, on the other
hand, normalizes each input with respect to statistics of the training
set at the same location (or more accurately of the statistics of
a mini-batch containing examples from the entire training set). The
fact that batch normalization relies on the statistics of a training
set may explain the fact that it improves the generalization power
of the representation.

One problem with batch normalization is its dependence on the mini-batch
size: It might not properly represent the training set at each iteration,
if it is chosen to be too small; alternatively, it can have a negative
effect of slowing down training, if it is too big (\ie since the
network has to see all training samples under the current weights
to calculate the mini-batch statistics). Also, batch normalization
is not easily applicable to recurrent neural networks since it relies
on statistics calculated over a mini-batch of training samples. For
this reason, layer normalization has been proposed in \cite{BaKH16}.
Layer normalization follows the same procedure proposed in batch normalization
and the only difference lies in the way normalization statistics are
calculated. While batch normalization calculates statistics over a
mini-batch, layer normalization calculates statistics for each input
separately using all feature maps or hidden units within any one layer.
Consequently, in batch normalization each unit is normalized with
different statistics relevant to that unit only, whereas layer norm
normalizes all units in the same way. 

While layer norm was shown to be effective on language related applications
where recurrent networks are usually more suitable, it failed to compete
with ConvNets trained with batch normalization for image processing
tasks \cite{IoffeS15}. One possible explanation proposed by the authors
is that in ConvNets all units do not make an equal contribution in
the activation of a unit at the output; therefore, the underlying
assumption that this is the case in layer normalization (\ie using
all units to calculate the statistics of the normalization) does not
hold for ConvNets.

\subsubsection*{Discussion}

The common thread across the contributions discussed in this subsection
is the fact that they all agree on the important role of normalization
in improving the representational power of multilayer architectures.
Another important point to note is that they all share the same goal
of reducing redundancies in the input as well as bringing it to the
same scale even while casting the problem under different forms. Indeed,
while early proposals of divisive normalization, \eg \cite{simoncelli2008},
explicitly cast the problem as a redundancy reduction problem, newer
proposals such as batch normalization \cite{IoffeS15} are also implicitly
enforcing this operation through whitening of the data at each layer.
Finally, reflecting back on the normalization problem from a biological
perspective, it is important to note that biological systems are also
efficient in encoding the statistical properties of natural signals
in the sense that they represent the world with small codes. Therefore,
one might hypothesize that they are also performing similar operations
of divisive normalization to reduce redundancies and obtain those
efficient codes.

\section{Pooling\label{sec:Pooling}}

Virtually any ConvNet model, be it biologically inspired, purely learning
based or completely hand-crafted, includes a pooling step. The goal
of the pooling operation is to bring a level of invariance to changes
in position and scale as well as to aggregate responses within and
across feature maps. Similar to the three building blocks of ConvNets
discussed in the previous sections, pooling is also supported by biological
findings as well as more theory driven investigations. The major debate
when it comes to this layer of processing in convolutional networks
is on the choice of the pooling function. The two most widely encountered
variations are average and max pooling. This section explores the
advantages and shortcomings of each and discusses other variations
as described in related literature.

\subsection{Biological perspective}

\begin{figure}[H]
\begin{centering}
\includegraphics[width=0.5\textwidth]{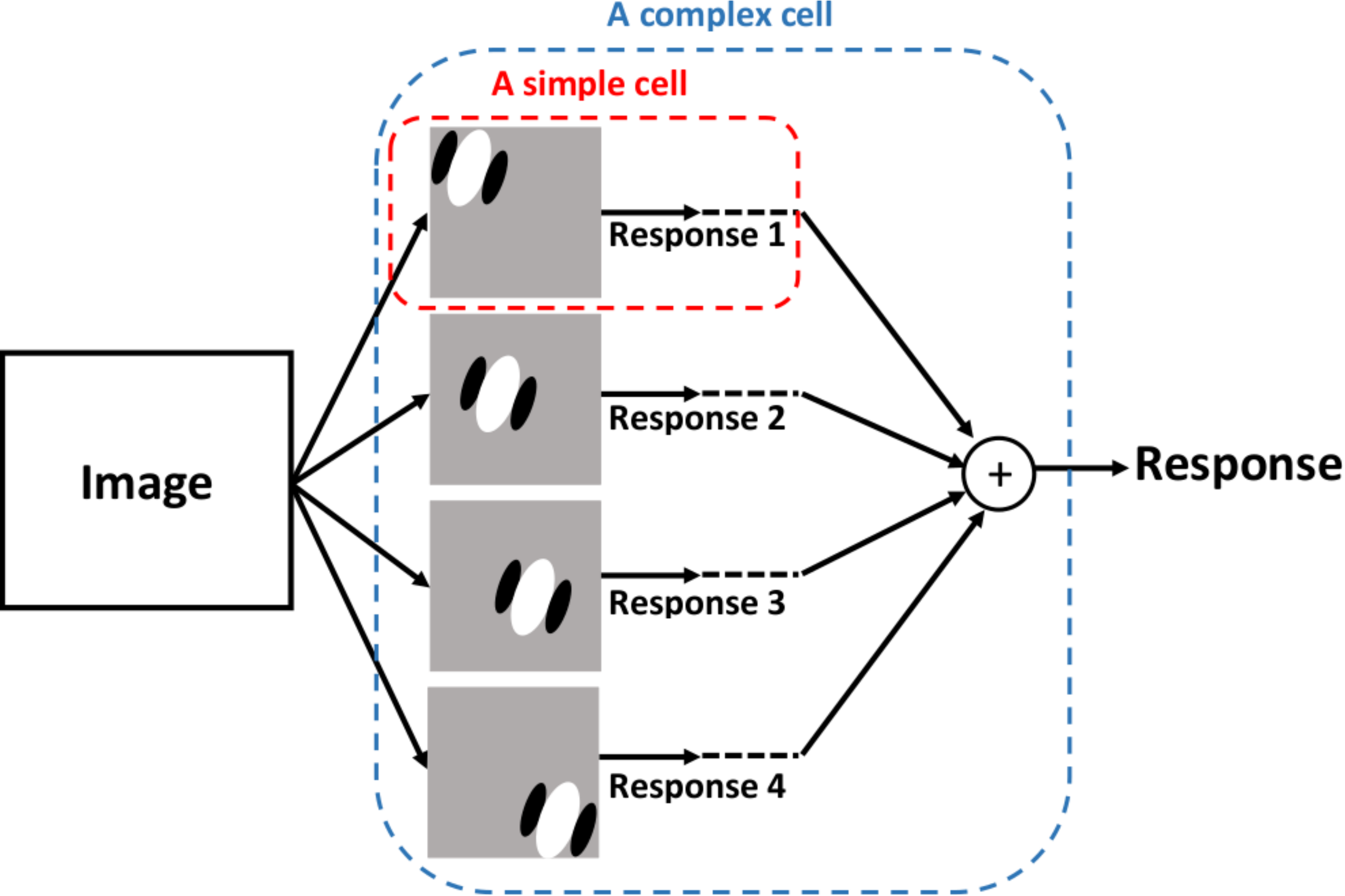}
\par\end{centering}
\caption[An Illustration of the Difference Between Simple and Complex Cells.]{\label{fig:simple-vs-complex} An Illustration of the Difference
Between Simple and Complex Cells. This illustration suggests that
complex cells responses result from combinations of simple cells responses.
\vspace{-10pt}}
\end{figure}

From a biological perspective, pooling is largely motivated by the
behavior of the cortical complex cells \cite{hubel1962,Movshon1978,heeger1992,carandini2006}.
In their seminal work, Hubel and Wiesel \cite{hubel1962} found that,
just like simple cells, complex cells are also tuned to a specific
orientation, but as opposed to simple cells, complex cells exhibit
a level of position invariance. They suggest that this result is achieved
through some sort of pooling in which the responses of simple cells
tuned to the same orientation%
{} are pooled across space and/or time as illustrated in Figure \ref{fig:simple-vs-complex}.

\begin{figure}[H]
\begin{centering}
\begin{tabular}{c}
\includegraphics[width=0.6\textwidth,height=0.1\paperheight]{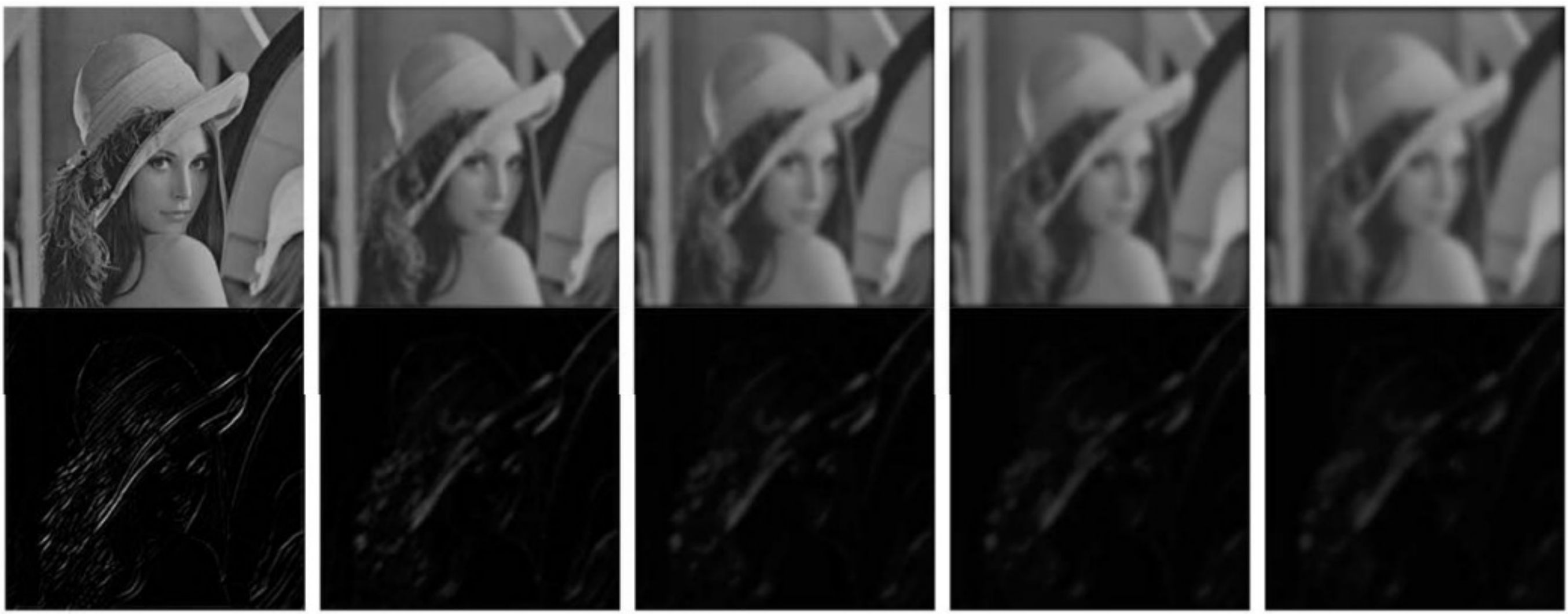}\tabularnewline
(a)\tabularnewline
\includegraphics[width=0.6\textwidth,height=0.1\paperheight]{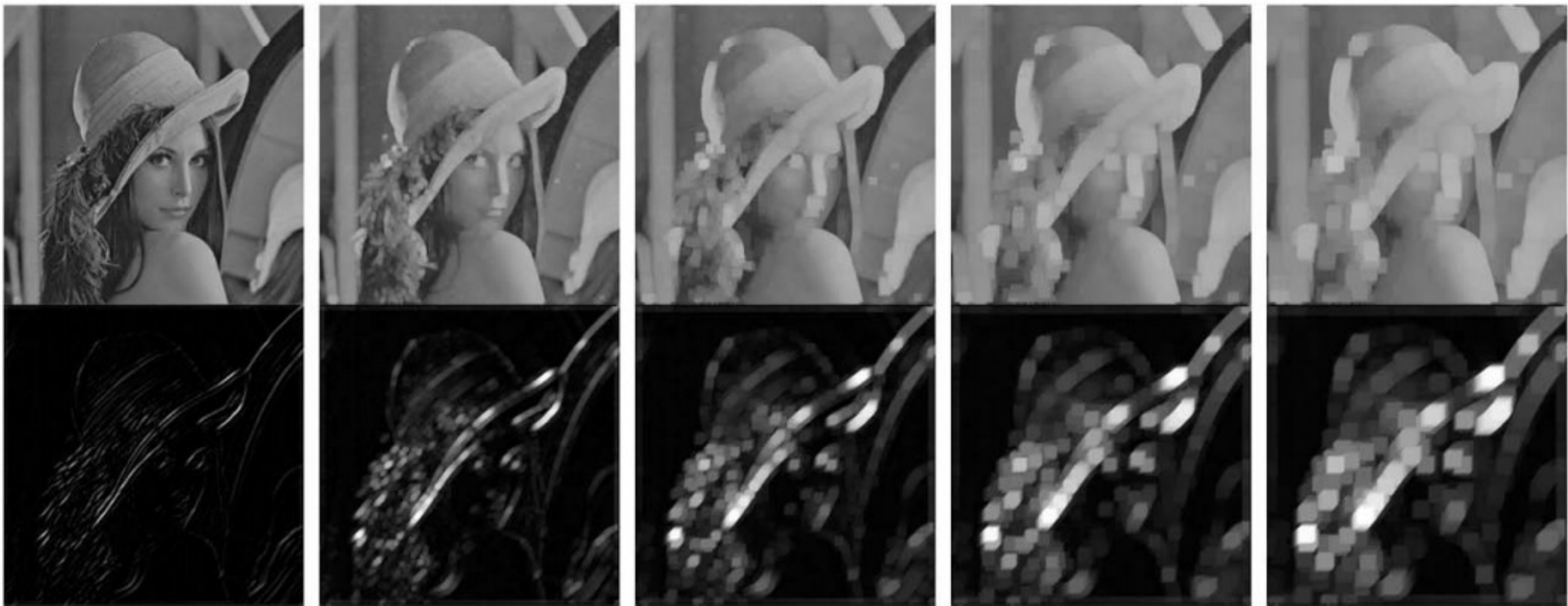}\tabularnewline
(b)\tabularnewline
\end{tabular}
\par\end{centering}
\caption[Average versus Max Pooling on Gabor Filtered Images.]{\label{fig:serremax}Average versus Max Pooling on Gabor Filtered
Images. This example illustrates the effect of average pooling at
various scales when applied to ((a) top row) an original gray-value
image and ((a) bottom row) its Gabor filtered version. While average
pooling leads to smoother versions of the gray-value image, the sparse
Gabor-filtered image fades away. In contrast, the example also illustrates
the effect of max pooling at various scales when applied to the same
gray-value image ((b) top row) and ((b) bottom row) its Gabor filtered
version. Here, max pooling causes the gray-values image to degrade
while the sparse edges in the Gabor filtered version are enhanced.
Figure reproduced from \cite{Serre2007}.\vspace{-10pt}}
\end{figure}

Some of the early biologically inspired convolutional networks such
as Fukushima's neocognitron \cite{Fukushima1980} and the original
LeNet network \cite{LeCun1998} relied on average pooling. In these
efforts, average pooling followed by sub-sampling is largely motivated
by the findings of Hubel and Wiesel and it is used to decrease the
network's sensitivity to position changes. On the other hand, the
HMAX \cite{Riesenhuber99} class of networks (\eg \cite{serre2005,MutchLowe2006,Serre2007,Jhuang2007})
rely on max pooling instead. Supporters of the max pooling strategy
claim that it is more plausible when the input to the pooling operator
is a set of Gabor filtered images (\ie the typical model for simple
cells). In fact, the authors argue that while a Gaussian scale space
(\ie similar to weighted average pooling) reveals new structures
at different scales when applied to a natural image, it causes features
to fade away when applied to a Gabor filtered image; see Figure \ref{fig:serremax}(a).
On the other hand, a max pooling operation enhances the strongest
responses in the filtered image at different scales as shown in Figure
\ref{fig:serremax}(b).

\begin{figure}[H]
\begin{centering}
\includegraphics[height=0.1\paperheight]{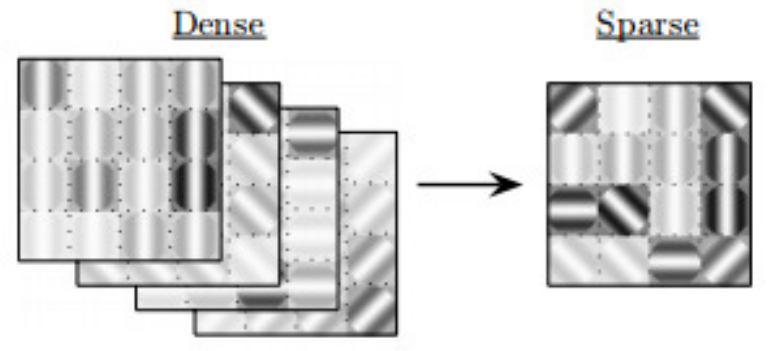}
\par\end{centering}
\caption[Cross-Channel Pooling Illustration.]{\label{fig:lowepooling}Cross-Channel Pooling Illustration. (left)
Dense simple cell responses resulting from filtering operations with
Gabor filters tuned to various orientations (4 orientations are shown
here for illustration purposes) (right) Sparsified simple cell responses
resulting from cross-channel pooling using the max operator (\ie
for each pixel location, the maximum response across feature maps
is kept). Figure reproduced from \cite{MutchLowe2006}.\vspace{-10pt}}
\end{figure}

The behavior of complex cells can also be viewed as a type of cross-channel
pooling, which is in turn another method of injecting invariances
into the representation. Cross channel pooling is achieved through
the combination of outputs from various filtering operations at a
previous layer. This idea was proposed by Mutch and Lowe \cite{MutchLowe2006}
as an extension to one of the most prominent biologically inspired
networks \cite{Serre2007}, previously introduced in Section \ref{sec:The-Convolutional-Layer}
and illustrated in Figure \ref{fig:serrenetwork}. In particular,
the authors introduce max pooling across channels at the second layer
of their network where the output of the S2 simple cells are pooled
across multiple orientations to keep the maximum responding unit at
each spatial position as illustrated in Figure \ref{fig:lowepooling}.

\subsubsection*{Discussion}

Overall, based on the description of complex cells, it seems that
from a biological perspective both average and max pooling are plausible,
although there is more work arguing in favor of average pooling. Independently
from the choice of the pooling operator, the fact is that there is
general agreement on the existence and significance of pooling. A
probably more important question lies in the choice of the receptive
field or the units over which pooling is performed. This aspect of
the pooling operation is further explored in more theory driven work,
as will be described in the next section.\vspace{-10pt}

\subsection{Theoretical perspective}

Pooling has been a component of the computer vision representational
pipelines for some time, \eg \cite{lowe04,Dalal05histogramsof,Lazebnik2006,Fukushima1980,LeCun1998},
with the goal of introducing some level of invariance to image transformations
and better robustness to noise and clutter. From a theoretical perspective,
probably one of the most influential works discussing the importance
and role of pooling was Koendrink's concept of locally orderless images
\cite{koenderink1999}. This work argued in favor of pooling whereby
the exact position of pixels within a Region Of Interest (ROI), \ie
a pooling region, can be neglected even while conserving the global
image structure. Currently, virtually all convolutional architectures
include a pooling block as part of its processing stages. As with
biologically motivated models, more theory driven approaches typically
employ either average or max pooling.

Recent work approaching their network's design from a purely theory
based perspective, \eg ScatNet \cite{Bruna2013} and SOE-Net \cite{ISMA2017},
rely on a form of average pooling. In particular, their networks rely
on a weighted sum pooling operation. These approaches tackle the pooling
problem from a frequency domain point of view; therefore, their choice
of average pooling is motivated by a desire to keep track of the frequency
content of the signal. Average pooling allows these networks to act
on different frequencies at each layer while downsampling the images
to increase invariance and reduce redundancies. At the same time their
controlled approach to specifying pooling parameters allows them to
avoid aliasing during the pooling operation. Notably, in SOE-Net's
investigation, the superiority of weighted average pooling was empirically
demonstrated over both simple box car pooling and max pooling.

Interestingly, most of the early convolutional architectures relied
on average pooling as well, \eg \cite{Fukushima1980,LeCun1998},
but it has slowly fallen out of favor in many learning based convolutional
architectures and been replaced by max pooling. This trend has been
mainly driven by small differences in performance. However, the role
of pooling in a network is significant and needs more careful consideration.
In fact, early work exploring the role of pooling \cite{Jarret2009}
demonstrated that the type of pooling plays such a significant role
in a ConvNet architecture that even an otherwise randomly initialized
network yielded competitive results on the task of object recognition
provided the appropriate type of pooling is used. In particular, this
work compared average and max pooling and demonstrated that with a
randomly initialized network average pooling yields superior performance. 

Other work more systematically compared average and max pooling empirically
\cite{Scherer2010} and suggested that there exists a complementarity
between the two types of pooling depending on the input type and the
transformations it undergoes. Therefore, this work implied that ConvNets
can benefit from using more than one pooling option throughout the
architecture. Yet other work considered the question from a purely
theoretical perspective \cite{Boureau2010}. Specifically, this work
examined the effect of average versus max pooling on the separability
of extracted features. The main conclusions of this paper can be summarized
in two points. First, the authors argue that max pooling is more suitable
when the pooled features are very sparse (\eg when pooling is preceded
by a ReLU). Second, the authors suggest that the pooling cardinality
should increase with the input size and that the pooling cardinality
affects the pooling function. More generally, it was shown that beyond
the pooling type, the pooling size plays an important role as well. 

The importance of the pooling cardinality was also explored in various
other studies, albeit empirically \cite{JiaHD12,Coats2011}. Indeed,
the role of pooling cardinality was first discussed in the context
of the earlier hand-crafted feature extraction pipeline \cite{JiaHD12}.
In particular, this work builds on the spatial pyramid pooling \cite{Lazebnik2006}
encoding method while highlighting the shortcoming of using predetermined
fixed-size pooling grids. The authors suggest learning the pooling
windows' sizes as part of the classifier training. More specifically,
the authors suggest randomly picking various pooling regions of different
cardinalities and training the classifier to pick the pooling region
that yields the highest accuracy. The main motivation behind this
learning based strategy is to make pooling adaptive to the dataset.
For example, the optimal pooling regions for an outdoor scene may
lie along the horizon, which does not necessarily apply to indoor
scenes. Similarly, for video action recognition it proved more perspicuous
to adapt the pooling region to the most salient parts of a video \cite{chris2015}.
The role of the pooling window size or cardinality was also directly
explored in a neural network context \cite{Coats2011}. Here, the
authors suggest that features that are most similar should be pooled
together. The authors propose finetuning the pooling support (\ie
pooling regions) of their network in an unsupervised manner. In particular,
pooling windows are chosen to group together similar features according
to a pairwise similarity matrix, where the similarity measure is squared
correlation. Beyond average and max pooling operations, the common
thread across these investigations is the importance of the pooling
region independently from the pooling function.

Other work approaches the choice of pooling and its corresponding
parameters from a pure machine learning point of view \cite{zeiler2013,Goodfellow2013,minlin14}.
From this perspective, pooling is advocated as a regularization technique
that allows for varying the network's structure during training. In
particular, pooling allows for the creation of sub-models within the
big architecture thanks to the variation of pathways that a back propagated
signal may take during training. These variations are achieved with
methods such as stochastic pooling \cite{zeiler2013} or cross-channel
pooling used in the maxout network \cite{Goodfellow2013} and Network
in Network (NiN) \cite{minlin14}. NiN was first introduced as a way
to deal with overfitting and correct for the over-complete representation
of ConvNets \cite{minlin14} . In particular, due to the large number
of kernels used at each layer, it was noticed that many networks often
end up learning redundant filters after training. Therefore, NiN is
introduced to reduce redundancies at each layer by training the network
to learn which feature maps to combine using a weighted linear combination.
Similar to NiN, the Maxout network \cite{Goodfellow2013} introduces
cross channel pooling wherein the output is set as the maximum across
$k$ feature maps on a channelwise basis. Notably, a recent proposal
also relied on cross channel pooling to minimize redundancies \cite{ISMA2017}
even while being completely learning free. In this work, the network
is based on a fixed vocabulary of filters and cross channel pooling
is designed to group together feature maps resulting from filtering
operations with the same kernel. Beyond minimizing redundancies, this
approach was adopted to allow the network size to remain manageable,
while maintaining interpretability.

Stochastic Pooling (SP) \cite{zeiler2013} was also introduced as
a regularization technique. However, different from maxout and NiN,
which perform cross channel pooling, SP acts within a feature map.
In particular, stochastic pooling is inspired from the dropout technique
that is widely used in fully connected layers, but SP is applied to
convolutional layers instead. It relies on introducing stochasticity
to the pooling operation that forces the back propagated signal to
randomly take different pathways at each iteration during training.
The method starts by normalizing feature map responses, $a_{i}$,
within each region to be pooled, $R_{j}$, as

\begin{equation}
p_{i}=\frac{a_{i}}{\sum_{k\in R_{j}}a_{k}}.
\end{equation}
The normalized values, $p_{i}$, are then used as the probabilities
of a multinomial distribution, which is in turn used to sample a location
$l$ within the region to be pooled. The corresponding activation
$a_{l}$ is the pooled value. Importantly, although stochastic pooling
relies on selecting one value from any region $R_{j}$ (\ie similar
to max pooling), the pooled value is not necessarily the largest in
$R_{j}$. Here, it is important to note that a different pooling strategy
is adopted during testing. At test time, the probabilities are no
longer used to sample a location during pooling; instead, they are
used as the weights of a weighted sum pooling operation. Hence stochastic
pooling is closer in spirit to max pooling during training and closer
to average pooling during testing. The authors argue that the adopted
pooling strategy during training allows for creating different models
thanks to varying pathways, while the pooling used during testing
allows for creating a rough average approximation over all possible
models seen during training. In summary, stochastic pooling can be
seen as an attempt to take the best of both average and max pooling.

Another approach that attempts to achieve a balance between average
and max pooling suggests letting the network learn the optimal pooling
method \cite{lee16}. This idea of multiple pooling strategies is
motivated by experiments demonstrating that the choice of optimal
pooling strategy is affected by the input \cite{Scherer2010}. In
particular, the authors propose three different methods of combining
the benefits of average and max pooling, namely; mixed, gated and
tree pooling. Mixed pooling combines average and max pooling independently
from the region to be pooled, where the network is trained to learn
the mixing proportion according to

\begin{equation}
f_{mix}(\mathbf{x})=a_{l}f_{max}(\mathbf{x})+(1-a_{l})f_{avg}(\mathbf{x}),
\end{equation}
subject to the constraint $a_{l}\in[0,1]$. In gated max-average pooling
the mixing proportion is adaptive to the region to be pooled. In particular,
the network is trained to learn a gating mask, $\mathbf{w}$, that
is applied to the input data via pointwise multiplication. Using this
gating mask, the mixing function is now defined as

\begin{equation}
f_{mix}(\mathbf{x})=\sigma(\mathbf{w}^{T}\mathbf{x})f_{max}(\mathbf{x})+(1-\sigma(\mathbf{w}^{T}\mathbf{x}))f_{avg}(\mathbf{x}),
\end{equation}
with $\sigma(\mathbf{w}^{T}\mathbf{x})=\frac{1}{(1+exp(-\mathbf{w}^{T}\mathbf{x}))}$.

\begin{figure}[H]
\begin{centering}
\includegraphics[width=0.3\textwidth]{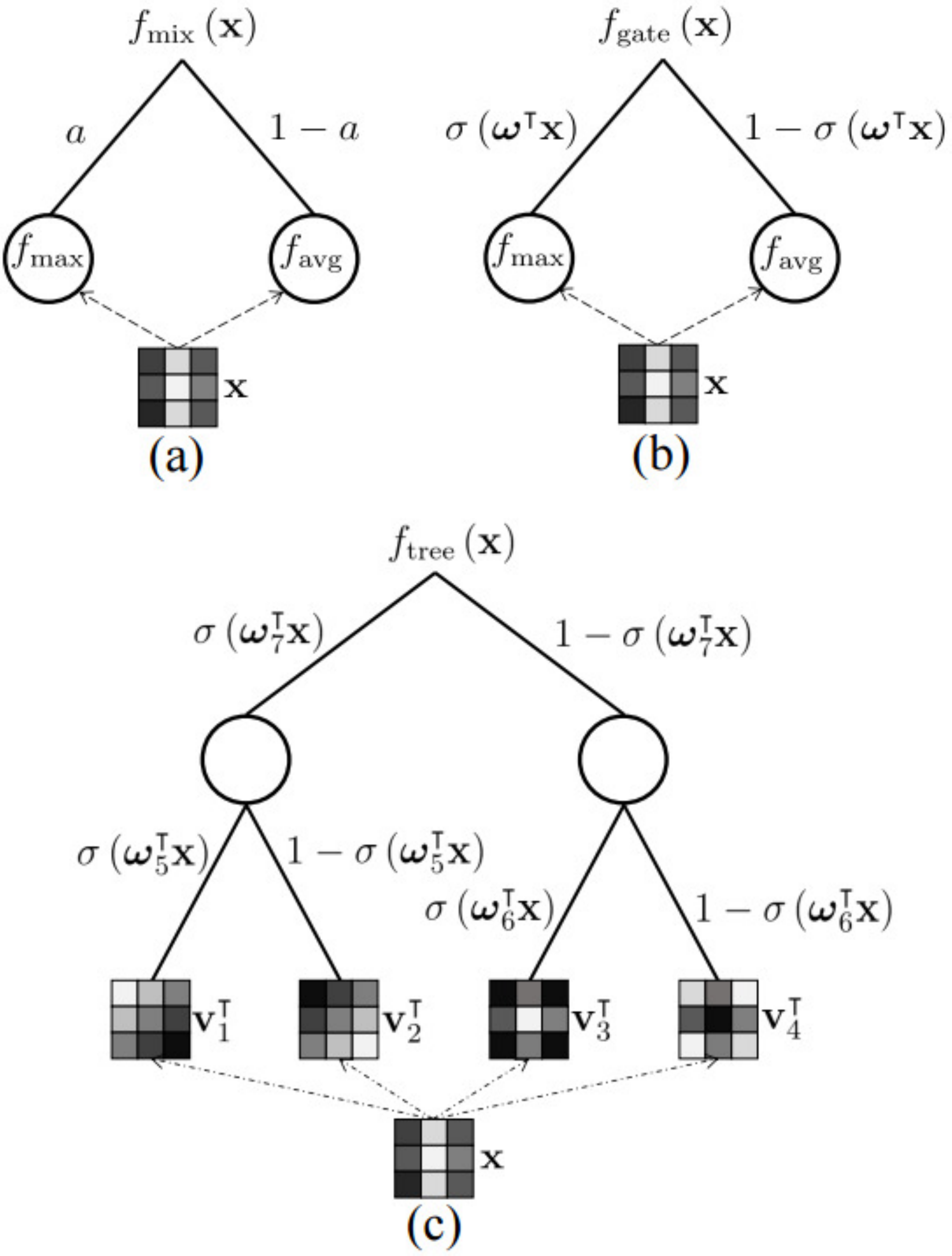}
\par\end{centering}
\caption[Mixed, Gated and Tree Pooling.]{\label{fig:mixed-gated-tree-pooling}Mixed, Gated and Tree Pooling.
Illustration of the described (a) mixed max-average pooling, (b) gated
max-average pooling and (c) tree pooling. Figure reproduced from \cite{lee16}.}
\end{figure}

The third pooling strategy proposed in this work is tree pooling,
which can be viewed as an extreme version of gated pooling. In tree
pooling, not only the mixing proportions are learned but the pooling
functions to be combined are learned as well. Specifically, a tree
structure is adopted to learn the parameters of the individual functions
and their mixing strategy as well. The difference between the three
pooling methods is illustrated in Figure \ref{fig:mixed-gated-tree-pooling}.
In sum, the main idea behind these proposals is letting the pooling
strategy adapt to the region being pooled. Following this strategy,
the authors were able to demonstrate the value of not only combining
average and max pooling but also that of adapting the pooling function
to the region to be pooled. 

\begin{figure}[H]
\begin{centering}
\includegraphics[width=0.7\textwidth]{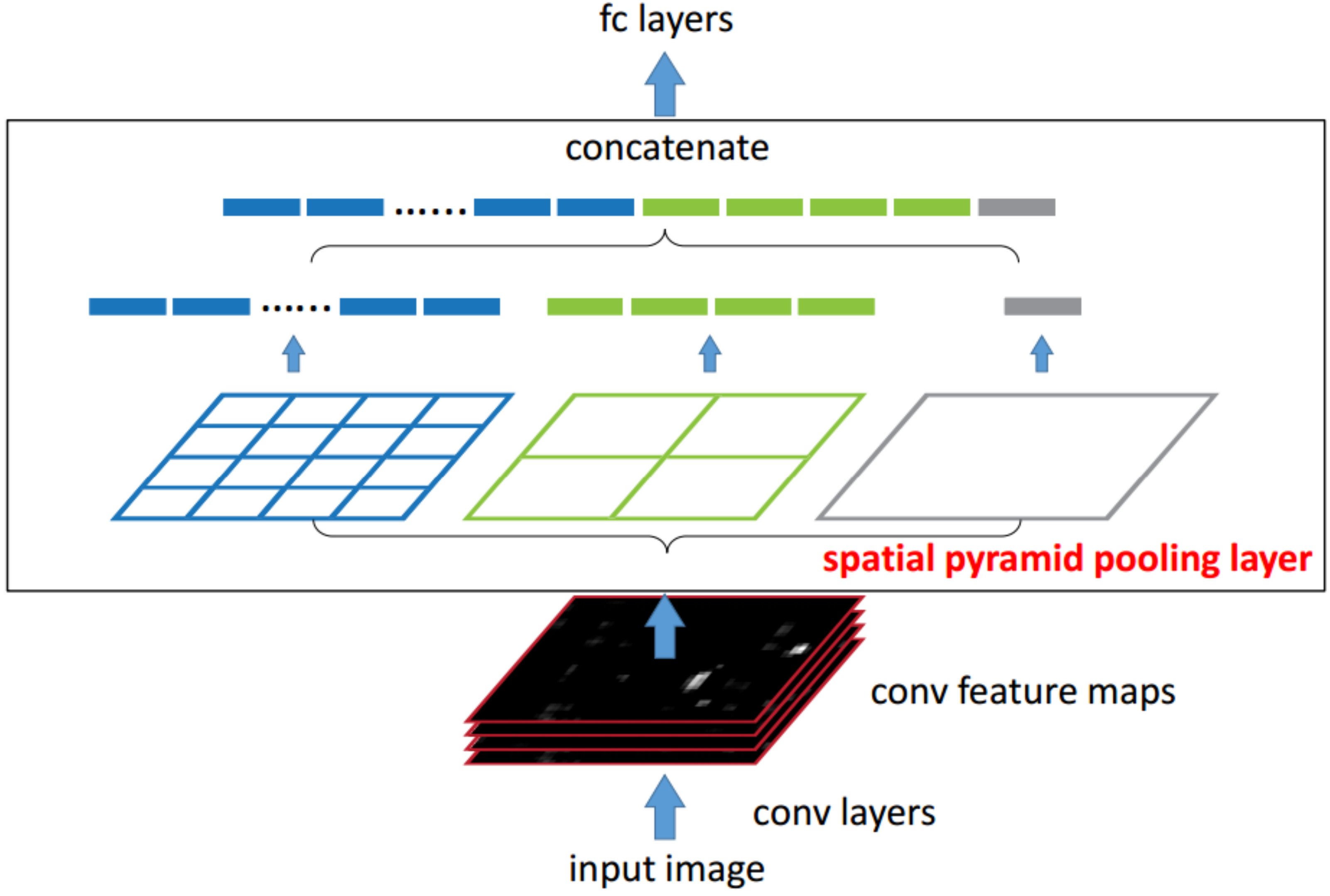}
\par\end{centering}
\caption[Spatial Pyramid Pooling Network.]{\label{fig:spp-net}Spatial Pyramid Pooling Network. SPP is applied
to the feature maps of the last convolutional layer of a network.
Because the spatial bins are proportional to the image size, SPP generates
feature vectors of the same size independently of the input image
size. Hence SPP-Net does not require input images to be pre-processed
such that they are of the same size. Figure reproduced from \cite{He2014}. }
\end{figure}

Finally, it is worth mentioning under this section one last type of
pooling, referred to as global pooling. Global pooling has been used
in some prominent ConvNet models in an effort to deal with more practical
issues relevant to ConvNet architecture design \cite{He2014,minlin14}.
For example, it is known that standard ConvNets rely on convolutional
layers for feature learning/extraction and fully connected layers
followed by a softmax for classification. However, fully connected
layers entail the use of a large number of parameters and are thereby
prone to overfitting. Many methods were introduced to deal with overfitting
induced by fully connected layers, perhaps the most widely used of
which is dropout \cite{Krizhevsky2012}. However, a more elegant way
that fits naturally in a convolutional framework was introduced in
NiN \cite{minlin14} and it is called global average pooling. It simply
relies on aggregating the last layer features across the entire feature
map support. Another example of reliance on global pooling is also
found in the so called SPP-Net \cite{He2014}. In this work, Spatial
Pyramid Pooling (SPP) \cite{Lazebnik2006}, %
{} is used to enable convolutional networks to accept input images of
any size. In fact, ConvNets require fixed size input due to the use
of fully connected layers. SPP-Net introduces spatial pyramid pooling
after the last convolutional layer to correct for this difficulty.
In particular, spatial pyramid pooling is used to generate a fixed
size representation independently from the size of the input image
as illustrated in Figure \ref{fig:spp-net}. Notably, global average
pooling used in NiN, is akin to performing spatial pyramid pooling
at the last layer of the ConvNet where the pyramid consists of only
the coarsest level.

\subsubsection*{Discussion}

Traditionally, the default functions used in pooling have relied on
either the average or max operators. However, several investigations
revealed a certain complementary between the two showing that more
parameters should be taken into account when choosing the pooling
operation. Due to such observations, recent research has been pushing
to extend the idea of training to include learning the pooling functions
and their parameters. However, this direction entails an increase
in the number of parameters to be learned and thereby more chances
of overfitting. Importantly, this approach is to be taken with caution
as it would likely further obscure our knowledge and understanding
of the learned representations. In complement, pooling parameters
can be specified on a theoretical basis for cases where previous stages
of processing have adequate analytic detail. Overall, pooling should
be viewed as a way to summarize information from multiple features
into a compact form that preserves the important aspects of the signal
while discarding details. Beyond deciding how to summarize the features,
it is clear that the harder problem is to determine what constitutes
data that should be pooled and where that data is present.

\section{Overall discussion}

This chapter discussed the role and importance of the most widely
used building blocks in a typical ConvNet architecture in an effort
to understand the working principles of ConvNets. In particular, the
details of each block were addressed from both biological and theoretical
perspectives. Overall, various common threads emerge from the exploration
of the discussed building blocks. In particular, it appears that all
blocks find relatively strong motivations from the operations taking
place in the visual cortex. Further, although all blocks play a significant
role in ConvNets, it appears that the selection of the convolutional
kernels is the most important aspect, as evidenced by the larger body
of literature tackling this block. More importantly, it seems that
more recent ConvNet architectures discussed throughout this chapter
(\eg \cite{Bruna2013,Jacobsen16,Cohen2017,Worral2017,ISMA2017})
are aiming at minimizing the need for heavy training based solutions
by incorporating more controlled building blocks at various stages
of their networks. These recent approaches are in turn motivated by
various efforts that revealed the sub-optimality of the learning based
ConvNets (\eg predominant redundancies in some of the widely used
learned ConvNets) via layerwise visualization and ablation studies,
as will be discussed in the next chapter.

\chapter{Current Status\label{cha:chapter4}}

The review of the role of the various components of ConvNet architectures
emphasized the importance of the convolutional block, which is largely
responsible for most abstractions captured by the network. In contrast,
this component remains the least understood block of processing given
that it entails the heaviest learning. This chapter reviews the current
trends in attempting to understand what is being learned at various
ConvNet layers. In light of these trends, various critical open problems
that remain will be highlighted.

\section{Current trends}

While various ConvNet models continue to push the state-of-the-art
further in several computer vision applications, understanding of
how and why these systems work so well is limited. This question has
sparked the interest of various researchers and in response several
approaches are emerging as ways of understanding ConvNets. In general,
these approaches can be divided into three tacks: those that rely
on visualizations of the learned filters and the extracted feature
maps, those that rely on ablation studies as inspired from biological
approaches to understanding the visual cortex and those that rely
on minimizing learning by introducing analytic principles into their
network's design. Each of these approaches will be briefly reviewed
in this section. 

\subsection{Understanding ConvNets via visualization}

Although several methods have been proposed in the literature for
visualizing the feature maps extracted by ConvNets, in this section
we will focus on the two most prominent approaches and discuss their
different variations. 

\begin{figure}[H]
\begin{centering}
\begin{tabular}{cc}
\includegraphics[width=0.5\textwidth]{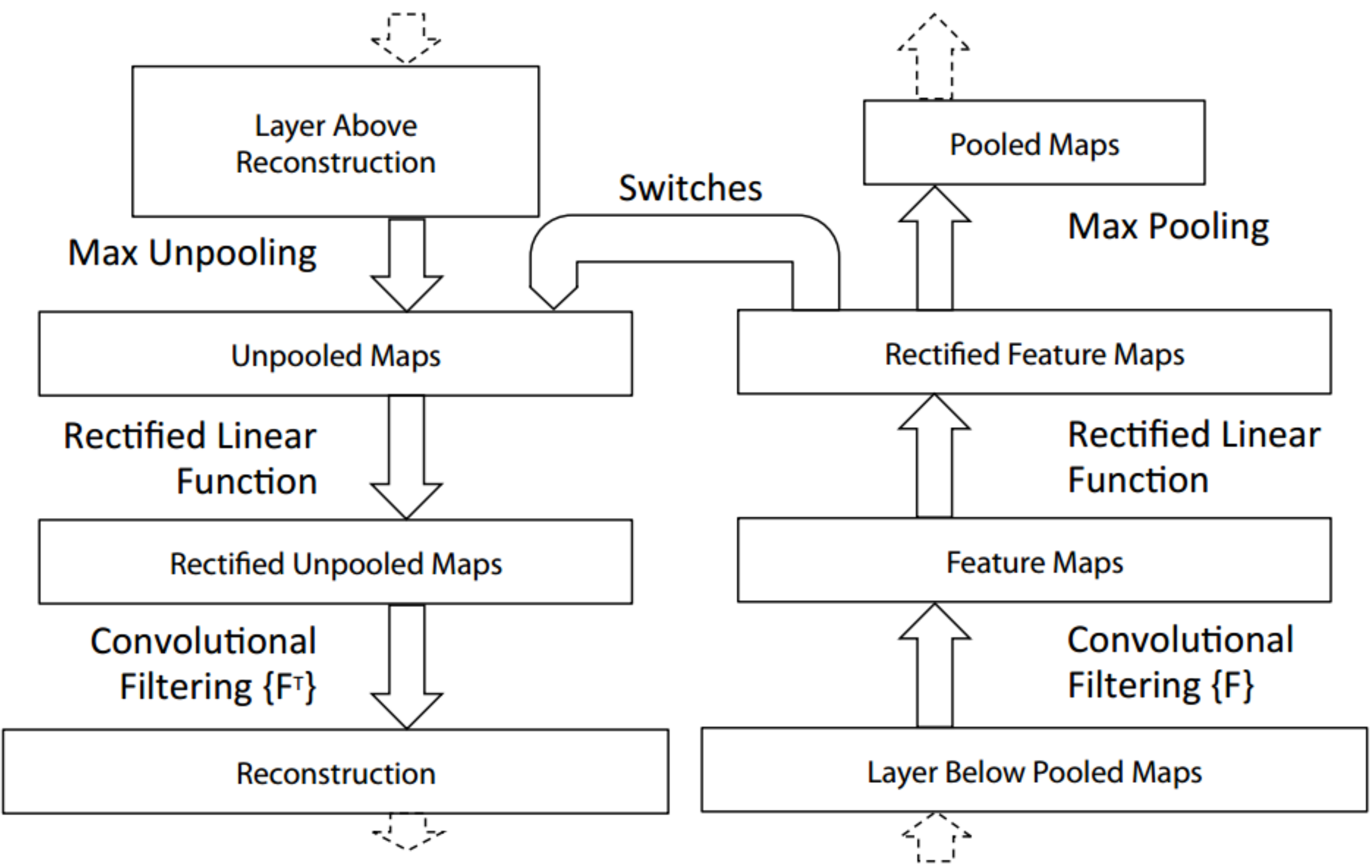} & \includegraphics[width=0.4\textwidth]{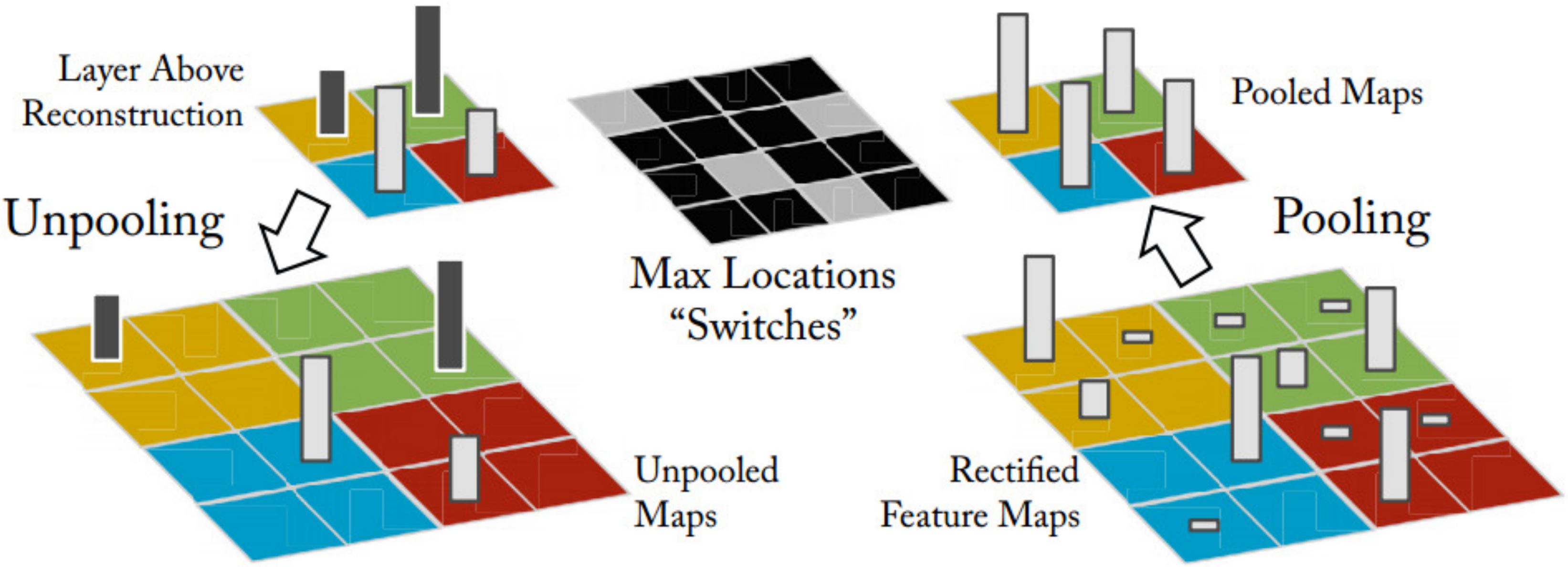}\tabularnewline
(a) & (b)\tabularnewline
\end{tabular}
\par\end{centering}
\caption[DeConvNet building blocks.]{\label{fig:deconvnet} DeConvNet building blocks. (a) Illustrates
a DeConvNet operation that can be applied to project the feature maps,
extracted from any layer of a ConvNet, back to image space. (b) Illustrates
the ``un-pooling'' operation via use of ``switches'', which correspond
to the locations responding to the max pooling operation. Figure reproduced
from \cite{Zeiler2014}.}
\end{figure}

The first approach to ConvNet visualization is known as a dataset-centric
approach \cite{yosinski2015} because it relies on probing the network
with input from a dataset to find maximally responding units in the
network. One of the earliest approaches falling under this category
is known as DeConvNet \cite{Zeiler2014}, where visualization is achieved
in two steps. First, a ConvNet is presented with several images from
a dataset and the feature maps responding maximally to this input
are recorded. Second, these feature maps are projected back to the
image space using the DeConvNet architecture, which consists of blocks
that invert the operations of the ConvNet used. In particular, DeConvNet
inverts the convolution operations (\ie performs ``de-convolution'')
via use of the transpose of the learned filters in the ConvNet under
consideration. Here, it is worth noting that taking the transpose
is not guaranteed to invert a convolution operation. For ``un-pooling'',
DeConvNet relies on recording the locations corresponding to max-pooled
responses in the ConvNet and uses those locations for ``un-pooling''
or upsampling the feature maps. These operations are summarized in
Figure \ref{fig:deconvnet}. 

\begin{figure}[H]
\begin{centering}
\includegraphics[width=0.75\textwidth]{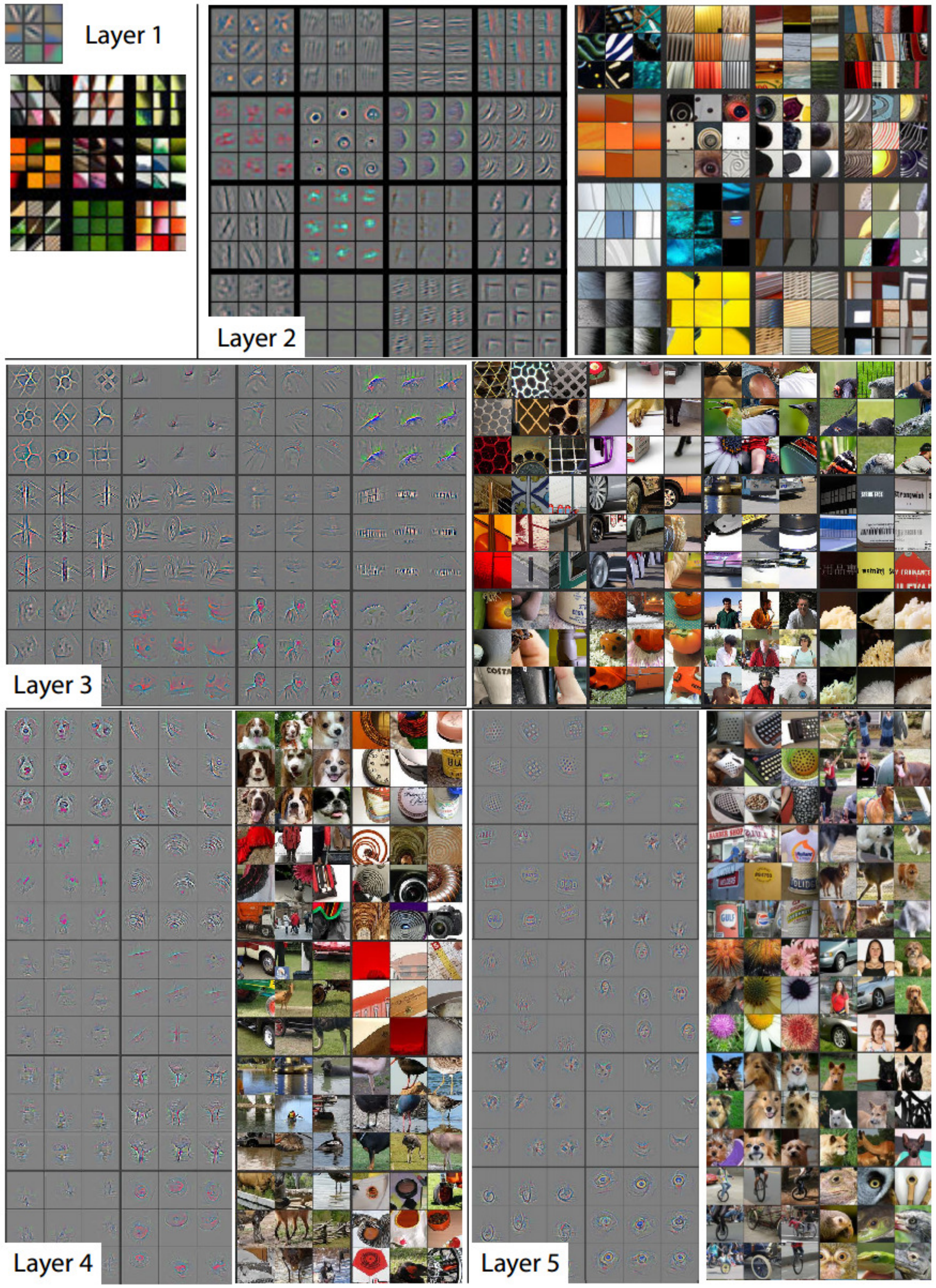}
\par\end{centering}
\caption[Visualization obtained by applying DeConvNet at various layer of a
standard ConvNet architecture]{\label{fig:deconv_visu}Visualization obtained by applying DeconvNet
at various layer of a standard ConvNet architecture such as AlexNet
\cite{Krizhevsky2012}. Figure reproduced from \cite{Zeiler2014}.}

\end{figure}

Typical visualizations resulting from these methods are shown in Figure
\ref{fig:deconv_visu}. Overall these visualization reveal that earlier
layers in the network tend to capture low level features such as oriented
bars and edges, \ie filters learned at lower layers are similar to
oriented bandpass filters. In contrast, at higher layers features
captured progress from simple textures to more complex objects. Interestingly,
these visualizations tend to conserve a high level of detail from
the images that yielded a high response in a network. In fact, it
seems like these visualization tend to emphasize the edges of the
input images and mainly reveal the part of the image that is responsible
for the high response (\ie they can be seen as methods for finding
the high contrast points of the input images and mainly reveal the
part that is responsible for high classification results). Motivated
by these observations, other approaches falling under the dataset-centric
paradigm proposed even simpler methods to visualize what a network
is learning. Examples include methods that progressively remove parts
from images yielding high responses to highlight what parts are responsible
for high responses \cite{Zhou2015,zhou2016}. Some of the conclusions
that emerged from these approaches are that objects are largely responsible
for recognizing scenes \cite{Zhou2015} or more generally that object
detectors emerge as we visualize higher layers of the network \cite{zhou2016}. 

The second approach to ConvNet visualization is known as a network-centric
approach \cite{yosinski2015} because it uses the network parameters
only without requiring any additional data for visualization purposes.
This approach was first introduced in the context of deep belief networks
\cite{Erhan2009} and later applied to ConvNets \cite{simonyan14deep}.
In this case, visualization is achieved by synthesizing an image that
will maximize some neuron's (or filter's) response. For example, starting
from the last layer of a network that yields a class score, $S_{c}$,
and an image initialized to random noise, $I$, the goal is to modify
the image such that its score for belonging to class $c$ is maximized.
This optimization problem is defined in \cite{simonyan14deep} according
to

\begin{equation}
\argmax_{I}S_{c}(I)-\lambda||I||_{2}^{2},\label{eq:vis}
\end{equation}
where $\lambda$ is a regularization parameter. Here the $L_{2}$
regularization is used to penalize large values. Most other methods
falling under this paradigm attempt to solve the same optimization
problem while enforcing different regularization techniques such as
total variation regularization to enforce smoothness and avoid high
frequency content in the synthesized image \cite{MahendranV15} or
simple clipping of pixels that do not participate strongly into the
filter's response to only highlight the patterns responsible for a
filter's activation \cite{yosinski2015}.

\begin{figure}
\begin{centering}
\begin{tabular}{c}
\includegraphics[width=0.7\textwidth]{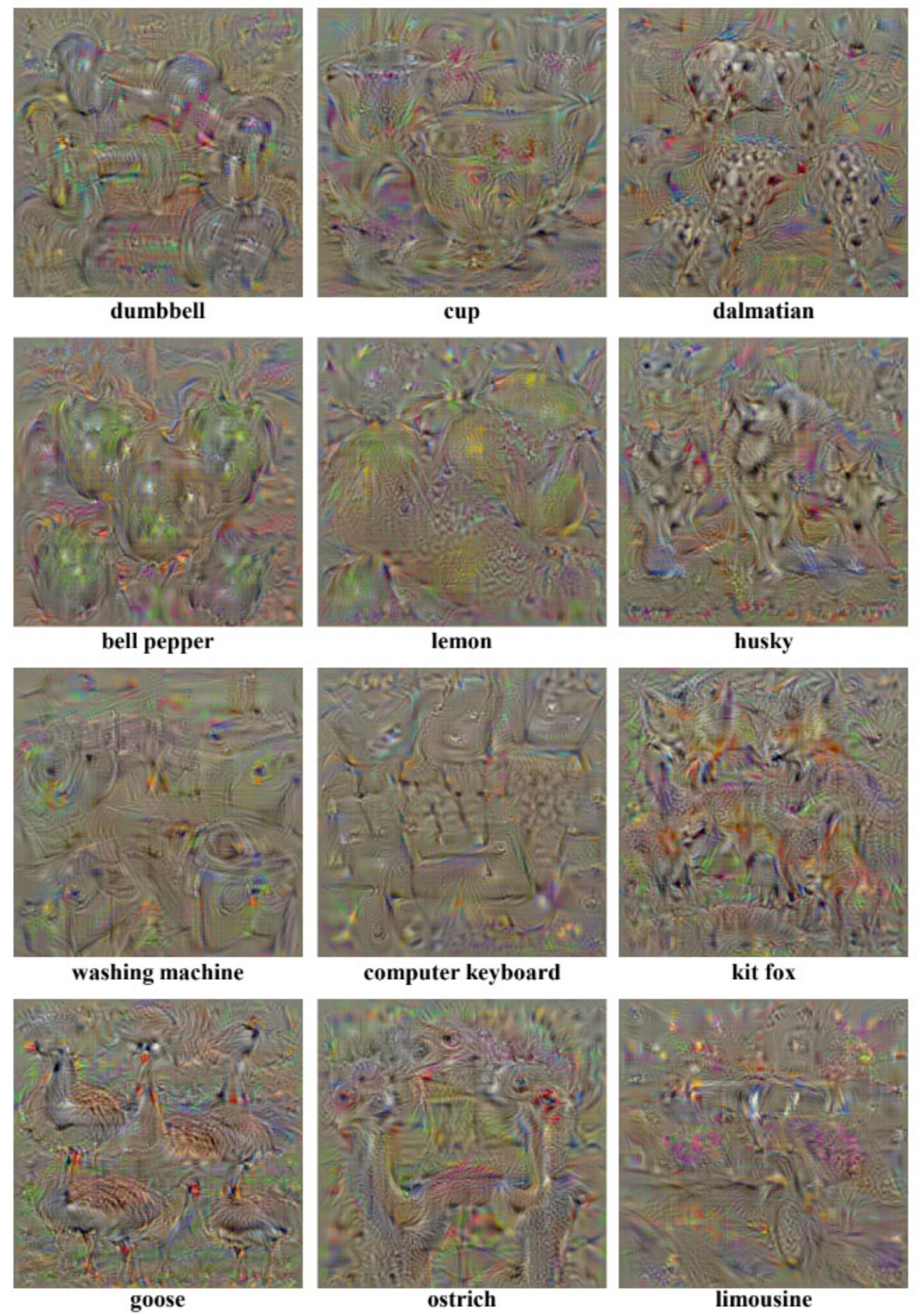}\tabularnewline
(a)\tabularnewline
\includegraphics[width=0.9\textwidth]{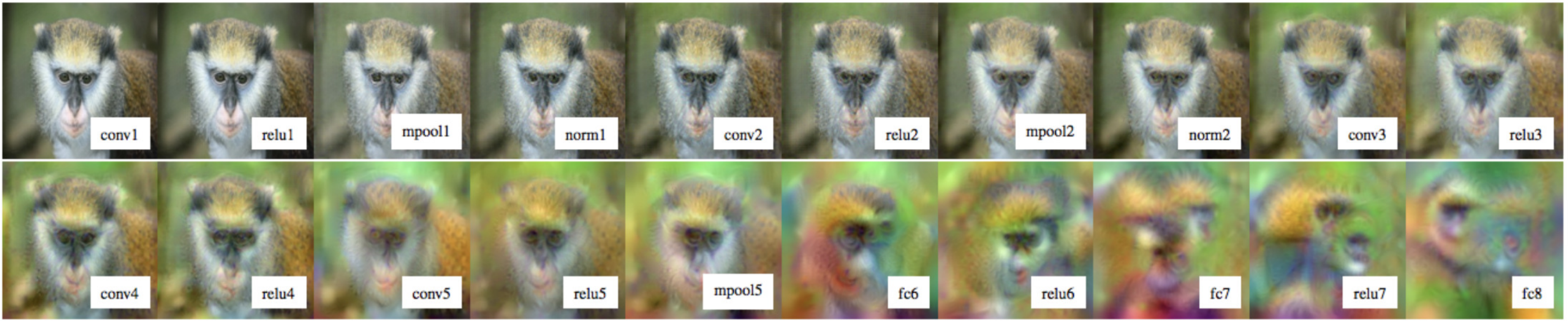}\tabularnewline
(b)\tabularnewline
\end{tabular}
\par\end{centering}
\caption[Visualization obtained by applying optimization in the image space
as done in the network-centric approaches.]{\label{fig:simo_visu}Visualization obtained by applying optimization
in the image space as done in the network-centric approaches. (a)
Visualization obtained by maximizing the score of belonging to various
classes, as shown below each image. Figure reproduced from \cite{simonyan14deep}.
(b) Visualizations obtained by maximizing the responses of a standard
network at various layers, as shown below each image. Figure reproduced
from \cite{MahendranV15}.}
\end{figure}

Typical visualization resulting from network-centric approaches when
used to maximize class scores as defined in \eqref{eq:vis} are shown
in Figure \ref{fig:simo_visu}. Usually, these visualizations suggest
that the network is learning high level shapes responsible for discriminating
various objects (\eg eyes, nose and ears when the target class is
animals faces); however, the exact locations of these features does
not matter. Therefore, these visualizations imply that the network
learns invariances progressively (\eg it becomes position agnostic
at higher layers), as expected from the use of various pooling operations.
However, when this network-centric visualization technique is applied
to invert lower layers, it does not necessarily show that they are
learning lower level features such as edges as opposed to the dataset-centric
technique. For example, in Figure \ref{fig:simo_visu}(b), it is seen
that lower layers retain most of the photographic information present
in the image. More generally, a major limitation of visualization
approaches to understanding ConvNets is the subjective nature of the
interpretation, which typically is based on visual inspection by the
authors of the method.

\subsection{Understanding ConvNets via ablation studies}

Another popular method to shed light on ConvNet capabilities that
is being widely used is the so called ablation study of the network.
In fact, many prominent ConvNet papers (\eg \cite{chatfield2014,Zeiler2014,SimonyanTwoStream,liminwang2016,Chris2017ACTIONS,dai17dcn,Jeon2017})
include an ablation study in their experimental section, where the
goal is to isolate the different components of the network to see
how removing or adding a block affects the overall performance. 

These ablation studies have the potential to guide ConvNet practitioner
towards ``good practices'' to achieve higher performance in different
applications. For example, one of the earliest ablation studies in
the context of ConvNets revealed the importance of proper rectification
and normalization even while using randomly initialized networks \cite{Jarret2009}.
Other work, revealed the importance of deeper representations while
using smaller filters at all layers \cite{chatfield2014,Simonyan14c}.
Yet other studies, investigated the role of pre-training and finetuning
as well as the number, location and magnitude of features. These investigations,
further highlighted the transferability of features extracted from
ConvNets across tasks \cite{pulkit2014}.

More interestingly, other work proposed to dissect ConvNets to investigate
the effect of each component on the interpretability of the representations
learned \cite{netdissect2017}. This approach relies on a dataset
with pixel level annotations, where each pixel is assigned several
``concept'' labels that include color, texture, object and scene
labels. Each unit in a ConvNet under consideration is evaluated for
its ability to generate a segmentation mask that matches a certain
concept-based segmentation mask $L_{c}$. In particular, each feature
map, $S_{k}$, in the network is converted to a binary mask, $M_{k}$,
where pixels are set to $1$ only if their activation exceeds a certain
pre-set threshold. Next, each unit, $k$, is assigned a score for
its ability to segment a given concept, $c$, according to 

\begin{equation}
IoU_{k,c}=\frac{\sum_{dataset}|M_{k}\cap L_{c}|}{\sum_{dataset}|M_{k}\cup L_{c}|}\label{eq:seg_concept}
\end{equation}
where $|.|$ is the cardinality of the set. With this measure, the
interpretability of each unit is defined based on its ability to generate
good segmentation masks. This measure revealed that units in lower
layers are able to generate better color or texture based masks, whereas
higher layers generate better object and scene segmentations. In line
with dataset-centric visualization approaches, this observation suggests
that lower layers are learning filters that capture lower level concepts
while higher layers learn more abstract features such as objects parts.
This approach also allowed for a systematic analysis of the effect
of various nonlinearities and training strategies on interpretability
and revealed that higher performance does not always yield highly
interpretable units. For example, it was shown that regularization
techniques such as batch normalization non-trivially affect a unit's
interpretability as defined in this approach. Also, it was found that
networks trained on scenes datasets yield more interpretable units
compared to the widely used ImageNet training. Unfortunately, one
of the main flaws of this approach is the fact that their definition
of interpretability depends highly on the dataset used for evaluation.
In other words, ConvNet units capable of capturing concepts that are
not represented in the dataset will yield low IoU scores; hence, deemed
not interpretable by this method, even if the concept is visually
interpretable. Therefore, this method can miss other important aspects
revealed by the network components responsible for higher performance. 

\subsection{Understanding ConvNets via controlled design}

Another method to understand ConvNets is to minimize the number of
learned parameters by injecting priors into the network design. For
example, some methods reduce the number of learned filters per layer
and include transformed versions of the learned filters in each layer
to model rotation invariances, \eg \cite{Luan2017,Zhou2017ORN}.
Other approaches rely on replacing learned filters with a basis set
and instead of learning filter parameters they learn how to combine
the basis set to form the effective filters at each layer, \eg\cite{Jacobsen16,Cohen2017,Worral2017,Luan2017,Zhou2017ORN}.
Yet other approaches, push the idea of injecting priors into their
network design even further by fully hand crafting their network and
adapting it to a given task, which yields especially interpretable
networks, \eg \cite{Bruna2013,Oyallon2015,ISMA2017}. Most of the
techniques falling under this paradigm were previously discussed in
details in Section \ref{subsec:Conv-Theoretical-perspective} of Chapter
\ref{cha:chapter3}.\vspace{-15pt}

\section{Open problems}

This report documented the significant progress made in the design
of various novel ConvNet architectures and building blocks. While
this progress resulted in a new state-of-the-art in several computer
vision applications, understanding of how these ConvNets achieve those
results lags behind. Moreover, there is little understanding of the
performance limitations (\eg failure modes) of these approaches.
Therefore, shedding light on what information is captured by those
ConvNets is becoming particularly relevant. Currently, approaches
that focus on understanding ConvNets are becoming apparent in the
related literature as discussed throughout this chapter. However,
while several of the techniques discussed here are taking good steps,
they all leave open critical questions that remain to be answered.%

In fact, the review on the different techniques that aim at explaining
ConvNets revealed that the most widely adopted approach relies on
visualizations. However, one of the biggest flaws of visualization
based techniques is the fact that they reduce understanding of a complex
and highly nonlinear model to a single image that is open to various
potential interpretations. Notably, these visualizations vary according
to the adopted technique (\eg dataset-centric versus. network-centric,
as shown in Figures \ref{fig:deconv_visu} and \ref{fig:simo_visu})
and usually also depend on the architecture under consideration as
well as the training strategy \cite{netdissect2017}. Their interpretation
also is highly subjective. More importantly, it was shown in a recent
study that replacing a strong response of a feature map with random
noise and propagating it back through the DeConvNet architecture yields
similar visualizations to that obtained when projecting back the feature
map response itself \cite{mahendran16salient}. Therefore, this study
showed that dataset-centric visualizations do not necessarily reveal
what a particular feature map is capturing because they themselves
rely on parameters learned by the network to generate their visualizations.
Thus, based on this discussion, the following points emerge as potential
key ways forward for visualization-based approaches:
\begin{itemize}
\item First and foremost, it is of paramount importance to develop ways
to make the evaluation of visualizations more objective, via introduction
of a metric that evaluates the quality and/or meaning of the generated
images.
\item Also, while it appears that network-centric visualization approaches
are more promising, as they don't rely on a network themselves in
generating their visualization (\eg DeConvNet), it appears necessary
to standardize their evaluation process as well. One possible solution
is to use a benchmark for generating the visualizations and networks
trained under the same conditions. Such standardization, can in turn
allow for a metric-based evaluation instead of the current interpretation
based analysis.
\item Another way forward is to visualize multiple units at the same time
to better capture distributed aspects of the representations under
study, even while following a controlled approach.
\end{itemize}
Considering ablation studies, while they allow for isolating parts
of a network to identify components responsible for better performance,
they cannot really elucidate what a network learned as they try to
explain ConvNets' highly intertwined components in isolation. Notably,
in their current application, ablation studies are simply used as
ways to glean a few percentage points in performance without really
adding value from an understanding point of view. Here, potentially
interesting ways forward for ablation-based approaches include:
\begin{itemize}
\item Use of a common and systematically organized dataset that captures
different challenges commonly encountered in computer vision (\eg
view point and illumination changes) and that entails categories with
an increased level of complexity (\eg textures, parts and objects).
In fact, an initial dataset of this kind was recently introduced \cite{netdissect2017}.
Use of ablation studies on such a dataset, together with an analysis
of the resulting confusion matrices can allow pinpointing the failure
modes of ConvNet architectures and thereby lend better understanding.
\item In addition, systematic studies of how multiple coordinated ablations
impacts performance is interesting. Such studies should extend insight
beyond how isolated units perform.
\end{itemize}
Finally, although there are flaws in both visualization and ablation
studies approaches to understanding ConvNets, they still shed some
light on ConvNet shortcomings, such as the redundancies in the learned
filters (\eg see Figure \ref{fig:CReU-1}) and importance of certain
nonlinearities (\eg \cite{Jarret2009,chatfield2014}). These insights
were in turn used in more controlled approaches to ConvNet realization
that are less reliant on data and more importantly less obscure (\eg
\cite{Bruna2013,Oyallon2015,Jacobsen16,Jeon2017,ISMA2017}). These
controlled approaches are emerging as a promising direction for future
research as they lend deeper understanding of the operations and representations
that these systems employ relative to purely learning based approaches.
Indeed, many such insights were reviewed in the previous chapter.
In turn, they also have the potential to support more rigorous performance
bounds based on their precise system specifications. Here interesting
ways forward include:
\begin{itemize}
\item Progressively fixing network parameters and analyzing impact on a
network's behavior. For example, fixing convolutional kernel parameters
(based on some prior knowledge of the task at hand) one layer at a
time to analyze the suitability of the adopted kernels at each layer.
This progressive approach has the potential to shed light on the role
of learning and can also be used as an initialization to minimize
training time.
\item Similarly, design of the network architecture itself can be investigated
(\eg number of layers or filters per layer) via analysis of the input
signal properties (\eg the frequency content of the signal). This
method can help adapt the architecture's complexity to the application.
\item Finally, use of controlled approaches to network realization can be
accompanied with systematic studies of the role of other aspects of
ConvNets that usually receive less attention due to the focus on learned
parameters. Examples include investigation of various pooling strategies
and the role of residual connections when most learned parameters
are otherwise fixed.
\end{itemize}

\newpage \addcontentsline{toc}{chapter}{Bibliography}

\bibliographystyle{plain}
\bibliography{mybibtex}

\end{document}